\documentclass{article} 
\usepackage{iclr2027_conference,times}

\usepackage{amsmath,amsfonts,bm}

\def\eqref#1{equation~\ref{#1}}

\def\1{\bm{1}}

\DeclareMathAlphabet{\mathsfit}{\encodingdefault}{\sfdefault}{m}{sl}
\SetMathAlphabet{\mathsfit}{bold}{\encodingdefault}{\sfdefault}{bx}{n}

\usepackage{hyperref}
\usepackage{url}

\usepackage{booktabs}  
\usepackage{amsfonts}       
\usepackage{nicefrac}       
\usepackage{microtype}      
\usepackage[dvipsnames, svgnames, x11names]{xcolor} 

\usepackage{amsthm}
\usepackage{amsmath}
\usepackage{graphicx}

\usepackage{amssymb}

\usepackage{multirow}
\usepackage{wrapfig}
\usepackage{subcaption}
\usepackage{makecell}
\usepackage{threeparttable}
\usepackage{longtable}

\usepackage[utf8]{inputenc}
\usepackage{algorithm}       
\usepackage{algpseudocode}
\usepackage[normalem]{ulem}
\usepackage{placeins}

\title{Think Before You Forecast: Scaling Test-Time Compute for Time Series Forecasting}

\author{
\vspace{2.5mm}
Daojun Liang \textsuperscript{\rm 1,\rm 2} \quad 
Xiao Hong \textsuperscript{\rm 3}  \thanks{Corresponding author.} \quad 
Pei Zhang \textsuperscript{\rm 1,\rm 2} \footnotemark[1] \quad 
Yinglong Wang \textsuperscript{\rm 1,\rm 2} \quad 
\textbf{Shuo Li} \textsuperscript{\rm 4, \rm 5} \\
\vspace{2.5mm}
\textbf{Jing Chen} \textsuperscript{\rm 1,\rm 2} \quad 
\textbf{Ying Guo} \textsuperscript{\rm 1,\rm 2} \quad
\textbf{Hu Zhang} \textsuperscript{\rm 1,\rm 2} \\
\textsuperscript{\rm 1} Key Laboratory of Computing Power Network and Information Security, Ministry of Education, \\
\quad Shandong Computer Science Center (National Supercomputing Center in Jinan), \\ 
\quad Qilu University of Technology (Shandong Academy of Sciences), Jinan, 250103, China \\
\textsuperscript{\rm 2} Shandong Provincial Key Laboratory of Computing Power Internet and Service Computing, \\ 
\quad Shandong Fundamental Research Center for Computer Science, Jinan, 250103, China \\
\textsuperscript{\rm 3}  Shandong Normal University, Jinan, 250103, China \\
\textsuperscript{\rm 4} Department of Computer and Data Sciences, Case Western Reserve University, Cleveland, USA \\
\textsuperscript{\rm 5} Department of Biomedical Engineering, Case Western Reserve University, Cleveland, USA \\ 
\quad \texttt{liangdj@sdas.org, hongxiao@sdnu.edu.cn} 
}

\iclrfinalcopy 
\begin{document}

\maketitle

\begin{abstract}
Test-time compute has become an important paradigm in computer vision and language modeling, where additional computation at inference time enables models to explore multiple candidate solutions and select or refine the most reliable one. However, time series forecasting models typically produce predictions through a single forward pass, leaving them vulnerable to distribution shifts and non-stationarity that frequently arise in real-world temporal data. 
We propose \textbf{T}ime \textbf{T}est \textbf{C}ompute for \textbf{T}ime \textbf{S}eries (TTC4TS), an architecture‑agnostic framework that improves the predictive performance of frozen forecasting models at inference time. Given a pretrained backbone, the generator produces multiple candidate forecast trajectories using augmentation, retrieval, adapter. The synthesizer then selects or combines these candidates using aggregation, classification, or candidate-scoring and ranking methods to produce the final forecast.
By increasing the number of candidates, TTC4TS scales test-time compute without retraining the forecasting backbone. 
Beyond improving point-forecast accuracy, TTC4TS can exploit candidate forecasts for uncertainty quantification, which supports improved interval forecasts.
We also theoretically analyze when additional test-time computation improves forecasting
Experiments across diverse datasets and backbones demonstrate consistent gains in accuracy and calibration. 
Code is provided in \href{https://github.com/Anoise/TTC4TS}{Github}.
\end{abstract}

\section{Introduction}

Time Series Forecasting (TSF) is a fundamental problem in machine learning, supporting decision-making in energy, transportation, finance, healthcare, retail, and scientific discovery, etc. Recent years have witnessed remarkable progress in neural forecasting models, including  recurrent \citep{connor1994recurrent, salinas2020deepar} or convolutional neural networks \citep{lea2016temporal, wu2022timesnet, luo2024moderntcn}, Transformers \citep{Zhou2021Informer,wu2021autoformer,zhou2022fedformer, Yu2023PatchTST,  liu2023itransformer, wang2024timexer}, linear architectures \citep{zeng2023transformers, yue2026olinear}, spectral models \citep{yi2024frequency, yi2024filternet}, and emerging time series foundation models \citep{garza2023timegpt, jin2024timellm, liu2024unitime, TimesFM2024, ansari2024chronos}. Despite these advances, most forecasting systems still follow the same inference paradigm: after training, a model receives a historical context window $X$ and produces a future trajectory $\hat{Y}=F(X)$ through a single forward propagation step.
This single-shot prediction paradigm is simple and efficient, but often fragile because the model tends to be overly confident in its one-off deterministic outputs \citep{lakshminarayanan2017simple}.


In contrast, recent progress in vision and Large Language Models (LLM) suggests a different direction: test-time compute \citep{ma2025scaling, snell2025scaling}. Rather than relying on a single output, a model can allocate additional computation during inference to generate, refine, compare, or select among multiple candidate solutions. This paradigm has shown that better decisions can often be obtained not only by scaling training, model size, or data, but also by using computation more effectively at test time. However, this powerful tool remains largely unexplored in TSF. Existing forecasting models, including large time series models \citep{TimesFM2024, ansari2024chronos} or LLM-based reasoning models \citep{LLM4TS2026}, still behave as deterministic one-pass predictors (quantile outputs are also one-pass), producing only one future trajectory even when the input is ambiguous, shifted, or highly non-stationary. 

This paper asks a simple question: \emph{Can we introduce test-time compute into TSF without retraining or modifying the deployed forecasting backbone?}

\begin{figure*}
  \centering
  \centerline{\includegraphics[width=\textwidth]{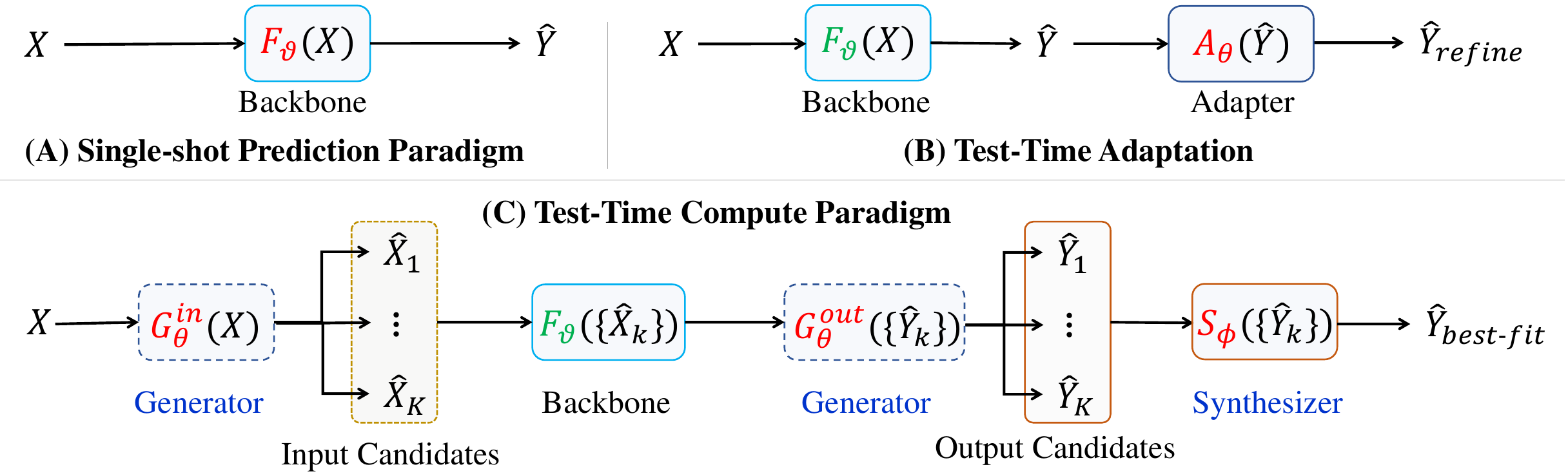}}
  \caption{Comparison of time series inference paradigms ({\color{red}red}: trainable; {\color{green}green}: frozen). (A) Standard learning relies on a single deterministic forward pass. (B) Test-Time Adaptation refines the single output, often requiring parameter updates. (C) Our proposed Test-Time Compute framework scales compute at inference by generating multiple diverse candidates, followed by a synthesis (Think) step.}
  \label{fig_paradigm}
  \vspace{-0.5cm}
\end{figure*}

We answer this question affirmatively by proposing Candidate Aggregation and Scaling at Test time (TTC4TS), a test-time compute mechanism tailored to TSF. 
The main idea is to keep the pretrained \text{forecaster} $F$ unchanged, generate a large number of candidate trajectories via the \textbf{Generator} $G_{\theta}$, and then select or aggregate these candidates through the \textbf{Synthesizer} $S_{\phi}$ to produce the best-fit prediction. Figure~\ref{fig_paradigm} illustrates three different inference paradigms for TSF. The traditional single-shot prediction paradigm setting shown in Figure ~\ref{fig_paradigm}(A), the model is trained or updated using observed data, but prediction itself still depends on a single deterministic forward pass. Test-time adaptation, shown in Figure~\ref{fig_paradigm}(B), attempts to refine the prediction at inference time, often by adapting model parameters or correcting the single output under some auxiliary objective. Although such methods improve flexibility, they still typically operate around one forecast trajectory and may require delicate update rules to avoid instability or leakage.

In contrast, the test-time compute paradigm shown in Figure~\ref{fig_paradigm}(C) changes the nature of inference: instead of committing immediately to one trajectory, the model generates multiple candidate forecasts and then performs a selection or aggregation step before producing the final prediction. This final step is the key ``thinking'' process in our framework.
TTC4TS generates multiple candidate forecasts by applying small, bounded modifications (e.g., multiple adapters) or augmentation (e.g., tunable transformation operators) to either the input or the output of the frozen forecaster $F$, or by retrieving similar historical contexts to produce diverse candidate trajectories $\{\hat{Y}_1,\hat{Y}_2,\ldots,\hat{Y}_K\}$.
The synthesizer then selects or combines these candidates to produce the final forecast. The synthesizer can be implemented in several forms, including simple averaging, weighted aggregation, horizon-wise selection, or a lightweight classifier/scorer that chooses the most reliable trajectory. In this way, TTC4TS converts TSF from a single-shot prediction process into a candidate-generation and synthesis process, enabling additional inference-time reasoning while preserving the original forecasts.


Beyond improving point forecasting accuracy, TTC4TS also provides an additional benefits: The multiple candidate trajectories naturally support Uncertainty Quantification (UQ). The diversity among candidates can be used to construct prediction intervals and estimate coverage that provide good and personalized uncertainty estimates.

Our contributions are summarized as follows:
\begin{itemize} 
\item  An architecture-agnostic framework is introduced, which improves frozen time-series forecasters through candidate generation and aggregation, instantiating 3 generator families and 8 synthesis strategies with explicit pre-test fitting and causal information access.
\item  We characterize when expanding candidate pools improves forecasting by separating gains
in attainable prediction quality from aggregation error, explaining why additional candidates
require reliable synthesis rather than guaranteeing lower squared-error risk.
\item  We extend TTC4TS through separate conformal calibration of candidate-derived prediction intervals, improving aggregate interval quality while demonstrating that candidate diversity alone does not provide calibrated uncertainty.
\item  Extensive experiments show that TTC4TS delivers substantial performance gains, achieving
an average improvement of 6.4\% and reducing errors in 420/432 evaluated configurations.
\end{itemize}


\begin{figure*}
  \centering
  \centerline{\includegraphics[width=\textwidth]{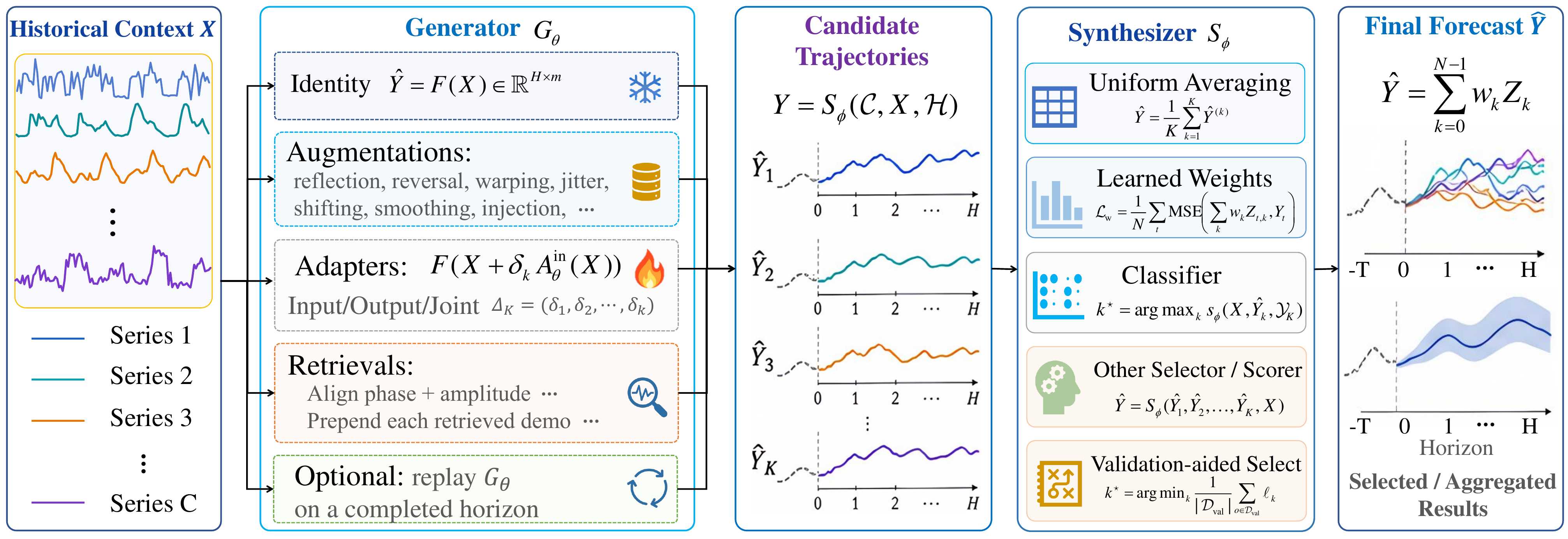}}
  \caption{Overview of the TTC4TS framework for test-time compute in time series forecasting.}
  \label{fig_arch}
\end{figure*}

\section{Methodology}

The overall architecture of TTC4TS is shown in Figure~\ref{fig_arch}. 
We first present the problem setup (Section~\ref{sec2_setup}), and then introduce the generator (Section~\ref{sec2_gen}) and the synthesizer (Section~\ref{sec2_syn}).


\subsection{Problem Setup}
\label{sec2_setup}

Let $\mathcal{D} = \{(X^{(t)}, Y^{(t)})\}_{t=1}^T$ denote a training set of context–target pairs, where each context window $X^{(t)} \in \mathbb{R}^{L \times C}$ covers a history of length $L$ over $C$ covariates, and each future target $Y^{(t)} \in \mathbb{R}^{H \times m}$ spans a forecast horizon $H$ across $m$ target dimensions (we omit the superscript $(t)$ hereafter for brevity). A pre-trained forecaster $F$ maps a context window $X$ to predictions $\hat{Y} = F(X) \in \mathbb{R}^{H \times m}$. The proposed framework, TTC4TS, comprises two components: a generator $G_\theta$, which produces a set of $K$ candidate forecasts, and a synthesizer $S_\phi$, which combines them. 
The framework replaces single-shot inference with a bank of candidate trajectories followed by synthesis:
\begin{equation}
\begin{aligned}
\mathcal C&=G_\theta(F,X;\mathcal B)
=[Z_{0},\ldots,Z_{N-1}],\quad \text{then,} \quad
\widehat Y&=S_\phi(\mathcal C,X,\mathcal H).
\end{aligned}
\label{eq_most_framework}
\end{equation}
Here, $\mathcal B$ is an eligible historical reference bank, and $\mathcal H=X^{\{0:t\}}$ denotes already observed history. The generator explores alternative forecasts through residual adapters, transformations, and retrieved demonstrations. The synthesizer assigns reliability weights using fixed coefficients, an offline-trained scorer, or observed retrospective errors. The current target $Y$ is never an input to either component. Auxiliary learning is permitted, but no backbone or auxiliary parameter is updated on a test query.

\subsection{Generator: Candidate Trajectories Generation}
\label{sec2_gen}

\textbf{Augmentation-based Candidates}:
For an input transformation $T_m$ and its associated forecast correction $U_m$, we form $ Z_{m}^{\mathrm{aug}}=U_m\!\left(F(T_m(X))\right)$.
The implemented bank contains $A$ candidates, including magnitude reflection, time reversal, magnitude warping, time warping, exponentially weighted moving-average smoothing, jitter, horizontal shifting, positive and negative vertical shifts, positive and negative slope changes, and spike injection. Transformation scales depend only on the observed context. Specifically, with $\mu_c=L^{-1}\sum_t X_{t,c}$, define $s_c=\max(\sqrt{\delta_{\mu}},\, \delta )$, where $\delta_{\mu} = L^{-1}\sum_t(X_{t,c}-\mu_c)^2$.

In particular, reversing the input history does not reverse the output forecast. Operators with $U_m$ equal to the identity are candidate-generating perturbations, not assumed invertible forecasting symmetries. All operator definitions, constants, and the jitter construction are provided in Appendix~\ref{app_gen_aug}.

\textbf{Multi-Bounded Adapters}:
For $Z\in\mathbb R^{T\times C}$, an adapter predicts a residual using a lightweight neural network. Rather than fitting a separate adapter for every candidate, we use two shared adapters, $r_{\theta_x}$ and $r_{\theta_y}$, and vary their adjustment strength. For $K\geq2$, $\delta_j=\frac{j}{K-1}\delta_{\max}, j\in [1,K].$
The input, output, and joint paths are
\begin{equation}
\begin{aligned}
Z_{j}^{\mathrm{in}}
&=F\!\left(X+\delta_jr_{\theta_x}(X)\right), \quad
Z_{j}^{\mathrm{out}}
&=Z_{0}+\delta_jr_{\theta_y}(Z_{0}), \quad
Z_{j}^{\mathrm{joint}}
&=Z_{j}^{\mathrm{in}}
+\delta_jr_{\theta_y}(Z_{j}^{\mathrm{in}}).
\end{aligned}
\label{eq_adapter}
\end{equation}
Each direct additive correction is bounded elementwise by $\delta_j$ in standardized units. This does not impose a corresponding bound on the backbone's response to an input perturbation. The input and joint paths share $\theta_x$, and the output and joint paths share $\theta_y$. 
Further, the adapters are fitted on $\mathcal D_{\mathrm{adapt}}$ by minimizing the mean forecast error across all paths:
\begin{equation}
\mathcal L_{\mathrm{adapt}}
=\frac{1}{|\mathcal D_{\mathrm{adapt}}|}
\sum_{(X,Y)\in\mathcal D_{\mathrm{adapt}}}
\frac{1}{3K}\sum_{j=0}^{K-1}
\sum_{v\in\{\mathrm{in,out,joint}\}}
\frac{\|Z_{j}^{v}-Y\|_F^2}{HC}.
\label{eq:most:adapter-loss}
\end{equation}
Please refer to Appendix~\ref{app_gen_adapter} for more implementation details.

\textbf{Retrieval-based Candidates}:
Retrieval supplies a separate candidate for each of $M$ historical neighbors. A training-bank entry contains both a reference past and its already observed continuation: $ R_n=\{X^{a_n}:X^{a_n+S+H} \}$,
where $a_n$ is the starting index and $S$ is the length of the reference past.
Further, using $Q=X[L-S:L]$ and $P_n=R_n[0:S]$, retrieval compares each channel after local normalization:
\begin{equation}
\begin{aligned}
\mathcal Z(z)_t&=\frac{z_t-\operatorname{mean}(z)}
{\max(\operatorname{std}(z),0.01)}, \qquad
d_{n,c}&=\frac{1}{S}\sum_{t=0}^{S-1}
\left[\mathcal Z(Q_{\cdot,c})_t-\mathcal Z(P_{n,\cdot,c})_t\right]^2.
\end{aligned}
\end{equation}
For each channel $c$, we retain the $M$ smallest-distance entries and assemble the $j$th demonstration as $R^{(j)}_{t,c}=R_{n_{j,c},t,c}$. 
Let $R^{\mathrm{phase},j}$ be the shifted pair. Its first $S$ positions determine $\mu_{r,j,c}$ and $\sigma_{r,j,c}$; the query determines $\mu_{q,c}$ and $\sigma_{q,c}$. The amplitude alignment is given by
\begin{equation}
\begin{aligned}
\alpha_{j,c}
&=\frac{\sigma_{q,c}\sigma_{r,j,c}}
{\sigma_{r,j,c}^2+(\sigma_{q,c}/K)^2},\quad
\beta_{j,c}&=\mu_{q,c}-\alpha_{j,c}\mu_{r,j,c},\quad
\widetilde R^{(j)}_{t,c}
&=\alpha_{j,c}R^{\mathrm{phase},j}_{t,c}+\beta_{j,c}.
\end{aligned}
\label{eq:most:alignment}
\end{equation}
The continuation inside each demonstration is historical data, not the current query's future. For further details, please refer to Appendix~\ref{app_gen_retrieval}, which specifies the exact phase-alignment calculation.

\subsection{Synthesizer: Candidate Selection or Aggregation}
\label{sec2_syn}

\textbf{Candidate-Bank Construction}:
Consistent with Eq.~\ref{eq_most_framework}, the generator concatenates the baseline, nonzero-strength adapter candidates, transformed-input candidates, and retrieved-demonstration candidates in a fixed order:
\begin{equation}
\begin{aligned}
\mathcal C=\big[&Z_{0},
\{Z_{m}^{\mathrm{aug}}\}_{m=1}^{A},
\{Z_{j}^{\mathrm{in}}\}_{j=1}^{K-1},
&\{Z_{j}^{\mathrm{out}}\}_{j=1}^{K-1},
\{Z_{j}^{\mathrm{joint}}\}_{j=1}^{K-1},
\{Z_{j}^{\mathrm{ret}}\}_{j=1}^{M}\big].
\end{aligned}
\label{eq:most:bank}
\end{equation}
Because every zero-strength adapter path recovers $F(X)$, this shared candidate is included only once. Thus, $N=1+3(K-1)+A+M$. Please refer to Appendix \ref{app_alg} for the specific algorithm.

\textbf{Synthesis Strategies}:
The synthesis strategies $S_\phi$ encompass a diverse range of approaches. In this work, we implement 8 distinct Synthesis Strategies, each designed to identify the synthesized forecast from a large pool of candidates. Therefore, after generating $N$ candidate trajectories $\mathcal{Y}_K=\{\hat{Y}_1,\hat{Y}_2,\ldots,\hat{Y}_K\}$, we map them to the final forecast using a function $\hat{Y}= S_\phi(\mathcal{Y}_K,X)$.
Here, $S_\phi$ follows a general rule, i.e., the synthesizer selects or combines generated candidates through a unified convex aggregation rule
\begin{equation}
\widehat Y=\sum_{k=0}^{N-1}w_{k}Z_{k},
\qquad w_{k}\geq0,
\qquad\sum_{k=0}^{N-1}w_{k}=1.
\label{eq_synthesis}
\end{equation}
Each candidate receives one weight for its entire forecast trajectory. One-hot weights implement hard selection, while distributed weights implement aggregation. 

We consider uniform averaging, learned input-dependent synthesis (ensemble learning, learned weights, classifier, risk regressor, pairwise ranker), Retrospective Weighting and Validation-aided Selection. 
Please refer to Appendix~\ref{app_syn} for implementation details.

\section{When Does Additional Computation Help?}
\label{when_ttc_help}

\textbf{Forecast Geometry and Useful Diversity}:
Condition on available information $\mathcal H_t$, including fitted parameters and generator draws. Set $d=HC$, $\langle a,b\rangle_d=a^\top b/d$, $\mu=\mathbb E[Y_t\mid\mathcal H_t]$, and $v=\mathbb E[\|Y_t-\mu\|_d^2\mid\mathcal H_t]$. For observable simplex weights and $z_w=\sum_kw_kZ_k$,
\begin{equation}
 R(w)=v+\|z_w-\mu\|_d^2=w^\top Qw
 =\sum_kw_kR_k- \tfrac12\sum_{i,j}w_iw_j\|Z_i-Z_j\|_d^2.
 \label{eq:geometry}
\end{equation}
Here, $Q_{ij}=\mathbb E[\langle Z_i-Y_t,Z_j-Y_t\rangle_d\mid\mathcal H_t]$ includes bias and $R_k=Q_{kk}$. Diversity reduces risk relative to weighted individual risks, but need not improve accuracy.
$z^*$ projects $\mu$ onto the candidate hull. Adding $z$ strictly improves this oracle exactly when $\langle z^*-\mu,z-z^*\rangle_d<0$ (Appendix~\ref{app:geometry}).

Moreover, adapter-generated candidates are not independent. Appendix~\ref{app:adapter_theory} derives approximation bounds under smoothness assumptions. Refitting adapters changes the paths and invalidates fixed-generator nesting.

\textbf{Aggregation Error and the Value of Computation}:
For static learned weights, let $\bar Q$ be the deployment error second-moment and $\widehat Q$ its fitting estimate. If their largest entrywise difference is at most $\eta$ and empirical optimization error is at most $\gamma$, then
\begin{equation}
 \bar R(\widehat w)-\min_{w\in\Delta_K}\bar R(w)\leq2\eta+\gamma,
 \label{eq:weights_bound}
\end{equation}
where $\Delta_K$ is the simplex. A nested pool improves deployed risk exactly when its oracle gain exceeds the increase in aggregation excess (Appendix~\ref{app:weights_theory}). 
For exponential weights $w_k\propto\exp(-s_k/T)$, $T>0$, the separate assumption $\max_k|s_k-R_k|\leq\varepsilon$ gives
\begin{equation}
 R(w)\leq\min_k R_k+2\varepsilon+T\log K-D(w).
 \label{eq:risk}
\end{equation}
Appendices~\ref{app:score_theory} and \ref{app:temporal_theory} prove the prior-dependent extension, exact identity-shrinkage correction, and historical-score concentration bound with a temporal-drift term. 
The recovered accuracy gain must be assessed against the measured computational cost. (Appendix~\ref{app:budget_theory}). 


\section{Candidate-informed Uncertainty Quantification}
\label{sec_uncertainty}

\textbf{From Candidate Diversity to Prediction Intervals}:
In this section, we study whether candidate information helps a separate uncertainty head, followed by residual calibration. This extension follows locally adaptive conformal prediction and conformalized quantile regression \citep{romano2019conformalized}.
Let $\mu_{t,h,c}$ be the selected point forecast and let $s_{t,h,c}>0$ be an uncertainty scale.
The family-balanced displacement is
\begin{equation}
 d_{t,h,c}^{2}=\frac{1}{3}\sum_{g\in\{\mathrm{adapter},\mathrm{augmentation},\mathrm{retrieval}\}}
 \frac{1}{|\mathcal K_g|}\sum_{k\in\mathcal K_g}(Z_{t,k,h,c}-\mu_{t,h,c})^2.
\end{equation}
where $\mathcal K_g$ is the set of candidate indices belonging to generator family $g$.
The identity is excluded from these three families but retained in empirical candidate quantiles, and equal‑family mass prevents the many adapter grid points from dominating this dispersion statistic.

\textbf{Candidate-scale CP}:
After the trained forecaster is fixed, a separate calibration block gives
\begin{equation}
 R_{i,h,c}=\frac{|Y_{i,h,c}-\mu_{i,h,c}|}{s_{i,h,c}},\qquad
 q_{h,c}(\alpha)=R_{(\lceil(n+1)(1-\alpha)\rceil),h,c}.
\end{equation}
Here, $n$ is the number of calibration origins and parentheses denote order statistics across origins, separately for each lead and channel. An order above $n$ returns infinity rather than an unjustified finite maximum. The interval is
\begin{equation}
 I_{t,h,c}(\alpha)=\left[\mu_{t,h,c}-q_{h,c}(\alpha)s_{t,h,c},\;
 \mu_{t,h,c}+q_{h,c}(\alpha)s_{t,h,c}\right].\label{eq:uq_interval}
\end{equation}
Under exchangeability of calibration and future scores, this construction has the usual marginal split-conformal interpretation. Temporal dependence and shifts can violate exchangeability. Our experiments report empirical coverage, without claiming finite-sample coverage under arbitrary time-series shifts or coverage conditional on every history.

\section{Experiments}
\label{sec_exp}

We validate TTC4TS on 6 datasets, 6 backbones, 4 horizons and 3 seeds (432 cells, see Appendix~\ref{app_imple} for more details). Results on three additional datasets are reported in Appendix~\ref{app_exp_reslut}. We test its gains when applied to pre-trained and SOTA models (Section \ref{sec_exp_main}), the effectiveness of generator (Section \ref{sec_exp_gen}), the effectiveness of synthesizer (Section \ref{sec_exp_syn}), and UQ performance in Section \ref{sec_exp_uq}.

\subsection{Main Results}
\label{sec_exp_main}

\begin{table}[H]
\centering
\caption{TTC4TS and matched adaptation/retrieval variants on 6 datasets, 6 backbones, 4 horizons and 3 seeds (432 cells). Relative errors are geometric means of ratios to Frozen in each cell. Detailed results are provided in Appendix~\ref{app:baseline_comparison}. Seed SD summarizes 3 seed-level aggregates. Wins counts lower MSE than Frozen, and TTC wins counts cells where TTC4TS beats the row.}\label{tab:baseline_summary}
\fontsize{9.2}{10.2}\selectfont
\setlength{\tabcolsep}{3.2pt}
\begin{tabular}{lrrrrr}
\toprule
Method & Rel. MSE & Rel. MAE & Seed SD & Wins & TTC wins \\
\midrule
Frozen backbone & 1.000 & 1.000 & 0.0000 & 0/432 & 420/432 \\
SOLID \citep{Chen2024SOLID} & 1.019 & 1.055 & 0.0002 & 2/432 & 428/432 \\
TAFAS \citep{Kim2025TAFAS} & 1.002 & 1.042 & 0.0014 & 166/432 & 402/432 \\
DynaTTA \citep{Grover2025DynaTTA} & 1.015 & 1.067 & 0.0068 & 112/432 & 424/432 \\
AdaPTS  \citep{Bene2025AdaPTS} & 0.993 & 1.009 & 0.0034 & 216/432 & 410/432 \\ 
$\delta$-Adapter  input \citep{Liang2026adapter} & 0.981 & 1.013 & 0.0097 & 251/432 & 372/432 \\
$\delta$-Adapter output \citep{Liang2026adapter}  & 0.994 & 1.017 & 0.0067 & 212/432 & 410/432 \\
$\delta$-Adapter composed \citep{Liang2026adapter} & 0.992 & 1.049 & 0.0163 & 189/432 & 386/432 \\
TS-RAG \citep{ning2026tsrag} & 0.985 & 1.007 & 0.0033 & 281/432 & 414/432 \\
Align-RAG \citep{asadi2026align} & 0.981 & 1.018 & 0.0000 & 246/432 & 347/432 \\
\midrule
TTC4TS + Validation-select & 0.940 & 0.993 & 0.0065 & 334/432 & 364/432 \\
TTC4TS + Uniform & 0.965 & 0.997 & 0.0012 & \uline{406/432} & 377/432 \\
TTC4TS + Risk regressor & \uline{0.961} & \uline{0.996} & 0.0029 & 407/432 & 412/432 \\
TTC4TS + Learned weights & \textbf{0.936} & \textbf{0.988} & 0.0023 & \textbf{420/432} & 0/432 \\
\bottomrule
\end{tabular}
\end{table}

We conducted a baseline comparison and component ablation study to evaluate whether TTC4TS aggregation improves forecasting accuracy. Table \ref{tab:baseline_summary} (see Appendix \ref{app_baseline_compar} for details) shows that TTC4TS with fixed learned weights achieves the best aggregate performance across 432 evaluation cells, with geometric-mean relative errors, reductions of 6.4\% compared with the frozen backbone. 
It improves MSE in 420/432 cells (97.2\%), the highest win count among all methods. Compared with the strongest adaptation/retrieval baselines, it achieves lower MSE than each baseline in 347-428 cells.
All four TTC4TS variants improve both aggregate metrics. These results demonstrate broad empirical gains across the evaluated settings and highlight the additional benefit of learned weighting.

\begin{figure}[H]
\centering\includegraphics[width=\linewidth]{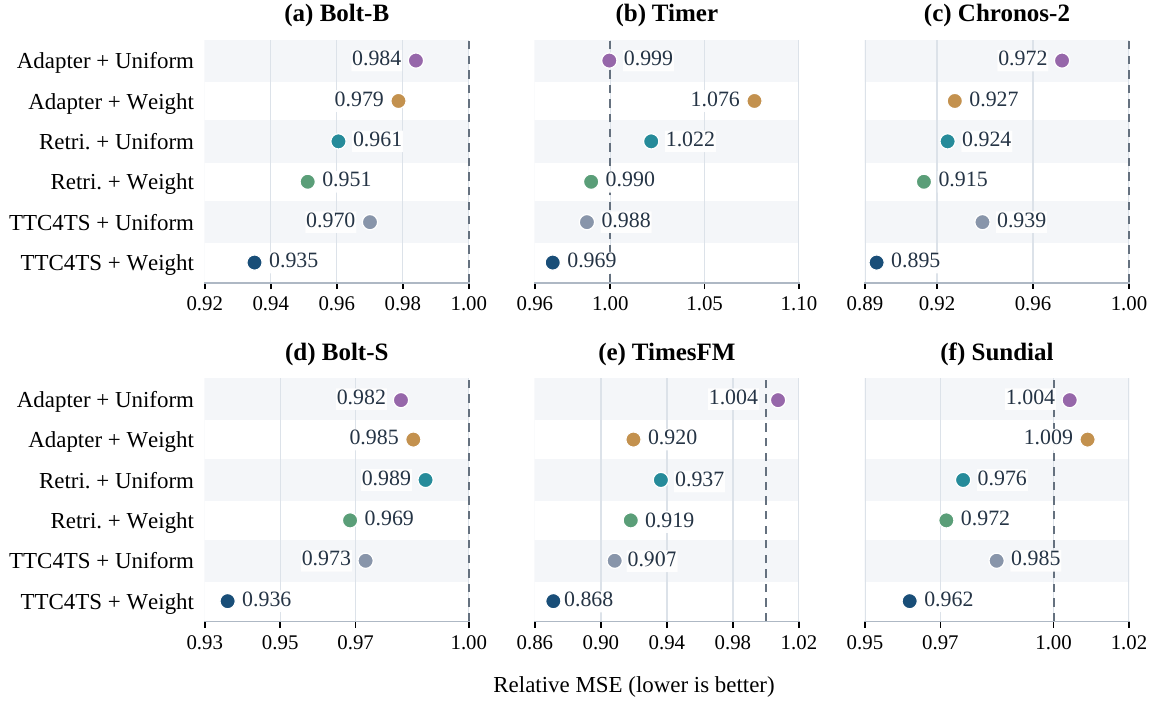}
\caption{Ablation studies of TTC4TS. Each subplot aggregates all 6 datasets, 4 horizons, and 3 seeds for one frozen checkpoint, using geometric MSE ratios to Frozen. The dashed line is the frozen reference, and horizontal axes are truncated and their ranges vary to keep values legible.}\label{fig:baseline_models}
\end{figure}

Figure \ref{fig:baseline_models} shows that TTC4TS with learned weights achieves the lowest relative error across all 6 models, yielding reductions of 3.1-13.2\% over the frozen backbones. It consistently outperforms standalone adapters, retrieval-based variants, and uniform aggregation. Within TTC4TS, weighting provides the largest improvement on Chronos-2. These results support the effectiveness of combining TTC4TS with weighting across the evaluated models.

\subsection{Effectiveness of the Generator}
\label{sec_exp_gen}

\begin{wrapfigure}{r}{0.5\textwidth}
  \vspace{-2em}
  \centering
  \centering\includegraphics[width=.5\textwidth]{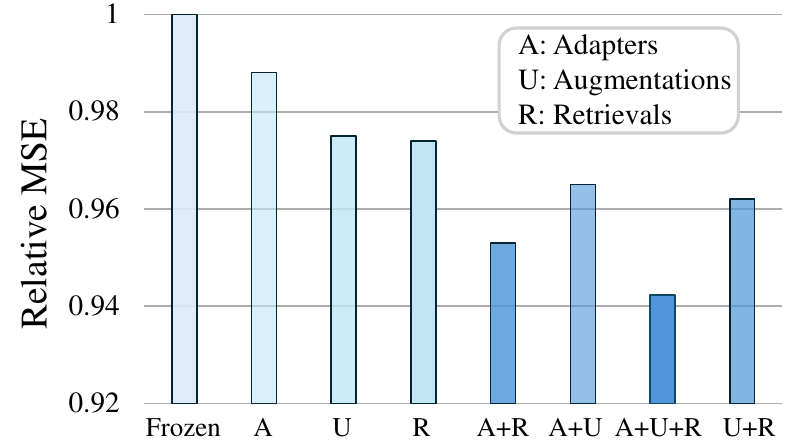}
  \vspace{-1.6em} 
  \caption{Generator-family comparison across 6 datasets. Results are averaged and normalized.}
  \label{fig:families}
  \vspace{-0.6em} 
\end{wrapfigure}


We conduct a generator ablation study to evaluate the individual contributions and complementarity of Adapters (A), Augmentations (U), and Retrieval (R). By comparing these strategies individually and in combination under fixed Uniform aggregation, we examine how candidate-pool composition affects forecasting accuracy relative to the frozen backbone. The results across 6 datasets and 6 backbones are shown in Figure~\ref{fig:families}, and the detailed results are in Appendix Table~\ref{tab:figure2_datasets}. All 3 generation strategies reduce MSE relative to the frozen backbone. Specifically, all pairwise combinations outperform their individual components, with U+R achieving the lowest relative MSE of 0.953 among these combinations. Combining all 3 strategies yields the best result, reducing relative MSE to 0.942, a 5.8\% improvement over the frozen backbone. These results suggest that the 3 generation strategies provide complementary candidates that improve forecasting accuracy when combined.
In addition, every generator is evaluated separately in Appendix Table~\ref{tab:candidates}, including poor stress candidates. 

\subsection{Effectiveness of the Synthesizer}
\label{sec_exp_syn}

\begin{table}[H]
\centering\small
\caption{Geometric mean MSE relative to the frozen backbone on the six datasets over four horizons and three seeds. The experimental results are averaged and normalized for a better comparison.}\label{tab:main_results}
\setlength{\tabcolsep}{4pt}
\resizebox{\linewidth}{!}{\begin{tabular}{lrrrrrrrrrrr}
\toprule
Backbone & Frozen & Uniform & Weights & Class. & Risk & Rank & Ensem. & Retro. & Single & V.Select & Oracle \\
\midrule
Bolt-S & 1.000 & 0.973 & \uline{0.936} & 1.286 & 0.977 & 0.992 & \textbf{0.927} & 0.980 & 0.988 & 0.957 & 0.736 \\
Bolt-B & 1.000 & 0.970 & \uline{0.935} & 1.149 & 0.974 & 0.979 & \textbf{0.913} & 0.977 & 0.979 & 0.954 & 0.730 \\
TimesFM & 1.000 & 0.907 & \textbf{0.868} & 1.163 & 0.904 & 0.913 & \textbf{0.868} & 0.945 & 0.930 & \uline{0.876} & 0.659 \\
Timer & 1.000 & 0.988 & 0.969 & 1.192 & 0.987 & 1.019 & \uline{0.975} & 0.986 & 1.012 & 0.980 & 0.804 \\
Sundial & 1.000 & 0.985 & \textbf{0.962} & 1.290 & 0.985 & 1.027 & \textbf{0.962} & 0.985 & 1.001 & \uline{0.974} & 0.794 \\
Chronos-2 & 1.000 & 0.939 & \uline{0.895} & 1.300 & 0.939 & 0.972 & \textbf{0.876} & 0.948 & 0.937 & 0.907 & 0.721 \\
\bottomrule
\end{tabular}}
\end{table}

We conducted a comparative evaluation of aggregation strategies across 6 frozen forecasting backbones and 6 datasets to assess how candidate selection and combination affect forecasting accuracy and whether these strategies consistently improve upon the frozen backbone. The results are presented in Table ~\ref{tab:main_results}. 
Relative errors are computed against the identity forecast in each task before taking a geometric mean, giving equal weight to datasets, horizons, and seeds. 
The domain-clustered sensitivity is reported in Appendix Table~\ref{tab:domain_summary}. 
It is noted that the test-label Oracle provides a non-deployable reference for the remaining scope to improve candidate selection.

Table ~\ref{tab:main_results} shows that ensemble learning (Ensem.) and learned weights (Weights) achieve the strongest non-oracle performance, with results aggregated over six datasets, four forecasting horizons, and three seeds. Ensem. reduces MSE by 2.5-13.2\% relative to Frozen across the six backbones and achieves the best or joint-best non-oracle result on five, with the largest gains on TimesFM (0.868) and Chronos-2 (0.876). Weights performs best on Timer (0.969) and matches Ensem. on TimesFM and Sundial. Uniform, Risk, Retro., and V.Select also improve over Frozen on every backbone, whereas Class. consistently increases MSE, and Rank and Single occasionally degrade performance. 
Despite these improvements, a clear gap remains to Oracle. This indicates a very promising direction: in the future, we can achieve better results with improved synthesizers.

\begin{wraptable}{r}{0.6\textwidth}
    \centering
    \caption{Aggregation over the same 80‑candidate pool across 6 datasets. Seed SD is the standard deviation of seed-level geometric mean relative MSE. Wins count model--dataset--horizon--seed cells. Oracle (targets) cannot be deployed.}
    \vspace{-0.5em}
    \label{tab:ablation_results}
     \resizebox{0.61\textwidth}{!}
     {
        \setlength{\tabcolsep}{4pt}
        \begin{tabular}{lrrrr}
        \toprule
        Aggregation & Rel. MSE & Rel. MAE & Seed SD & Wins \\
        \midrule
        Frozen backbone & 1.000 & 1.000 & 0.0000 & 0/432 \\
        Uniform & 0.965 & 0.997 & 0.0012 & 406/432 \\
        Learned weights & \uline{0.936} & \uline{0.988} & 0.0023 & \textbf{420/432} \\
        Classifier & 1.228 & 1.156 & 0.0186 & 73/432 \\
        Risk regressor & 0.961 & 0.996 & 0.0029 & \uline{407/432} \\
        Pairwise ranker & 0.983 & 1.026 & 0.0086 & 213/432 \\
        Ensemble learning & \textbf{0.919} & \textbf{0.985} & 0.0015 & 384/432 \\
        Retrospective & 0.970 & 0.996 & 0.0011 & 375/432 \\
        Validation-best candidate & 0.974 & 1.005 & 0.0006 & 240/432 \\
        Validation-Selection & 0.940 & 0.993 & 0.0065 & 334/432 \\
        Oracle (unavailable) & 0.739 & 0.905 & 0.0003 & 432/432 \\
        \bottomrule
        \end{tabular}
     }
     \vspace{-0.8em}
\end{wraptable}


Table~\ref{tab:ablation_results} shows that Ensemble learning has the lowest aggregate MSE among deployable rules (0.919), followed by learned weights. Validation-based Selection does not outperform either fixed rule in aggregate. Selecting the best validation rule is therefore useful relative to Frozen but is not a reliable way to beat the strongest combination baseline. The trajectory Classifier increases relative MSE to 1.228. Pairwise ranking lowers MSE to 0.983 while increasing MAE to 1.026, so its effect depends on the metric. The Oracle uses current test labels and is unavailable at deployment. 
Its relative MSE of 0.739 diagnoses candidate opportunity rather than achievable selection performance.

In addition, ensemble learning achieves the lowest errors, whereas Learned weights improves over the backbone in more configurations. This distinction reflects the magnitude and frequency of improvements: win counts ignore their size, while relative MSE accounts for both gains and degradations.

\begin{figure}[H]
\centering\includegraphics[width=\linewidth]{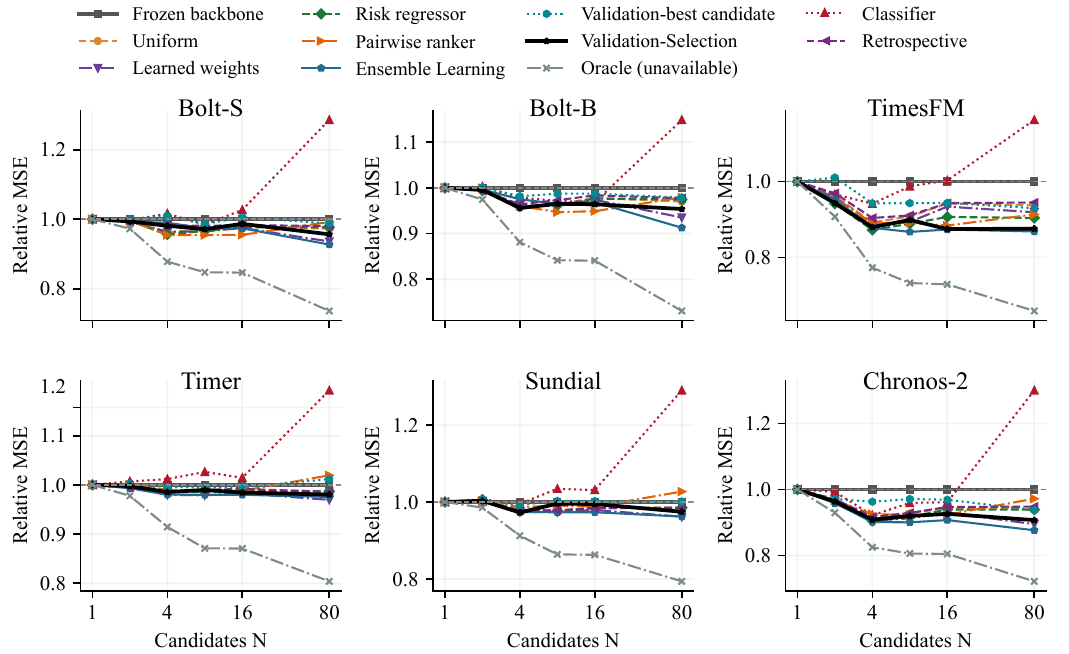}
\caption{Candidate-budget curves on 6 datasets for all eleven rules in the legend. Values are geometric mean relative MSE over four horizons and three seeds. Ordering is fixed before evaluation, and Oracle uses unavailable test labels. Candidate count differs from wall-clock cost.}\label{fig:scaling}
\end{figure}

Further, we conducted a candidate-budget scaling experiment to investigate whether increasing the number of candidate forecasts improves prediction accuracy and how this relationship depends on the aggregation strategy. Specifically, we varied the candidate budget from 1 to 80 across 6 frozen forecasting backbones and compared different aggregation methods. The results shown in Figure \ref{fig:scaling} shows that the benefits of increasing the candidate budget depend strongly on the aggregation strategy. At $N=80$, Learned weights and Ensemble learning improve over the frozen backbone across all 6 models, whereas the Classifier consistently deteriorates. 
In addition, the Oracle error continues to decrease as the pool expands, revealing useful candidates that practical synthesizers do not fully exploit. These results highlight that effective aggregation is essential for converting additional candidates into forecasting gains.

\begin{table}[H]
\centering
\caption{Comparison of methods for quantifying uncertainty. C80/C90/C95 are empirical pointwise coverage percentages, and W90 is mean 90\% interval width in fitting-standardized units. R-WIS is the geometric mean WIS ratio to Frozen split CP over the three interval levels; lower is better. Coverage and width are arithmetic means over cells.}\label{tab:uq_summary}
\fontsize{9.6}{10.6}\selectfont
\setlength{\tabcolsep}{4pt}
\begin{tabular}{lrrrrr}
\toprule
Method & C80 & C90 & C95 & W90 & R-WIS \\
\midrule
Frozen Gaussian \citep{hyndman2018forecasting} & 84.9 & 90.5 & 93.5 & 2.930 & 1.155 \\
Frozen split CP \citep{lei2018distribution} & 81.6 & 89.9 & \uline{94.9} & 2.168 & 1.000 \\
Raw candidate quantiles \citep{koenker2001quantile} & 25.6 & 31.4 & 38.8 & 0.314 & 1.596 \\
History-scale CP \citep{romano2019conformalized} & 81.0 & 89.7 & \uline{94.9} & 2.254 & 1.022 \\
CQR (non-shrinking) \citep{romano2019conformalized} & 83.8 & 93.1 & 97.0 & 2.453 & 1.029 \\
Candidate-spread CP \citep{xu2023sequential} & \uline{80.9} & \textbf{90.0} & 95.2 & 2.539 & 1.043 \\
\midrule
TTC4TS split CP  & 81.3 & \uline{89.8} & \uline{94.9} & 2.126 & 0.985 \\
TTC4TS-quantile CP & 82.2 & 90.7 & 95.7 & 2.116 & \textbf{0.971} \\
TTC4TS-scale CP & \uline{80.8} & 89.8 & \textbf{95.0} & 2.276 & 1.018 \\
TTC4TS bounded ACI & \textbf{80.3} & 90.2 & 95.4 & 2.143 & \uline{0.974} \\
\bottomrule
\end{tabular}
\vspace{0.3em}

{\footnotesize\emph{Note.} Coverage ranks minimize deviation from nominal, and WIS ranks minimize error within each block.\par}
\end{table}

\subsection{Candidate-informed UQ}
\label{sec_exp_uq}

We investigate whether TTC4TS candidates can support informative prediction intervals in addition to accurate point forecasts. The uncertainty experiment covers all 6 backbones, all 6 datasets, 3 seeds, totally 108 complete cells. 
Comparators include Gaussian intervals, split Conformal Prediction (CP), a non-shrinking CQR implementation \citep{romano2019conformalized}, and bounded adaptive conformal inference \citep{gibbs2021adaptive}. Protocol details appear in Appendix~\ref{app:uncertainty}.

Table \ref{tab:uq_summary} summarizes the uncertainty quantification performance across 108 experimental settings. 
It shows that TTC4TS-quantile CP achieves the lowest R-WIS (0.971), corresponding to a 2.9\% improvement over Frozen split CP, while reducing W90 from 2.168 to 2.116 (2.4\%). Its empirical coverage is slightly conservative at 82.2\%, 90.7\%, and 95.7\%. TTC4TS bounded ACI achieves a similarly low R-WIS with coverage closer to the nominal targets (80.3\%, 90.2\%, and 95.4\%). TTC4TS-scale CP attains 95.0\% coverage at the 95\% level, but produces wider intervals (W90 = 2.276) and a higher R-WIS (1.018) than Frozen split CP. In contrast, raw candidate quantiles produce very narrow intervals but exhibit severe undercoverage, with C90 reaching only 31.4\%. Overall, the quantile and bounded ACI variants offer favorable balances between coverage calibration and interval sharpness, although the improvements do not extend to every TTC4TS variant.

It is worth noting that calibration is essential: uncalibrated candidate quantiles cover only 31.4\% of targets at nominal 90\%. Moreover, the learned TTC4TS-scale CP head has relative WIS 1.018, showing that a more complex uncertainty estimator does not necessarily improve interval quality. Overall, the results demonstrate the utility of conformally calibrated candidate quantiles, while distinguishing empirical coverage from guarantees under temporal dependence.

\section{Conclusion}

In this paper, we proposed TTC4TS, a generate-and-synthesize framework for improving forecasts from a fixed backbone. Through augmentation-, adapter-, and retrieval-based candidates, it makes the inference procedure an additional design choice without requiring backbone retraining or test-time parameter updates. Across the 432 main evaluation cells, learned-weight synthesis achieves a 6.4\% reduction in geometric-mean relative MSE and improves on the frozen backbone in 420 cells.
The central finding is conditional rather than monotonic: additional computation helps when candidates provide useful alternatives and synthesis recovers their benefit. Our analysis and candidate-budget experiments distinguish this benefit from redundant sampling and unreliable selection. Separately calibrated candidate quantiles further improve aggregate interval quality, while raw candidate disagreement does not constitute calibrated uncertainty.
Future work should focus on cost-aware candidate generation, more reliable synthesis, and calibration under dependence. More broadly, TTC4TS shifts attention from scaling the forecaster alone to determining which additional computations are worth performing before issuing a forecast.


\clearpage


\bibliography{iclr2027_conference}
\bibliographystyle{iclr2027_conference}

\clearpage
\appendix

\section{Related Work}
\label{sec_rework}

\subsection{Deep Learning for Time Series Forecasting}
Time series (TS) forecasting is a fundamental problem characterized by a rich trajectory of methodological advancements. While classical statistical methods (e.g., ARIMA \citep{piccolo1990distance} and Exponential Smoothing \citep{gardner1985exponential}) and traditional machine learning algorithms (e.g., Random Forests \citep{liaw2002classification} and SVR \citep{castro2009online}) offer strong theoretical guarantees and interpretability, they often struggle to capture the complex, high-dimensional dependencies inherent in real-world data. Consequently, deep learning paradigms have dominated recent literature. Early deep models leveraged CNNs \citep{lecun1998gradient, lai2018modeling} and RNNs \citep{connor1994recurrent, salinas2020deepar} to capture non-linear spatiotemporal dynamics. Subsequently, Transformers \citep{NIPS2017_Transformer} revolutionized the field, with variants introducing mechanisms like sparse attention \citep{Zhou2021Informer, li2019enhancing}, frequency-domain analysis \citep{zhou2022fedformer, yi2024frequency, yi2024filternet}, and period-based formulations \citep{wu2022timesnet, liang2024periodformer}.

Recently, the literature has diversified into several advanced paradigms: (1) \textit{Smarter Transformers} that optimize context length and computational efficiency, such as PatchTST \citep{nie2022time}, iTransformer \citep{liu2023itransformer}, and TimeXer \citep{wang2024timexer}; (2) \textit{Linear and State-Space Models} that rival Transformers at a fraction of the computational cost, including DLinear \citep{zeng2023transformers}, N-HiTS \citep{challu2023nhits}, and Mamba-based architectures \citep{ma2024mamba, ahamed2024timemachine, zhang2025fldmamba}; and (3) \textit{Foundation Models} (e.g., TimeGPT \citep{garza2023timegpt}, TimeLLM \citep{jin2024timellm}, UniTime \citep{liu2024unitime}), which leverage massive cross-domain corpora to enable zero-shot capabilities. Furthermore, some models explicitly integrate physical laws \citep{XiaoHong2023Integrated, XiaoHong2026Integrating, XiaoHong2022Personalized} or utilize decomposition architectures \citep{wu2021autoformer, liu2022non, liang2024minusformer, liang2025distpred} to disentangle complex patterns. Despite these architectural innovations, most of these static models implicitly assume stationarity, leaving them highly vulnerable to concept drift when deployed in dynamic, real-world environments.

\subsection{Adapting to Concept Drift: Online Learning and Test-Time Adaptation}
To circumvent the performance degradation caused by non-stationarity \citep{1976_timeseries, liang2024act}, researchers have increasingly explored continuous adaptation strategies. 

Online learning approaches integrate drift-awareness directly into the forecasting pipeline. For instance, FSNet \citep{pham2023learning} balances a slow-learning backbone with fast-adapting components inspired by complementary learning systems, whereas OneNet \citep{wen2023onenet} dynamically ensembles models specialized in temporal and cross-series dependencies, significantly improving online forecasting robustness \citep{liang2026deepboots}.

Parallel to online learning, Test-Time Adaptation (TTA) aims to adjust pre-trained models during inference. Prominent adapter-based methods include SOLID \citep{Chen2024SOLID}, which retrains specific layers using retrieved similar samples, and TAFAS \citep{Kim2025TAFAS}, which updates linear adapters based on detected temporal cycles. Extensions like PETSA \citep{medeiros2025PETSA}, DynaTTA \citep{Grover2025DynaTTA}, and ELF \citep{Lee2025ELF} further refine this via dynamic gating and auxiliary losses. However, these methods typically rely on linear adapters, parallel fusion, or computationally expensive fine-tuning of internal weights. More critically, they predominantly overlook the insidious problem of label leakage \citep{liang2024act, lau2025dsof}. In contrast, our proposed TTC4TS fundamentally departs from these paradigms by executing non-linear adaptations exclusively at the input and output boundaries. It achieves rapid updates utilizing only the most recent samples, serves dual functions as a feature selector and corrector, and provides robust theoretical guarantees against label leakage.

\subsection{Parameter-Efficient Fine-Tuning and Uncertainty Calibration}
Our approach draws conceptual inspiration from Parameter-Efficient Fine-Tuning (PEFT) methods originally pioneered in Natural Language Processing (NLP). Techniques such as bottleneck adapters \citep{houlsby2019param, pfeiffer2020mad}, Low-Rank Adaptation (LoRA) \citep{hu2022lora}. While $\delta$-Adapter shares the high-level philosophy of freezing the backbone, it is uniquely tailored to the continuous, multi-horizon nature of TS forecasting. Unlike NLP adapters that require white-box access to modify internal attention or feed-forward layers, $\delta$-Adapter operates strictly at the input or output interface, allowing it to wrap around any black-box forecaster. 

Furthermore, handling output uncertainty is critical in TS adaptation. Conformal prediction \citep{vovk2017nonparametric, romano2019conformalized} has emerged as a popular non-parametric framework to construct valid prediction intervals \citep{xu2021conformal, xu2023sequential,Liang2026adapter}. However, conformal prediction heavily relies on the assumption of exchangeable residuals, which is routinely violated in TS data due to strong temporal autocorrelations and non-i.i.d. dynamics.
In contrast, TTC4TS keeps the forecasting backbone and all fitted auxiliary parameters fixed during testing. It generates alternative forecasts through transformations, bounded residual adapters, and historical retrieval, then selects or aggregates these candidates using information available at the forecast origin.

\section{Generator Operator Definitions}
\label{app:most:operators}

This appendix specifies the transformations used in the candidate generator. We omit the forecast-origin subscript and use zero-based indices $t=0,\ldots,L-1$ and $h=0,\ldots,H-1$. All quantities operate on the series standardized using the training prefix. Let
\begin{equation}
\mu_c=\frac{1}{L}\sum_{t=0}^{L-1}X_{t,c},
\qquad
s_c=\max\!\left(
\sqrt{\frac{1}{L}\sum_{t=0}^{L-1}(X_{t,c}-\mu_c)^2}, \delta
\right).
\end{equation}
For transformation $m$, write $B^{(m)}=F(T_m(X))$ for the uncorrected backbone forecast and $Z^{(m)}$ for the resulting candidate.

\subsection{Augmentation-based Candidates}
\label{app_gen_aug}

\textbf{Magnitude Flip}:
Reflect the history around its channel mean and apply the same reflection to the forecast:
\begin{equation}
T_{\mathrm{mf}}(X)_{t,c}=2\mu_c-X_{t,c},
\qquad
Z^{(\mathrm{mf})}_{h,c}=2\mu_c-B^{(\mathrm{mf})}_{h,c}.
\end{equation}

\textbf{Time Flip}:
Reverse the observed history without reversing the forecast:
\begin{equation}
T_{\mathrm{tf}}(X)_{t,c}=X_{L-1-t,c},
\qquad
Z^{(\mathrm{tf})}=B^{(\mathrm{tf})}.
\end{equation}
This is a perturbation-based candidate, not an assertion of time-reversal equivariance.

\textbf{Magnitude Warp}:
Define the positive modulation
\begin{equation}
g(u)=\exp\!\left(\epsilon \sin\frac{2\pi u}{L}\right),
\end{equation}
where, $\epsilon$ is a small constant factor.  Then, we use
\begin{equation}
T_{\mathrm{mw}}(X)_{t,c}
=\mu_c+(X_{t,c}-\mu_c)g(t),
\qquad
Z^{(\mathrm{mw})}_{h,c}
=\mu_c+\frac{B^{(\mathrm{mw})}_{h,c}-\mu_c}{g(L+h)}.
\end{equation}

\textbf{Time Warp}:
Distort the sampling locations by
\begin{equation}
u_t=\frac{t}{L-1},
\qquad
\tau_t=(L-1)\left[u_t+0.05\sin(2\pi u_t)\right].
\end{equation}
Let $a_t=\lfloor\tau_t\rfloor$, $b_t=\min(a_t+1,L-1)$, and $\lambda_t=\tau_t-a_t$. Linear interpolation yields
\begin{equation}
T_{\mathrm{tw}}(X)_{t,c}
=(1-\lambda_t)X_{a_t,c}+\lambda_tX_{b_t,c},
\qquad
Z^{(\mathrm{tw})}=B^{(\mathrm{tw})}.
\end{equation}

\textbf{EWMA Smoothing}:
Construct a smoothed history recursively:
\begin{equation}
\widetilde X_{0,c}=X_{0,c},
\qquad
\widetilde X_{t,c}=0.3X_{t,c}+0.7\widetilde X_{t-1,c}\quad(t\geq1),
\qquad
Z^{(\mathrm{ewma})}=F(\widetilde X).
\end{equation}

\textbf{Jitter}:
The jittered history and candidate are
\begin{equation}
\widetilde X_{t,c}
=
X_{t,c}
+\alpha s_c\epsilon_{t,c},
\qquad
\epsilon_{t,c}\sim\mathcal N(0,1),
\end{equation}
where $s_c$ is the history’s standard deviation and $\alpha$ controls the noise strength.
The implementation constructs reproducible pseudo-random noise rather than drawing fresh independent noise on each call. For seed $s$, define
\begin{equation}
\widetilde X^{(j)}
=
X+\alpha_j s\odot\epsilon^{(j)},
\qquad
\widehat Y^{(j)}=F(\widetilde X^{(j)}).
\end{equation}
Specifically, in our paper implementation, the noise is deterministic for a fixed query and seed, and the construction should not be described as supplying independent posterior samples.

\textbf{Horizontal Shift}:
With $d=\max(1,\lfloor L/K\rfloor)$, shift the history using left-endpoint replication:
\begin{equation}
T_{\mathrm{hs}}(X)_{t,c}=X_{\max(0,t-d),c},
\qquad
Z^{(\mathrm{hs})}=B^{(\mathrm{hs})}.
\end{equation}

\textbf{Positive and Negative Vertical Shifts}:
Each sign produces a separate candidate. For $b_c^\pm=\pm0.5s_c$,
\begin{equation}
T_{\mathrm{vs}\pm}(X)_{t,c}=X_{t,c}+b_c^\pm,
\qquad
Z^{(\mathrm{vs}\pm)}_{h,c}=B^{(\mathrm{vs}\pm)}_{h,c}-b_c^\pm.
\end{equation}

\textbf{Positive and Negative Slopes}:
Each sign again gives a separate candidate. For $a_c^\pm=\pm0.5s_c/L$,
\begin{equation}
T_{\mathrm{sl}\pm}(X)_{t,c}
=X_{t,c}+a_c^\pm(t-L+1),
\qquad
Z^{(\mathrm{sl}\pm)}_{h,c}
=B^{(\mathrm{sl}\pm)}_{h,c}-a_c^\pm(h+1).
\end{equation}
The added historical slope is zero at the last observation, and the forecast correction uses the corresponding future offset.

\textbf{Spike Injection}:
Let $t_\star=L-\max(2,\lfloor \frac{L}{2K}\rfloor)$. Inject a channel-scaled spike:
\begin{equation}
T_{\mathrm{sp}}(X)_{t,c}
=X_{t,c}+2s_c\mathbf1[t=t_\star],
\qquad
Z^{(\mathrm{sp})}=B^{(\mathrm{sp})}.
\end{equation}
The ten transformation types above yield twelve candidates because vertical shifts and slopes each have two signs. Only magnitude flip, magnitude warp, vertical shifts, and slope changes apply non-identity forecast corrections.

\subsection{Multi-Bounded Adapters}
\label{app_gen_adapter}

Here, we omit the batch dimension and adopt zero-based indexing for the time dimension. All operations are performed on data standardized using statistics computed from the training prefix. The adapters generate candidate representations by scaling a learned residual. Specifically, given an input tensor $Z \in \mathbb{R}^{T \times C}$, the Adapter computes
\begin{equation}
\begin{aligned}
u&=\operatorname{vec}(Z),\\
h_1&=\mathcal N\!\left(\operatorname{ReLU}(W_1u+b_1)\right),\\
h_2&=\mathcal N\!\left(\operatorname{ReLU}(W_2h_1+b_2)\right),\\
r_\theta(Z)
&=\operatorname{reshape}_{T\times C}
  \left[\tanh(W_3h_2+b_3)\right].
\end{aligned}
\end{equation}

Here, $\mathcal N$ is the normalization used by the implementation. Because of $\tanh$,
\begin{equation}
|r_\theta(Z)_{t,c}|\leq 1.
\end{equation}
For $K\geq2$, the adjustment strengths are
\begin{equation}
\delta_j=\frac{j}{K-1}\delta_{\max},
\qquad j=0,\ldots,K-1.
\end{equation}
Then, the input, output, and joint candidates are respectively

\begin{equation}
    \begin{aligned}
\widehat Y_j^{\mathrm{in}} & = F\!\left(X+\delta_j r_{\theta_x}(X)\right) \\
\widehat Y_j^{\mathrm{out}} & = F(X)+\delta_j r_{\theta_y}(F(X)) \\
\widehat Y_j^{\mathrm{joint}} & = \widehat Y_j^{\mathrm{in}} +\delta_j r_{\theta_y}   \left(\widehat Y_j^{\mathrm{in}}\right).
\end{aligned}
\end{equation}

The same input Adapter $\theta_x$ is shared by the input and joint paths; the same output Adapter $\theta_y$ is shared by the output and joint paths. Their training objective is

\begin{equation}
\min_{\theta_x,\theta_y}
\frac{1}{|\mathcal D_{\mathrm{adapt}}|}
\sum_{(X,Y)\in\mathcal D_{\mathrm{adapt}}}
\frac{1}{3K}
\sum_{j=0}^{K-1}
\sum_{v\in\{\mathrm{in,out,joint}\}}
\frac{\|\widehat Y_j^v-Y\|_F^2}{HC},
\end{equation}

with all parameters of $F$ fixed. Backpropagation through $F$ remains enabled when training the input Adapter.
Further, the code also supports the optional multiplicative correction

\begin{equation}
A_{\theta,\delta}^{\mathrm{mul}}(Z)
=
Z\odot\left(1+\delta r_\theta(Z)\right).
\end{equation}
Since $|r_\theta(Z)_{t,c}|\leq1$, the multiplicative factor lies in $[1-\delta,1+\delta]$. This bounds the relative factor rather than the absolute size of the correction independently of $Z$.

\subsection{Historical Retrieval and Alignment}
\label{app_gen_retrieval}

References are complete historical pairs $R_n=V[16n:16n+S+H]$, with $S=64$, and must lie entirely inside the training bank. At origin $o$, only entries satisfying $16n+S+H\leq o$ are eligible. The query is $Q=X[L-S:L]$. For each channel, retrieve the $M$ smallest local-z-score distances between $Q$ and the reference past, using the standard-deviation floor $0.01$. Candidate $j$ assembles channel $c$ from its $j$th nearest reference, yielding $R^{(j)}_{t,c}=R_{n_{j,c},t,c}$. This is channel-wise retrieval, not retrieval of one common multivariate window for all channels.

\textbf{Phase Alignment}:
Center the query and each reference past:
\begin{equation}
q^0_{t,c}=Q_{t,c}-\overline Q_c,
\qquad
p^0_{j,t,c}=R^{(j)}_{t,c}
-\frac{1}{S}\sum_{u=0}^{S-1}R^{(j)}_{u,c}.
\end{equation}
For lag $\ell$, define the valid overlap
\begin{equation}
I_\ell=\{t:0\leq t<S,\ 0\leq t-\ell<S\}.
\end{equation}
The selected lag is
\begin{equation}
\ell^\star_{j,c}
=\arg\max_{\ell\in\{-8,\ldots,8\}}
\frac{
\sum_{t\in I_\ell}q^0_{t,c}p^0_{j,t-\ell,c}
}{
\max\!\left(
\sqrt{
\sum_{t\in I_\ell}(q^0_{t,c})^2
\sum_{t\in I_\ell}(p^0_{j,t-\ell,c})^2
},10^{-6}\right)
}.
\end{equation}
Apply that lag to the complete historical pair, replicating endpoints:
\begin{equation}
R^{\mathrm{phase},j}_{t,c}
=R^{(j)}_{\operatorname{clip}(t-\ell^\star_{j,c},0,S+H-1),c}.
\end{equation}

\textbf{Amplitude Alignment}:
Compute $\mu_{r,j,c}$ and $\sigma_{r,j,c}$ from the first $S$ positions of $R^{\mathrm{phase},j}$, and $\mu_{q,c},\sigma_{q,c}$ from $Q$. Then
\begin{equation}
\alpha_{j,c}
=\frac{\sigma_{q,c}\sigma_{r,j,c}}
{\sigma_{r,j,c}^{2}+(\sigma_{q,c}/5)^2+10^{-6}},
\qquad
\beta_{j,c}=\mu_{q,c}-\alpha_{j,c}\mu_{r,j,c},
\end{equation}
\begin{equation}
\widetilde R^{(j)}_{t,c}
=\alpha_{j,c}R^{\mathrm{phase},j}_{t,c}+\beta_{j,c}.
\end{equation}
The denominator is the implemented stabilized alignment rule; it is not replaced by ordinary variance-ratio rescaling. One forecast is generated from each concatenation $[\widetilde R^{(j)},X]$. Historical neighbors are not averaged before forecasting.

\textbf{Causal and Interface Requirements}:
The full retrieval configuration requires $L\geq64$, at least $M$ eligible entries per channel, and a backbone that accepts $L+64+H$ input steps. At retrospective origin $o-H$, causal replay requires the eligibility bound $16n+64+H\leq o-H$. This requirement concerns the historical replay call, not just current-origin retrieval.

\section{Synthesizer for Trajectory-level synthesis}
\label{app_syn}

\subsection{Static Aggregation}
\textbf{Uniform Averaging}: It sets $w_{k}=1/N$ and requires no synthesizer fitting. Learned weights introduce one logit $a_k$ per candidate and solve
\begin{equation}
w_k=\operatorname{softmax}(a)_k,
\qquad
\min_a\frac{1}{|\mathcal D_{o,\mathrm{syn}}|}
\sum_{o\in\mathcal D_{o,\mathrm{syn}}}
\ell\!\left(\sum_kw_kZ_{o,k},Y_o\right).
\label{eq:most:static-weights}
\end{equation}
Here and below, $o\in\mathcal D_{o,\mathrm{syn}}$ indexes forecast windows in a fitting block. These weights are fixed across test inputs and are fitted separately for each model--dataset--horizon--seed configuration.

\subsection{Learning-based Synthesizer}
\label{app_syn_define}

\textbf{Ensemble learning}: The Ensemble learning instead performs $40$ selections with replacement on $\mathcal D_{\mathrm{syn}}$. If $E_{n,o}$ averages the first $n$ selections, the next candidate is
\begin{equation}
j_{n+1}=\arg\min_j
\frac{1}{|\mathcal D_{o,\mathrm{syn}}|}
\sum_{o\in\mathcal D_{o,\mathrm{syn}}}
\ell\!\left(\frac{nE_{n,o}+Z_{o,j}}{n+1},Y_o\right),
\label{eq:most:greedy}
\end{equation}
with the initial selection minimizing single-candidate fitting loss. A candidate selected $c_k$ times receives $w_k=c_k/40$. These coefficients also remain fixed at test time.

\textbf{Learned Weights}: Learn one logit $a_k$ per candidate and convert the logits into convex weights:
\begin{equation}
    w_k=\operatorname{softmax}(a)_k.
\end{equation}
Optimize the combined forecast directly on the aggregator-training block:
\begin{equation}
    \mathcal L_{\mathrm{weights}}    =\frac1N\sum_t    \operatorname{MSE}\!\left(\sum_kw_kZ_{t,k},Y_t\right).
\end{equation}


\textbf{Classifier}: The classifier attempts to predict the best candidate. For each fitting window, let $k^\star=\arg\min_k\ell_{k}$. The classifier minimizes
\begin{equation}
\mathcal L_{\mathrm{class}}
=-\frac{1}{|\mathcal D_{\mathrm{syn}}|}
\sum_{o\in\mathcal D_{\mathrm{syn}}}
\log\operatorname{softmax}(s)_{k^\star}.
\end{equation}
At inference it returns $Z_{\arg\max_k s_{k}}$. Thus, classification implements multiclass hard selection rather than probability-weighted averaging.

\textbf{Risk Regressor}: The risk regressor aggregates candidate-level features to predict relative error, with the regression target defined as a stabilized transformation of each candidate’s error relative to that of the backbone:
\begin{equation}
q_{k}=\log\!\left[
1+\min\!\left(100,\frac{\ell_{k}}{\max(\ell_{0},10^{-5})}\right)
\right].
\label{eq:most:risk-target}
\end{equation}
The regression objective applies SmoothL1 loss to the score and target arrays over fitting windows and candidates:
\begin{equation}
\mathcal L_{\mathrm{risk}}
=\operatorname{SmoothL1}\!\left(
\{s_{k}\}_{o\in\mathcal D_{\mathrm{syn}},k},
\{q_{k}\}_{o\in\mathcal D_{\mathrm{syn}},k}\right).
\end{equation}
At inference,
\begin{equation}
w_{k}=\operatorname{softmax}(-s)_k.
\label{eq:most:risk-weights}
\end{equation}
Lower predicted transformed relative error therefore receives higher weight. The risk regressor softly combines trajectories; it neither predicts raw MSE nor selects only the minimum-score candidate. Adapter candidate-count experiments use this synthesis rule.

\textbf{Pairwise Ranker}: The pairwise ranker learns to assign higher scores to lower-error candidates through pairwise comparisons of training errors, then selects the highest-scoring forecast trajectory at inference.
For unequal-error pairs, define $d_{ij}=\operatorname{sign}(\ell_{i}-\ell_{j})$ and optimize
\begin{equation}
\mathcal L_{\mathrm{rank}}
=\operatorname{mean}_{i,j:\,d_{ij}\ne0}
\operatorname{softplus}\!\left[d_{ij}(s_{i}-s_{j})\right].
\label{eq:most:ranking}
\end{equation}
This objective assigns higher scores to lower-error candidates. Equal-error pairs are excluded. Inference selects $Z_{\arg\max_k s_{k}}$, as in classification, but supervision uses pairwise preferences rather than only the winning index.

\subsection{Retrospective Weighting}
\label{app_syn_retro}

At origin $o$, the immediately preceding horizon is already observed. We replay each candidate operator with context $X[o-H-L:o-H]$ and historical origin $o-H$, obtaining $Z_{o-H,k}$. Its retrospective error is
\begin{equation}
r_{k}=\ell\!\left(Z_{o-H,k},V[o-H:o]\right).
\label{eq:most:retrospective-error}
\end{equation}
To ensure causal replay, retrieval eligibility in this call must be checked against $o-H$, not $o$. Retrospective synthesis then uses
\begin{equation}
w_{k}=(1-\lambda)
\operatorname{softmax}_k\!\left(
-\frac{r_{k}}{\tau\max(r_{0},10^{-5})}
\right)+\lambda\mathbf1[k=0].
\label{eq:most:retrospective-weights}
\end{equation}
Validation selects $\tau\in\{0.1,0.3,1\}$ and $\lambda\in\{0,0.5,1\}$. The temperature controls concentration on recently successful operators, while $\lambda$ shrinks the forecast toward the frozen backbone. This policy changes weights without fitting a synthesizer or updating parameters at test time. It requires at least $L+H$ historical observations and an additional candidate bank unless the relevant forecasts are cached.

\subsection{Validation‑aided Selection}
\label{app_syn_val}

Adapter supervision uses $\mathcal D_{\mathrm{adapt}}$, while learned synthesis rules use the aggregator-training block $\mathcal D_{\mathrm{syn}}$. Validation targets select hyperparameters and deployment rules; current test-future targets are unavailable throughout generation and synthesis. The learned-synthesizer optimization configuration uses $160$ AdamW updates, learning rate $0.01$, and weight decay $0.03$. Offline auxiliary fitting is distinct from backbone retraining and from online parameter adaptation.

\textbf{Validation-best Candidate}:
In addition to the synthesis rules above, validation-best selects a single candidate operator:
\begin{equation}
k^\star=\arg\min_k
\frac{1}{|\mathcal D_{\mathrm{val}}|}
\sum_{o\in\mathcal D_{\mathrm{val}}}\ell_{k},
\qquad \widehat Y=Z_{k^\star}.
\label{eq:most:validation-candidate}
\end{equation}
The operator is fixed throughout testing, but its trajectory is regenerated for each input. This is not per-query oracle selection.

\textbf{Validation-aided Selection}: 
The validation-based selection chooses an entire deployment rule from
\begin{equation}
\begin{aligned}
\mathcal M=\{&\text{Frozen},\text{Uniform},\text{Learned weights},
\text{Classifier},\text{Risk},\\
&\text{Ranker},\text{Greedy},\text{Retrospective},
\text{Validation-best}\},
\end{aligned}
\end{equation}
using
\begin{equation}
m^\star=\arg\min_{m\in\mathcal M}
\frac{1}{|\mathcal D_{\mathrm{val}}|}
\sum_{o\in\mathcal D_{\mathrm{val}}}
\ell(\widehat Y^{(m)},Y).
\label{eq:most:validation-method}
\end{equation}
The selected rule is held fixed across the test split, although its candidate weights may remain input-dependent. This outer selection is a deployment policy, not an additional learned aggregator or a per-query selector between synthesis methods. It may choose the unchanged backbone.

\section{When Additional Computation is Beneficial}\label{app:proof}
This appendix develops Section~\ref{when_ttc_help} in three stages: the geometry of the candidate set, the error incurred when learning or scoring its combination, and the cost of obtaining that set. All claims concern squared error in the fitting-standardized coordinates. They do not assert corresponding guarantees for MAE, geometric averages of relative errors, or probabilistic coverage. The arguments specialize classical convex projection, forecast combination, and exponential weighting \citep{winkler1983combination,cesa2006prediction}; their purpose is to make the conditions relevant to TTC4TS explicit.

\subsection{Information, randomness, and the risk being compared}\label{app:theory_setup}
Flatten an $H\times C$ forecast to $\mathbb R^d$, $d=HC$, with inner product $\langle a,b\rangle_d=a^\top b/d$ and norm $\|a\|_d$. Let $\mathcal H_t$ contain all observed information used at the forecast origin: history, fitted parameters, normalization, the causal reference bank, and the realized algorithmic random draws. Such draws use no future information; conditionally on the observed history they are independent of the future target. Each candidate $Z_k$ and deployed weight is $\mathcal H_t$-measurable. Assume finite conditional second moments and a finite nonempty candidate set. Conditional equalities below hold almost surely; suppress $t$ to simplify notation.
Define
\begin{equation}
 \mu=\mathbb E[Y\mid\mathcal H_t],\qquad
 v=\mathbb E[\|Y-\mu\|_d^2\mid\mathcal H_t],\qquad
 R(z)=\mathbb E[\|z-Y\|_d^2\mid\mathcal H_t].
 \label{eq:conditional_setup}
\end{equation}
The simplex is $\Delta_K=\{w\in\mathbb R^K:w_k\geq0,\ \sum_kw_k=1\}$. We write $R(w)=R(z_w)$ for $z_w=\sum_kw_kZ_k$, and $R_k=R(Z_k)$. The conditional mean $\mu$ and conditional risks are analytical quantities, not inputs to the implementation. No claim below permits reading the current future target.

Conditional risk describes one observed history with its remaining future uncertainty. In contrast, static Learned weights minimize a single risk averaged over a deployment distribution of histories and futures. We denote that risk by $\bar R$ and analyze it separately in Appendix~\ref{app:weights_theory}. An oracle allowed to choose different weights for every history is generally stronger than the best static weights. Neither oracle is the test-label selector reported in the experiments.

\subsection{Exact risk geometry and the value of a new candidate}\label{app:geometry}
\textbf{Proposition 1 (Risk and Diversity Identities)}:
Define $Q_{ij}=\mathbb E[\langle Z_i-Y,Z_j-Y\rangle_d\mid\mathcal H_t]$. For every $w\in\Delta_K$,
\begin{align}
 Q_{ij}&=v+\langle Z_i-\mu,Z_j-\mu\rangle_d,\label{eq:q_bias}\\
 R(w)&=w^\top Qw=v+\|z_w-\mu\|_d^2
       =\sum_kw_kR_k-D(w),\label{eq:risk_identity}\\
 D(w)&=\sum_kw_k\|Z_k-z_w\|_d^2
       =\tfrac12\sum_{i,j}w_iw_j\|Z_i-Z_j\|_d^2.\label{eq:diversity_identity}
\end{align}
In particular, $Q$ is positive semidefinite and $D(w)\geq0$.

\textbf{Proof}:
Write $Z_i-Y=(Z_i-\mu)-(Y-\mu)$. The cross terms have conditional mean zero, proving Equation~\ref{eq:q_bias}. Since $\sum_kw_k=1$, the mixture error is $\sum_kw_k(Z_k-Y)$, whose expected squared norm is $w^\top Qw$. Expanding $\|z_w-Y\|_d^2$ around $\mu$ gives the second equality in Equation~\ref{eq:risk_identity}. Also,
\[
 \tfrac12\sum_{i,j}w_iw_j\|Z_i-Z_j\|_d^2
 =\sum_kw_k\|Z_k\|_d^2-\|z_w\|_d^2
 =\sum_kw_k\|Z_k-z_w\|_d^2.
\]
Subtracting this identity from $\sum_kw_kR_k=v+\sum_kw_k\|Z_k-\mu\|_d^2$ proves the remaining equality. Finally, for any vector $a$, $a^\top Qa=\mathbb E[\|\sum_ka_k(Z_k-Y)\|_d^2\mid\mathcal H_t]\geq0$.

\textbf{Interpretation}:
$D(w)$ is observable, but the weighted individual risk in Equation~\ref{eq:risk_identity} is not. Increasing both by adding a bad forecast need not improve their difference. After conditioning on the history and generator draws, candidates are fixed; their errors share the same random future. They are not independent draws of that future. Across random histories, an error second moment can be decomposed into bias products and centered cross moments. Calling the whole matrix an error covariance would omit the bias term.

For example, under the restrictive unconditional assumptions that every error has common bias $b$, centered second moment $\sigma^2$, and pairwise centered cross moment $\rho\sigma^2$, uniform averaging has risk
\begin{equation}
 \|b\|_d^2+\frac{K\sigma^2+K(K-1)\rho\sigma^2}{K^2}
 =\|b\|_d^2+\sigma^2\left(\rho+\frac{1-\rho}{K}\right).
 \label{eq:correlation_floor}
\end{equation}
For a family with fixed $0\leq\rho\leq1$, only the last term decreases with $K$. Bias and correlated error remain. This derivation explains a possible saturation mechanism; the heterogeneous TTC4TS pool is not assumed to satisfy these exchangeable-moment conditions.

\textbf{Proposition 2 (Necessary and Sufficient Improvement of The Convex Oracle)}:
Let $\mathcal C=\operatorname{conv}\{Z_0,\ldots,Z_{K-1}\}$ and let $z^*$ be the projection of $\mu$ onto $\mathcal C$. Then
\begin{equation}
 R^*_{\mathcal C}=\min_{z\in\mathcal C}R(z)
 =v+\operatorname{dist}_d(\mu,\mathcal C)^2.\label{eq:hull_risk}
\end{equation}
For an additional candidate $z$, put $u=z-z^*$, $c=\langle z^*-\mu,u\rangle_d$, and $q=\|u\|_d^2$. The enlarged hull has strictly smaller optimal risk if and only if $c<0$. Its risk reduction is at least
\begin{equation}
 g(c,q)=
 \begin{cases}
 0,&q=0\ \text{or}\ c\geq0,\\
 c^2/q,&q>0,\ -q\leq c<0,\\
 -2c-q,&q>0,\ c<-q.
 \end{cases}\label{eq:segment_gain}
\end{equation}
More generally, nested candidate hulls have non-increasing optimal risk.

\textbf{Proof}:
The finite hull is compact and convex. The squared distance has a minimizer, and strict convexity in the forecast $z$ makes it unique (its weight representation need not be unique). Equation~\ref{eq:risk_identity} proves Equation~\ref{eq:hull_risk}. The projection satisfies
\begin{equation}
 \langle z^*-\mu,x-z^*\rangle_d\geq0
 \quad\text{for every }x\in\mathcal C.\label{eq:projection_condition}
\end{equation}
Indeed, a negative inner product would make a sufficiently short feasible step toward $x$ decrease the squared distance. Conversely, expanding the squared distance shows that this inequality implies optimality.

Every point of the enlarged hull has the form $(1-a)x+az$ with $x\in\mathcal C$ and $a\in[0,1]$. If $c\geq0$, its displacement from $z^*$ has nonnegative inner product with $z^*-\mu$ by Equation~\ref{eq:projection_condition}; hence $z^*$ remains optimal. If $c<0$, the feasible segment $z^*+au$ gives
\begin{equation}
 R(z^*+au)-R(z^*)=2ac+a^2q.\label{eq:directional_quadratic}
\end{equation}
This is negative for sufficiently small positive $a$. For $q>0$, its minimizer on $[0,1]$ is $a^*=\min\{1,\max\{0,-c/q\}\}$. Substituting the three cases proves Equation~\ref{eq:segment_gain}. If $q=0$, then $u=0$ and $c=0$. The full enlarged hull may improve more than this single segment, which is why $g$ is a lower bound on its gain. Finally, minimizing over a superset cannot increase the minimum.

\subsubsection{Why Uniform Can Get Worse}
Let $\bar z$ be the old mean of $K$ candidates and add $z$. Set $u=z-\bar z$, $c=\langle\bar z-\mu,u\rangle_d$, and $q=\|u\|_d^2$; these quantities now use $\bar z$, not $z^*$. The new mean is $\bar z+u/(K+1)$. Equation~\ref{eq:directional_quadratic} gives strict improvement exactly when
\begin{equation}
 2(K+1)c+q<0.\label{eq:uniform_condition}
\end{equation}
A direction can be helpful for a small weight yet harmful at Uniform's forced weight. For a scalar deterministic target $Y=0$, adding $z=M\ne0$ to the perfect frozen forecast $Z_0=0$ raises Uniform's risk to $M^2/4$. Conversely, candidates $1$ and $-1$ each have risk one but their equal mixture has risk zero. Neither trajectory distance nor the best individual risk fully characterizes the best ensemble.

\subsection{Adapter sample count: resolution, curvature, and refitting}\label{app:adapter_theory}
\textbf{Proposition 3 (A Fixed Output Adapter has a One-Dimensional Hull)}:
Fix $Z_0$, a residual $r$, and an interval $[0,m]$, where $m=\delta_{\max}>0$. For any finite grid $\mathcal D\subset[0,m]$ containing both endpoints,
\begin{equation}
 \operatorname{conv}\{Z_0+\delta r:\delta\in\mathcal D\}
 =\{Z_0+\delta r:0\leq\delta\leq m\}.\label{eq:affine_hull}
\end{equation}
Thus its optimal convex risk is independent of grid size. A uniform linspace grid with $N\geq2$ points has mean forecast $Z_0+(m/2)r$ for every $N$. If $h$ is the maximum adjacent grid gap, the best single grid candidate obeys
\begin{equation}
 0\leq\min_{\delta\in\mathcal D}R(Z_0+\delta r)
       -\min_{0\leq\delta\leq m}R(Z_0+\delta r)
 \leq\tfrac14\|r\|_d^2h^2.\label{eq:adapter_grid_bound}
\end{equation}

\textbf{Proof}:
Every convex combination has effective adjustment $\sum_jw_j\delta_j\in[0,m]$. Conversely, each $\delta\in[0,m]$ is a convex combination of $0$ and $m$, proving Equation~\ref{eq:affine_hull}. The arithmetic mean of linspace is $m/2$. For hard selection, set $b=\langle Z_0-\mu,r\rangle_d$ and $q=\|r\|_d^2$. Risk is $R(Z_0)+2b\delta+q\delta^2$. If $q=0$, the claim is immediate. Otherwise the continuous minimizer is $\delta^*=\min\{m,\max\{0,-b/q\}\}$. When it is an endpoint, the grid attains it. When it is interior, the derivative vanishes and
\[
 R(Z_0+\delta r)-R(Z_0+\delta^*r)=q(\delta-\delta^*)^2.
\]
A nearest grid point lies within $h/2$, yielding Equation~\ref{eq:adapter_grid_bound}.

\subsubsection{Scope for The Implemented Adapters}
Here $r=R_\psi(F(X))$ is fixed after fitting and $m=0.3$. The statement concerns the output-only grid with one identity, not the mean of the entire mixed pool. In that pool, family proportions and the identity's mass also matter. Two linspace grids with different sizes need not be nested. For example, with $Z_0=0$, $r=1$, and deterministic $Y=m/2$, a three-point grid attains zero selection error while a four-point grid does not. A smaller upper bound on discretization error is not a monotonicity theorem for non-nested grids. Refitting adapters, as in the sample-count experiment, additionally changes $r$ and the input/composed paths.

\textbf{Proposition 4 (Discretizing a Nonlinear Adapter Path)}:
Let $z:[0,m]\to\mathbb R^d$ be a twice continuously differentiable forecast path with $\sup_\delta\|z''(\delta)\|_d\leq B$. Let $\mathcal C_{\rm path}$ be the convex hull of the full path, $\mathcal C_{\rm grid}$ the hull of an endpoint-inclusive grid with maximum gap $h$, and $a=\operatorname{dist}_d(\mu,\mathcal C_{\rm path})$. With $\epsilon_h=Bh^2/8$,
\begin{equation}
 0\leq R^*_{\mathcal C_{\rm grid}}-R^*_{\mathcal C_{\rm path}}
 \leq2a\epsilon_h+\epsilon_h^2.\label{eq:curved_grid_bound}
\end{equation}
If only $L_z$-Lipschitz continuity is known, the same statement holds with $\epsilon_h=L_zh/2$.

\textbf{Proof}:
For neighboring grid points $a_0,b_0$ with gap $\ell\leq h$ and $\delta=(1-\theta)a_0+\theta b_0$, linearly interpolate $\widetilde z(\delta)=(1-\theta)z(a_0)+\theta z(b_0)$. Taylor expansion about $\delta$ cancels the first-order terms in this weighted sum. The integral remainders have norms bounded by $B|a_0-\delta|^2/2$ and $B|b_0-\delta|^2/2$. Therefore
\[
 \|\widetilde z(\delta)-z(\delta)\|_d
 \leq\tfrac B2\big[(1-\theta)\theta^2+\theta(1-\theta)^2\big]\ell^2
 =\tfrac B2\theta(1-\theta)\ell^2\leq Bh^2/8.
\]
Under Lipschitz continuity the corresponding bound is
$ (1-\theta)L_z\theta\ell+\theta L_z(1-\theta)\ell\leq L_zh/2$.
The continuous path is compact, and its finite-dimensional convex hull is compact: each point admits a combination of at most $d+1$ path points. Replace those points in a representation of the projection onto $\mathcal C_{\rm path}$ by their interpolants. Their combination lies in $\mathcal C_{\rm grid}$ and is within $\epsilon_h$ of that projection. Its distance from $\mu$ is at most $a+\epsilon_h$. Equation~\ref{eq:hull_risk} then gives the upper bound $(a+\epsilon_h)^2-a^2$; hull inclusion gives the lower bound.

Input and composed adapters in Equation~\ref{eq:adapter_grid_bound} can have curvature because adjusted contexts pass through $F$. Their neural wrappers are not assumed globally twice differentiable with a known $B$; even a useful Lipschitz constant is not established here. Proposition 4 is a conditional approximation result, not a certified rate for these checkpoints. It also bounds approach to a continuous-path oracle, not recovery of that oracle by a fitted scorer.

\subsection{Static Learned weights and aggregation estimation error}\label{app:weights_theory}
Condition on the fitted generator. Let $P$ be the deployment distribution of a history--future pair, with finite second moments. For one weight vector shared across origins, define
\begin{equation}
 \bar Q_{ij}=\mathbb E_P[\langle Z_i(X)-Y,Z_j(X)-Y\rangle_d],\qquad
 \bar R(w)=w^\top\bar Qw.\label{eq:static_risk}
\end{equation}
The fitting matrix $\widehat Q$ averages error products on aggregator-training windows. It can differ from $\bar Q$ through finite data, temporal dependence, or distribution shift. Our implementation optimizes mixture MSE through softmax logits; it does not explicitly need to store $\widehat Q$.

\textbf{Proposition 5 (Estimation and Optimization Bound)}:
Suppose $\max_{i,j}|\widehat Q_{ij}-\bar Q_{ij}|\leq\eta$ and the returned $\widehat w\in\Delta_K$ satisfies
\begin{equation}
 \widehat w^\top\widehat Q\widehat w
 \leq\min_{w\in\Delta_K}w^\top\widehat Qw+\gamma,\qquad\gamma\geq0.
 \label{eq:approx_optimization}
\end{equation}
Then Equation~\ref{eq:weights_bound} holds. These are deterministic conditions on the realized fitting problem; no independence of candidate errors is required.

\textbf{Proof}:
For every simplex vector $w$,
\[
 |w^\top(\widehat Q-\bar Q)w|
 \leq\sum_{i,j}w_iw_j\eta=\eta.
\]
Let $w^*$ minimize $w^\top\bar Qw$ on the compact simplex. Successively applying this inequality, Equation~\ref{eq:approx_optimization}, and the same inequality again gives
\[
 \bar R(\widehat w)
 \leq\widehat w^\top\widehat Q\widehat w+\eta
 \leq (w^*)^\top\widehat Qw^*+\gamma+\eta
 \leq\bar R(w^*)+2\eta+\gamma.
\]
This proves the claim. The optimization condition compares against the closed simplex, including zero weights. Finite softmax logits give positive weights, but can approximate its boundary; any remaining approximation or optimization gap belongs in $\gamma$.

\subsubsection{Why A Larger Pool Need Not Improve Learned Aggregation}
For nested sets $S\subset S'$ of unchanged candidate rules, define $\bar R_S^*=\min_{w\in\Delta_{|S|}}\bar R_S(w)$ and the actual excess $E_S=\bar R_S(\widehat w_S)-\bar R_S^*\geq0$. Then
\begin{equation}
 \bar R_S(\widehat w_S)-\bar R_{S'}(\widehat w_{S'})
 =\underbrace{\bar R_S^*-\bar R_{S'}^*}_{\text{candidate opportunity}}
  -\underbrace{(E_{S'}-E_S)}_{\text{change in aggregation excess}}.
 \label{eq:opportunity_excess}
\end{equation}
The first term is nonnegative because the new weights can put zero mass on added rules. The second can have either sign. Under Proposition 5 for $S'$, an oracle gain exceeding $2\eta_{S'}+\gamma_{S'}$ is sufficient for improvement over the old deployed rule, since $E_S\geq0$; this is not necessary. Candidate count may increase estimation difficulty and optimization error, while a redundant hull gives no additional opportunity.

This comparison holds for the same deployment distribution and unchanged candidate rules. For static weights it concerns a convex hull of forecast \emph{functions}, not separate pointwise optima. Pointwise hull equality in Proposition 3 also implies equality for static mixtures of the fixed affine output path, because the effective $\delta$ is constant across histories. Refitting generators at each budget invalidates simple nesting. AdamW with a fixed number of updates, including the 160-update budget study, supplies no certificate for $\gamma$ or $\eta$. Thus the bound diagnoses requirements; it does not declare every measured configuration statistically certified.

\subsection{Risk scoring, temperature, and identity shrinkage}\label{app:score_theory}
\textbf{Proposition 6 (Conditional Exponential-Weight Bound)}:
Return to the conditional setting in Appendix~\ref{app:theory_setup}. Let observable scores $s_k$ satisfy $\max_k|s_k-R_k|\leq\varepsilon$. Let the temperature $T>0$ and prior $\pi\in\Delta_K$ be observable, with every $\pi_k>0$. Set
\begin{equation}
 w_k=\frac{\pi_k\exp(-s_k/T)}{\sum_j\pi_j\exp(-s_j/T)}.
 \label{eq:prior_gibbs}
\end{equation}
For any comparison weights $q\in\Delta_K$, with $0\log0=0$,
\begin{align}
 R(w)\leq{}&R(q)+D(q)-D(w)+2\varepsilon\nonumber\\
 &+T\big[\operatorname{KL}(q\|\pi)-\operatorname{KL}(w\|\pi)
                         -\operatorname{KL}(q\|w)\big].\label{eq:gibbs_comparator}
\end{align}
Consequently,
\begin{equation}
 R(w)\leq\min_k\{R_k+T\log(1/\pi_k)\}+2\varepsilon-D(w).
 \label{eq:prior_bound}
\end{equation}
Uniform $\pi$ gives Equation~\ref{eq:risk}, strengthening the version without $D(w)$.

\textbf{Proof.}
Let $A=\sum_j\pi_j e^{-s_j/T}$. Then $\log(w_k/\pi_k)=-s_k/T-\log A$. Substituting this into the definition of KL divergence yields, for every $q$,
\[
 q^\top s+T\operatorname{KL}(q\|\pi)
 =T\operatorname{KL}(q\|w)-T\log A.
\]
For $q=w$ the right side is $-T\log A$. Subtracting the two identities gives
\[
 w^\top s=q^\top s+T\big[\operatorname{KL}(q\|\pi)
 -\operatorname{KL}(w\|\pi)-\operatorname{KL}(q\|w)\big].
\]
Writing $\mathbf R=(R_0,\ldots,R_{K-1})^\top$, score accuracy implies $w^\top\mathbf R\leq w^\top s+\varepsilon$ and $q^\top s\leq q^\top\mathbf R+\varepsilon$. Applying Equation~\ref{eq:risk_identity} to both mixtures proves Equation~\ref{eq:gibbs_comparator}. For the vertex selecting candidate $k$, $D(q)=0$ and $\operatorname{KL}(q\|\pi)=\log(1/\pi_k)$. Dropping the other nonpositive KL terms and minimizing over $k$ proves Equation~\ref{eq:prior_bound}. All steps are conditional and allow fully dependent candidates.

\subsubsection{What The Bound Does and Does Not Establish}
Writing $\Delta=R_0-\min_kR_k\geq0$ when identity is retained, Equation~\ref{eq:risk} guarantees improvement over Frozen whenever
\begin{equation}
 \Delta+D(w)>2\varepsilon+T\log K.\label{eq:score_sufficient}
\end{equation}
This is a sufficient analytical condition, not an observable acceptance test: $\Delta$ and a valid $\varepsilon$ are unknown. Lower temperature reduces the entropy allowance, but does not repair inaccurate score ordering. In the hard-selection limit, the usual $2\varepsilon$ comparison with the best individual risk remains; hard selection can still miss a better convex combination. Taking $q$ as an optimal mixture in Equation~\ref{eq:gibbs_comparator} leaves a $D(q)$ term. Scores of individual losses alone therefore do not ensure competitive optimization of mixture loss, which depends on off-diagonal error products.

Even exact scores do not imply monotonicity at fixed temperature. Let $Y=0$, with identity $0$ and $n$ copies of forecast $1$. At $T=1$ and uniform prior, all bad candidates have exact risk one, yet their total weight is $n e^{-1}/(1+n e^{-1})$. The ensemble risk is its square and strictly increases with $n$. The hull never changes. A prior assigning fixed total mass to a group of identical forecasts, divided among its copies, would remove this duplication effect; the implemented uniform-prior rule does not do so.

\subsubsection{Identity Shrinkage Has an Exact Diversity Correction}
For any observable $\lambda\in[0,1]$ and $z_\lambda=(1-\lambda)z_w+\lambda Z_0$, Proposition 1 with two candidates gives
\begin{equation}
 R(z_\lambda)=(1-\lambda)R(w)+\lambda R_0
 -\lambda(1-\lambda)\|z_w-Z_0\|_d^2.\label{eq:shrink_exact}
\end{equation}
Replacing $R(w)$ by any valid upper bound gives a corresponding shrinkage bound, including its negative diversity term. It does not guarantee that a validation-chosen $\lambda<1$ improves on Frozen after a shift. Indeed, for any observed history where a rule outputs $z\ne Z_0$, a compatible future law with $Y=Z_0$ deterministically makes Frozen perfect and that rule strictly worse. With no constraint on changes in the future conditional law, universal dominance is impossible. Identity membership only makes a safe forecast \emph{available}; it does not make an imperfect selector safe.

The Risk regressor uses a clipped, normalized log-loss target; Classifier and Pairwise ranker use classification objectives. Their outputs are not calibrated absolute MSE estimates by construction. Proposition 6 must not be applied to those scores without an additional calibration argument. For Retrospective, $s_k=r_{t,k}$ and $T=\tau\max(r_{t,0},\epsilon)$ are observable; the algebra applies if score accuracy holds, but the use of an observed horizon alone does not establish that assumption.

\subsection{Historical scores under dependence and temporal drift}\label{app:temporal_theory}
The following sufficient conditions explain what would be needed to turn historical losses into a score-accuracy bound. They are not asserted for the reported data. Fix $K$ candidate rules before an audit sequence. Let $\mathcal F_0\subset\cdots\subset\mathcal F_n\subseteq\mathcal H_t$ be a filtration in which the prediction for block $i$ is available in $\mathcal F_{i-1}$ and its fully resolved loss $L_{ik}$ is measurable in $\mathcal F_i$. Completed, non-overlapping forecast blocks with intervening history included in the filtration are one admissible construction. Overlapping target blocks cannot simply be treated as independent observations.

Assume $0\leq L_{ik}\leq M$ almost surely for a known constant $M$, and use deterministic nonnegative weights $a_i$ summing to one. Define
\begin{equation}
 \begin{aligned}
 p_{ik}&=\mathbb E[L_{ik}\mid\mathcal F_{i-1}],&
 s_k&=\sum_i a_iL_{ik},\\
 \beta_t&=\max_k\left|\sum_i a_ip_{ik}-R_k\right|,&
 n_{\rm eff}&=\left(\sum_i a_i^2\right)^{-1}.
 \end{aligned}\label{eq:history_setup}
\end{equation}
Here, $R_k$ is current conditional risk, whereas $p_{ik}$ is the earlier predictable risk. Their difference includes changing regimes and histories, even for a stationary process with state-dependent predictability.

\textbf{Proposition 7 (Concentration Plus Drift)}:
For any $0<\alpha<1$, with probability at least $1-\alpha$ over the observed sequence,
\begin{equation}
 \max_k|s_k-R_k|
 \leq\beta_t+M\sqrt{\frac{\log(2K/\alpha)}{2n_{\rm eff}}}.\label{eq:history_bound}
\end{equation}
No independence across candidates or across blocks is assumed. The probability statement is over histories, not a post-observation guarantee conditional on each realized history.

\textbf{Proof}:
For each $k$, $L_{ik}-p_{ik}$ is a martingale difference. Given $\mathcal F_{i-1}$ it lies in an interval of length $M$. To verify the needed exponential bound, let $U$ be any conditionally centered variable in an interval of length $M$. Its conditional log moment generating function $\psi(\xi)=\log\mathbb E[e^{\xi U}\mid\mathcal F_{i-1}]$ satisfies $\psi(0)=\psi'(0)=0$. Its second derivative is the variance under exponential tilting. Tilting preserves the interval, so that variance is at most $M^2/4$: squared distance to the interval midpoint is at most $M^2/4$, and variance is no larger than the expected squared distance to any fixed point. Integrating the derivative bound twice gives $\psi(\xi)\leq\xi^2M^2/8$ for either sign of $\xi$. Applied to the scaled increment, this conditional Hoeffding argument \citep{azuma1967weighted,cesa2006prediction} yields
\[
 \mathbb E\!\left[\exp\{\xi a_i(L_{ik}-p_{ik})\}\mid\mathcal F_{i-1}\right]
 \leq\exp(\xi^2a_i^2M^2/8).
\]
Iterating conditional expectation across $i$ bounds the moment generating function of $S_k=\sum_i a_i(L_{ik}-p_{ik})$ by $\exp(\xi^2M^2\sum_i a_i^2/8)$. Markov's inequality and minimization over $\xi>0$, attained at $4u/(M^2\sum_i a_i^2)$, give
\[
 \Pr(S_k\geq u)\leq\exp\!\left(-\frac{2u^2}{M^2\sum_i a_i^2}\right).
\]
Applying the same argument to $-S_k$ and a union bound over $K$ rules bounds $\Pr(\max_k|S_k|\geq u)$ by $2K\exp(-2n_{\rm eff}u^2/M^2)$. Choose $u=M\sqrt{\log(2K/\alpha)/(2n_{\rm eff})}$, then use $|s_k-R_k|\leq|S_k|+|\sum_i a_ip_{ik}-R_k|$. This proves Equation~\ref{eq:history_bound}; $M=0$ is immediate without division.

\textbf{Implications and Limits}:
For equal weights $n_{\rm eff}=n$. This is a concentration parameter for martingale innovations around their changing predictable means, not a claim that dependent blocks behave as $n$ independent samples of current risk. The drift term can remain large as $n$ increases. If an external assumption bounds $\beta_t\leq b$, substituting $\varepsilon=b+M\sqrt{\log(2K/\alpha)/(2n_{\rm eff})}$ into Proposition 6 yields a high-probability current-risk comparison. A high-probability drift bound requires its own failure probability, added by a union bound. Without drift control, a longer scoring history offers no such certification.

The implemented retrospective rule uses one completed horizon, so this construction has $n=1$; the $HC$ entries of that horizon are not automatically independent audit samples. Standardized squared error is also not known to be bounded by a universal $M$. Clipping audit losses would change the risk being estimated; a claim about unclipped MSE would additionally require a tail bound. Data-dependent weighting, choosing rules after seeing audit labels, and overlapping forecasts require separate arguments. Thus Proposition 7 explains why score reliability can fail under shift without attributing an unverified statistical guarantee to the experiments.

\subsection{Compute budgets, oracle definitions, and experimental scope}\label{app:budget_theory}
\subsubsection{A Decision Criterion that Includes Cost}
On a fixed deployment distribution, let $C(S)$ denote the expected measured cost of generating and scoring set $S$, and let $\nu\geq0$ convert that cost to the units of risk. Define $J(S)=\bar R_S(\widehat w_S)+\nu C(S)$. For nested unchanged rules, Equation~\ref{eq:opportunity_excess} gives exactly
\begin{equation}
 J(S)-J(S')=(\bar R_S^*-\bar R_{S'}^*)-(E_{S'}-E_S)
             -\nu\{C(S')-C(S)\}.\label{eq:cost_criterion}
\end{equation}
Increasing computation is beneficial for this objective if and only if the right side is positive. This is a decomposition, not an algorithm that knows future risk. Output-only candidates share $F(X)$; input and composed paths require adjusted-context forecasts. Retrieval changes both the number of calls and context length. Historical scoring may need another complete pool. Consequently, $K$ is not a substitute for $C(S)$, and duplicated outputs can raise scoring cost without enlarging the forecast hull. Fitting a separate aggregator or adapter for each $K$ also incurs offline costs beyond the inference budget.

\subsubsection{The Reported Oracle is a Different Comparator}
The unavailable test-label selector chooses $k$ after observing $Y$. Its conditional expected loss satisfies
\begin{equation}
 \mathbb E[\min_k\|Z_k-Y\|_d^2\mid\mathcal H_t]
 \leq\min_k\mathbb E[\|Z_k-Y\|_d^2\mid\mathcal H_t]
 =\min_kR_k,\label{eq:hindsight_oracle}
\end{equation}
because the minimum realized loss is at most each realized loss before taking expectations. It is not ordered in general relative to the conditional convex optimum. With $Y=0$ and candidates $-1,1$, hindsight hard selection loses one whereas their mean loses zero. With $Y\in\{-1,1\}$ equiprobable and the same candidates, hindsight selection loses zero whereas the best observable convex forecast is zero with risk one. An observed Oracle improvement therefore measures hindsight candidate opportunity; it neither proves a deployable gain nor estimates the convex optimum in Proposition 2.

\subsubsection{Reading the Experimental Evidence}
Uniform has an exact risk identity but no general monotonicity guarantee. Learned weights and Greedy ensemble depend on mixture estimation and optimization; Greedy's restricted search can add approximation error. Classifier and Pairwise ranker aim at selection and need generalization and ranking accuracy, rather than automatically satisfying Proposition 6. Risk regression and Retrospective require additional score calibration to invoke that proposition. Validation-best candidate and Validation-Selection further depend on validation-to-test transfer. None gains a guarantee merely by including the frozen forecast.

The measured budget and adapter sweeps can test whether a particular fitted configuration improves, but cannot establish the conditional means, drift bounds, or uniform estimation guarantees assumed above. Seed variation, correlated datasets, and finite test origins further separate observed aggregate metrics from these population statements. The appropriate conclusion is conditional: additional candidates help when they extend useful forecast directions, their benefit survives aggregation error, and that recovered benefit justifies actual computation. More samples by themselves satisfy none of these three requirements.

\clearpage

\section{Algorithm}
\label{app_alg}

Algorithm~\ref{alg:ttc} specifies the complete workflow. Its generator includes every family, and its weight procedure covers every deployable rule. Algorithms~\ref{alg:generate} expand these calls.

\begin{algorithm}[H]
\caption{TTC4TS: offline fitting, validation selection, and test-time inference}\label{alg:ttc}
\small
\begin{algorithmic}[1]
\Require Pretrained $F$; pre-test series; context $L$, horizon $H$; observed test history
\Ensure Fitted adapters and aggregators, selected rule $m^*$, and forecasts $\widehat Y_t$
\State \textbf{Offline fitting:} freeze $F$; form purged chronological blocks $D_A,D_S,D_V$
\State Fit normalization on $D_A$; standardize data; build reference bank $\mathcal R$ from $D_A$
\State Initialize input/output residuals $R_\phi,R_\psi$; set $\Delta=\operatorname{linspace}(0,0.3,20)$
\For{each adapter update on a minibatch $(X,Y)\subset D_A$}
  \State Construct $Z_j^{\rm in},Z_j^{\rm out},Z_j^{\rm joint}$ by Equation~\ref{eq_adapter}, for all $\delta_j\in\Delta$
  \State Update only $(\phi,\psi)$ using $\frac{1}{3|\Delta|}\sum_j\sum_{v\in\{{\rm in,out,joint}\}}\operatorname{MSE}(Z_j^v,Y)$
\EndFor
\State Freeze adapters; fix the ordered pool: identity, adapters, augmentations, Retrieval-demo
\For{each origin $u$ in $D_S\cup D_V$}
  \State $Z_u\gets\Call{Generate}{X_u,u}$; $\widetilde Z_u\gets\Call{Generate}{X_{u-H},u-H}$ \Comment{Alg.~\ref{alg:generate}}
  \State $r_{u,k}\gets\operatorname{MSE}(\widetilde Z_{u,k},x_{u-H:u})$; $f_u\gets\operatorname{Features}(X_u,Z_u,r_u)$
\EndFor
\State On $D_S$, fit learned weights, Classifier, Risk, ranker and Greedy (Appendix~\ref{app_syn_define})
\State \textbf{Validation:} choose Retrospective $(\tau,\lambda)$ and best fixed candidate $k^*$ on $D_V$
\State Select $m^*$ by mean validation MSE over the nine deployment options; exclude Oracle
\State Freeze all fitted parameters, feature statistics and validation choices
\For{each test origin $t$ with only observations through $t-1$ available}
  \State $Z_t\gets\Call{Generate}{X_t,t}$; $r_t,f_t\gets\emptyset$ \Comment{Standardized history}
  \If{$m^*\in\{\text{Risk, ranker, Retrospective}\}$}
    \State $\widetilde Z_t\gets\Call{Generate}{X_{t-H},t-H}$
    \State $r_{t,k}\gets\operatorname{MSE}(\widetilde Z_{t,k},x_{t-H:t})$; $f_t\gets\operatorname{Features}(X_t,Z_t,r_t)$
  \EndIf
  \State $w_t\gets\Call{Weights}{m^*,Z_t,f_t,r_t}$ \Comment{no future labels}
  \State Output $\widehat Y_t=\sum_k w_{t,k}Z_{t,k}$ for this origin; no parameter update
\EndFor
\end{algorithmic}
\end{algorithm}

\begin{algorithm}[t]
\caption{\textsc{Generate}: all TTC4TS candidate families}\label{alg:generate}
\small
\begin{algorithmic}[1]
\Require Frozen $F,R_\phi,R_\psi$; fixed $\Delta$, 12 augmentations $\mathcal A$, bank $\mathcal R$; $Q=10$
\Function{Generate}{$X,u$}
  \State $B\gets F(X)$; $\mathcal I,\mathcal O,\mathcal J,\mathcal U,\mathcal D\gets$ empty ordered lists
  \For{each $\delta\in\Delta\setminus\{0\}$ in increasing order}
    \State Append $F(X+\delta R_\phi(X))$ to $\mathcal I$ \Comment{Input adapter}
    \State Append $B+\delta R_\psi(B)$ to $\mathcal O$ \Comment{Output adapter}
    \State $P\gets F(X+\delta R_\phi(X))$ \Comment{Separate call in the measured code}
    \State Append $P+\delta R_\psi(P)$ to $\mathcal J$ \Comment{Composed adapter}
  \EndFor
  \For{each augmentation $a\in\mathcal A$ in fixed order}
    \State $(X^a,h_a)\gets\operatorname{Transform}(a,X,H,\mathrm{seed})$
    \State Append $h_a(F(X^a))$ to $\mathcal U$
  \EndFor
  \State Mask bank pairs whose observed continuation ends after $u$
  \State For each channel, retrieve $Q$ nearest standardized 64-step pasts to $X_{L-64:L}$
  \State Assemble $Q$ multichannel pairs $(P_q,V_q)$ from these channel-wise neighbors
  \For{$q=1,\ldots,Q$}
    \State $(\bar P_q,\bar V_q)\gets\operatorname{Align}(P_q,V_q;X_{L-64:L})$ \Comment{Phase and amplitude}
    \State $C_q\gets[\bar P_q;\bar V_q;X]$ \Comment{Aligned past, observed continuation, query}
    \State Check $\operatorname{length}(C_q)\leq F$'s context capacity; otherwise raise an error
    \State Append $F(C_q)$ to $\mathcal D$ \Comment{One forecast per demonstration}
  \EndFor
  \State $Z\gets\operatorname{stack}([B]\mathbin{+}\mathcal I\mathbin{+}\mathcal O\mathbin{+}\mathcal J\mathbin{+}\mathcal U\mathbin{+}\mathcal D)$
  \State Check finite values and common shape $H\times C$; \Return $Z$
\EndFunction
\end{algorithmic}
\end{algorithm}

\subsection{Complete Procedures and Code Mapping}

Algorithm~\ref{alg:ttc} covers the complete training and deployment protocol; the two procedures below expand its generator and aggregator. $D_A$, $D_S$, and $D_V$ denote the adapter-fitting, aggregator-fitting, and validation blocks, respectively. $X_u=x_{u-L:u}$ and $x_{u-H:u}$ use half-open time intervals. All tensors below use the fixed normalization fitted on $D_A$. Returned forecasts are in those coordinates; the inverse scaler is applied when original-unit predictions are requested. $M=|\Delta|=20$ is the number of samples per adapter view, whereas $K=80$ is the total candidate count and $Q=10$ is the retrieval count.

The output-coordinate map $h_a$ undoes magnitude reflection, vertical translation, magnitude warping, or the future continuation of an added slope. It is the identity for perturbations without a justified forecast-space inverse, including time reversal. The $\delta=0$ identity is included only once. The ordered pool has $1+3(M-1)+12+Q=80$ candidates; retrieval contributes only independently generated demonstration forecasts. No direct retrieved future or cumulative retrieval average is a candidate. The separate input/joint calls in Algorithm~\ref{alg:generate} preserve the measured implementation's 61 wrapper calls per current-origin pool, even though reusing them could reduce cost.

\paragraph{Transformation details.} Magnitude warping applies a positive sinusoidal scale to the mean-centered history and divides the forecast's centered values by its continuation through the horizon. Added slopes are removed using the corresponding future slope. Time reversal, smoothing, time warping, horizontal shifts, jitter and spikes perturb the conditioning history; they do not define invertible transformations of future dynamics. In particular, the predicted future is not reversed for time-flipped inputs. Stateless jitter depends only on the observed window and fixed seed. Historical forecasts use the same transformations and timestamp masks as current forecasts; on a horizon-spaced stream, they may be cached once their targets become observed.

\subsection{Fitting objectives and validation choices}
For each origin $u$ in $D_S$, form candidate errors $\ell_{u,k}=\operatorname{MSE}(Z_{u,k},Y_u)$ and historical errors $r_{u,k}=\operatorname{MSE}(Z_{u-H,k},x_{u-H:u})$. The latter use only an already observed horizon. Fit the following components on $D_S$ only; $\mathbb E_S$ denotes the empirical mean over its fitting origins:
\begin{align}
 b^*&=\arg\min_b\;\mathbb E_S\operatorname{MSE}\!\left(\sum_k\operatorname{softmax}(b)_kZ_{u,k},Y_u\right),\label{eq:fitweights}\\
 \theta_C^*&=\arg\min_{\theta_C}\;\mathbb E_S\operatorname{CE}\!\left(c_{\theta_C}(Z_u),\arg\min_k\ell_{u,k}\right),\label{eq:fitclassifier}\\
 \theta_R^*&=\arg\min_{\theta_R}\;\mathbb E_{S,k}\operatorname{smoothL}_1\!\left(s_{\theta_R}(f_{u,k}),\rho_{u,k}\right),\label{eq:fitrisk}\\
 \rho_{u,k}&=\log\!\left(1+\min\!\left(100,\frac{\ell_{u,k}}{\max(\ell_{u,0},10^{-5})}\right)\right),\nonumber\\
 \theta_P^*&=\arg\min_{\theta_P}\;\mathbb E_{S,i,j:\ell_{u,i}\ne\ell_{u,j}}\operatorname{softplus}\!\left[
 (s_{\theta_P}(f_{u,i})-s_{\theta_P}(f_{u,j}))\operatorname{sign}(\ell_{u,i}-\ell_{u,j})\right].\label{eq:fitrank}
\end{align}
These are numerical optimization objectives, not claims that a global optimum is attained. $c$ is the supplied trajectory Classifier, applied independently to flattened $HC$-dimensional candidates; $s$ is the feature MLP plus candidate bias. Feature means/scales are fitted on $D_S$. All four objectives use 160 full-batch AdamW updates at learning rate 0.01 and weight decay 0.03 in the main comparison. Adapter fitting in Algorithm~\ref{alg:ttc} instead uses 32 Adam minibatch updates, differentiating through $F$ with its parameters frozen. The zero-Delta loss is constant and contributes no parameter gradient.

Ensemble learning starts with $\bar Z_u^{(0)}=0$ and zero counts. At step $n=1,\ldots,40$, select
\begin{equation}
 \begin{aligned}
 k_n&=\arg\min_k\mathbb E_S\operatorname{MSE}\!\left(\frac{(n-1)\bar Z_u^{(n-1)}+Z_{u,k}}{n},Y_u\right),\\
 \bar Z_u^{(n)}&=\frac{(n-1)\bar Z_u^{(n-1)}+Z_{u,k_n}}{n}.
 \end{aligned}
\end{equation}
Its fixed weight $g_k$ is the fraction of the 40 selections assigned to candidate $k$. Selection is with replacement; ties use the first candidate in the fixed order. Uniform needs no fitting. Validation-best uses $k^*=\arg\min_k\mathbb E_V\ell_{u,k}$. Retrospective selects $(\tau,\lambda)$ by mixture MSE on $D_V$ over $\{0.1,0.3,1\}\times\{0,0.5,1\}$. Validation-selected TTC4TS selects one rule from the nine options, in that order for exact ties. Oracle is an evaluation-only diagnostic, $Z_{t,\arg\min_k\operatorname{MSE}(Z_{t,k},Y_t)}$, and is never available to that choice or deployment.

\section{Experimental Setup and Results}

\subsection{Implementation Details}
\label{app_imple}

The information of the experiment datasets used in this paper are summarized as follows: (1) Electricity Transformer Temperature (ETT) dataset \cite{Zhou2021Informer}, which contains the data collected from two electricity transformers in two separated counties in China, including the load and the oil temperature recorded every 15 minutes (ETTm) or 1 hour (ETTh) between July 2016 and July 2018. (2) Electricity (ECL) dataset \footnote[1]{https://archive.ics.uci.edu/ml/datasets/ElectricityLoadDiagrams20112014} collects the hourly electricity consumption of 321 clients (each column) from 2012 to 2014. (3) Exchange \cite{lai2018modeling} records the current exchange of 8 different countries from 1990 to 2016. (4) Traffic dataset \footnote[2]{http://pems.dot.ca.gov} records the occupation rate of freeway system across State of California measured by 861 sensors. (5) Weather dataset \footnote[3]{https://www.bgc-jena.mpg.de/wetter} records every 10 minutes for 21 meteorological indicators in Germany throughout 2020.
Appliances, Beijing PM2.5  dataset provides by \cite{appliances2017} and \cite{beijing2015}. 
The detailed statistics information of the datasets is shown in Table \ref{tb_dataset}.

\begin{wraptable}{r}{0.45\textwidth}
   \vspace{-1.3em}
    \centering
    \caption{Summary statistics of the datasets, including sequence length, number of variables, and sampling frequency.}
    \vspace{-0.5em}
    \label{tb_dataset}
     \resizebox{0.46\textwidth}{!}
     {
        \setlength{\tabcolsep}{4pt}
        \begin{tabular}{cccc}
          \toprule
          Dataset    & length  & features & frequency \\
          \midrule
          ETTh1       & 17,420  & 7       & 1h \\
          ETTh2       & 17,420  & 7       & 1h \\
          ETTm1       & 69,680  & 7       & 15m \\
          ETTm2       & 69,680  & 7       & 15m \\
          Electricity & 26,304  & 321     & 1h  \\
          Exchange    & 7,588   & 8       & 1d  \\
          Traffic     & 17,544  & 862     & 1h  \\
          Weather     & 52,696  & 21      & 10m \\
          \bottomrule
        \end{tabular}
     }
     \vspace{-0.8em}
\end{wraptable}


ETT uses the standard chronological 12/4/4-month boundaries. The other datasets use 70/10/20 chronological splits. 
The first 60\% of the nominal training period fits adapters, normalization, and the retrieval bank; the rest fits aggregators after a one-horizon purge. The middle development block selects hyperparameters and the deployment rule. Test origins are spaced by $H$, for $H\in\{96,192,336,720\}$, and all channels are retained. We report MSE and MAE in fitting-standardized units. These choices differ from stride-one LTSF evaluation and from fitting a scaler on the complete nominal training block; only comparisons within the same protocol are directly interpretable.

The context length is 512, with at most 96 fitting windows per component. Adapters receive 32 Adam updates at learning rate $10^{-4}$, averaging the three adapter views and all 20 Delta values; batch size is one for horizon 720 or checkpoints named base, otherwise two. Adapter and Classifier hidden widths are 512; maximum Delta is 0.3 and retrieval uses top-$K=10$. Aggregators receive 160 updates at learning rate 0.01 and weight decay 0.03; budget-specific fits use 40 updates. Seeds 0--2 vary initialization and adapter minibatches. The 80-candidate pool contains identity, 57 adapter candidates, 12 augmentations, and 10 retrieval demonstrations. Saved predictions hold targets and alternatives fixed across rules, and parameter hashes verify the frozen backbone. PyTorch experiments use eight 32GB V100 GPUs, one task per device. Appendix Table~\ref{tab:deployment_cost} reports synchronized cold-query latency including retrospective forecasts.


\subsection{Reproduction details}
With half-open index intervals, adapter origins satisfy $t\ge L$ and $t+H\le e_{\rm adapt}$. Aggregator origins satisfy $t\ge e_{\rm adapt}+H$ and $t+H\le e_{\rm train}$; validation uses $t\ge e_{\rm train}$ and $t+H\le e_{\rm val}$; testing uses $t\ge e_{\rm val}$ and $t+H\le e_{\rm test}$. Retrospective targets are $[t-H,t)$, so none used to train an aggregator lies before the adapter boundary. Audits verify these inequalities; earlier test observations may legitimately enter a later test origin's history.

The supplied PyTorch implementation stores per-origin targets, all candidate forecasts, learned adapter and aggregator states, fitted feature statistics, exact split boundaries, dataset hashes, frozen-backbone hashes, and software versions. Every result table is reconstructed from these files. Model parameters are checked before and after adapter fitting. Test labels enter only the metric computation and the explicitly marked oracle diagnostic. Training retrieval excludes reference pairs ending after the relevant origin. Unit tests cover these information boundaries, degenerate normalization gradients, forecast-space inverses, and the shared-residual adapter forward and joint-gradient path.

All six checkpoints use controlled point decoders. Bolt emits median predictions in 64-step blocks, TimesFM in 128-step blocks, and Timer uses its point head recursively. Chronos-2 predicts channel-independent median quantiles directly. Sundial shares one transformer representation across eight fixed Gaussian noise trajectories and applies 50 flow-decoder steps to each trajectory before averaging. A recorded backbone call means one wrapper prediction, not one internal transformer or flow-network pass; latency includes all internal computation.



\section{Detailed Experimental Results}
\label{app_exp_reslut}

\subsection{Comparison with SOTA Methods}
\label{app:baseline_comparison}
\label{app_baseline_compar}

This comparison fixes the TTC4TS synthesizer to \textbf{learned weights}: one global simplex vector per model--dataset--horizon--seed cell, fitted on the aggregator-training block.

The lower block of Table~\ref{tab:baseline_summary} keeps the existing shared TTC4TS adapters: one input, output, or composed prediction at $\delta=0.3$, a validation-best adapter candidate including identity, and learned weights over identity plus each generator family. These are component ablations rather than independently trained prior methods. Full Uniform and aligned-demo averaging with identity are retained to distinguish changing the generator pool from changing its weights.

\subsection{Measured comparison}

Component controls further limit the claim that the complete generator is necessary. Learned weights over identity and augmentations alone reach relative MSE 0.935, slightly below the full pool's 0.936; adapter-only learned weights reach 0.985. Full TTC4TS has better relative MAE than the augmentation-only control (0.988 versus 0.990), but it cannot claim uniformly better error or efficiency. All three generator families provide an interface for experimentation; this evaluation does not establish that retaining every family is optimal.

The 432-cell audit recomputes 7,344 test metric records and their validation counterparts from saved forecasts, verifies frozen model and source hashes, and checks every validation-selected placement and checkpoint. One preliminary TimesFM identity check required reproducing the original cache batch to remove batch-layout rounding differences; its prediction then matched exactly. The correction changed only verification, and both source versions and the initial failure record are retained. All final cell statuses are successful. Seed standard deviations summarize repeated fitting, not independent datasets; deterministic retrieval variants reuse verified seed-zero predictions and accordingly have zero seed dispersion.

\clearpage
\begin{table}[H]
\centering
\caption{Geometric relative MSE by model and dataset on 6 datasets, aggregating horizons 96/192/336/720 and seeds 0--2. Each ratio uses the matching Frozen result.}\label{tab:baseline_datasets_mse}
\fontsize{8.2}{10.2}\selectfont
\setlength{\tabcolsep}{3.2pt}
\begin{tabular}{lrrrrrrrr}
\toprule
Dataset & AdaPTS & $\delta$-Adp+In  & $\delta$-Apt+Out & TS-RAG  & Align-RAG  & TTC4TS+U & TTC4TS+RR & TTC4TS+LW \\
\midrule
\multicolumn{9}{l}{\textbf{Bolt-S}} \\
ETTh1 & \uline{0.985} & 0.998 & 0.986 & 0.996 & 0.996 & 0.996 & 0.995 & \textbf{0.906} \\
ETTh2 & 0.996 & 1.011 & 1.010 & 0.997 & \textbf{0.983} & 1.000 & 0.995 & \uline{0.988} \\
ETTm1 & 0.970 & 0.990 & 0.965 & 0.976 & 1.022 & 0.996 & \uline{0.958} & \textbf{0.888} \\
ETTm2 & 1.003 & 1.012 & 1.024 & 1.004 & 1.016 & \uline{0.973} & \textbf{0.972} & 0.980 \\
Appliances & 1.025 & 0.991 & 1.023 & \uline{0.990} & 1.045 & 1.071 & 0.992 & \textbf{0.969} \\
BeijingWeather & 0.920 & 0.974 & \uline{0.880} & 0.931 & \textbf{0.861} & 0.902 & 0.903 & 0.892 \\
\midrule
\multicolumn{9}{l}{\textbf{Bolt-B}} \\
ETTh1 & \uline{0.967} & 1.000 & 0.969 & 0.992 & 0.996 & 0.980 & 0.981 & \textbf{0.889} \\
ETTh2 & 1.004 & 1.013 & 1.015 & 1.004 & \textbf{0.931} & \uline{0.950} & 0.973 & 0.970 \\
ETTm1 & \uline{0.970} & 0.999 & \uline{0.970} & \uline{0.970} & 1.001 & 0.981 & 0.975 & \textbf{0.893} \\
ETTm2 & 0.982 & 1.015 & 1.000 & 0.982 & 1.039 & \textbf{0.960} & 0.970 & \uline{0.963} \\
Appliances & 1.003 & 0.999 & 1.028 & 0.988 & 1.039 & 1.018 & \textbf{0.982} & \uline{0.987} \\
BeijingWeather & 0.977 & 0.981 & 0.983 & 0.968 & \uline{0.878} & 0.880 & \textbf{0.835} & 0.913 \\
\midrule
\multicolumn{9}{l}{\textbf{TimesFM}} \\
ETTh1 & 0.996 & 1.012 & 0.999 & 1.006 & \uline{0.975} & 0.989 & 0.985 & \textbf{0.938} \\
ETTh2 & 1.011 & 1.012 & 1.039 & 1.021 & 1.058 & \textbf{0.927} & \uline{0.947} & 0.974 \\
ETTm1 & 0.995 & 1.009 & 1.005 & 1.013 & 1.000 & 0.962 & \uline{0.930} & \textbf{0.897} \\
ETTm2 & 1.002 & 1.016 & 1.010 & 0.997 & 1.085 & 0.970 & \uline{0.969} & \textbf{0.931} \\
Appliances & 1.011 & 0.999 & 1.011 & 1.007 & 0.970 & 0.925 & \uline{0.921} & \textbf{0.918} \\
BeijingWeather & 0.743 & 0.981 & 0.881 & 0.981 & \uline{0.557} & \textbf{0.551} & \textbf{0.551} & 0.861 \\
\midrule
\multicolumn{9}{l}{\textbf{Timer}} \\
ETTh1 & 1.001 & 1.018 & 1.017 & 1.011 & 0.985 & \textbf{0.971} & 0.982 & \uline{0.972} \\
ETTh2 & 1.014 & 1.017 & 1.031 & 1.016 & 1.060 & \textbf{0.974} & \uline{0.976} & \uline{0.976} \\
ETTm1 & \uline{0.995} & 0.998 & 1.000 & 0.999 & 1.246 & 1.269 & 1.018 & \textbf{0.978} \\
ETTm2 & 1.005 & 1.015 & 1.026 & 1.001 & 1.165 & \textbf{0.956} & \uline{0.964} & \uline{0.964} \\
Appliances & \uline{0.956} & 1.009 & 0.966 & 0.961 & 1.065 & 0.994 & 0.983 & \textbf{0.954} \\
BeijingWeather & 0.998 & 1.009 & 1.011 & 1.009 & \textbf{0.964} & 0.997 & 1.017 & \uline{0.972} \\
\midrule
\multicolumn{9}{l}{\textbf{Sundial}} \\
ETTh1 & 1.005 & 1.012 & 1.016 & 1.008 & \uline{0.984} & 0.994 & 0.990 & \textbf{0.978} \\
ETTh2 & 0.996 & 1.011 & 1.006 & 1.000 & 1.007 & \textbf{0.968} & \uline{0.970} & 0.972 \\
ETTm1 & 1.015 & 1.007 & 1.025 & 1.007 & 0.993 & 1.000 & \uline{0.978} & \textbf{0.963} \\
ETTm2 & 0.999 & 1.013 & 1.013 & 1.005 & 1.125 & 0.970 & \uline{0.969} & \textbf{0.963} \\
Appliances & 1.029 & 1.000 & 1.034 & 1.025 & 0.998 & 0.941 & \textbf{0.937} & \uline{0.940} \\
BeijingWeather & 0.986 & 0.992 & 0.986 & 0.981 & \uline{0.956} & 0.985 & 0.986 & \textbf{0.955} \\
\midrule
\multicolumn{9}{l}{\textbf{Chronos-2}} \\
ETTh1 & 0.990 & 1.001 & 0.994 & 0.994 & \uline{0.943} & 0.962 & 0.965 & \textbf{0.883} \\
ETTh2 & 0.992 & 1.011 & 1.009 & 0.994 & 0.993 & \textbf{0.936} & 0.951 & \uline{0.950} \\
ETTm1 & 1.023 & 0.990 & 1.026 & 1.013 & \uline{0.913} & 0.917 & 0.917 & \textbf{0.885} \\
ETTm2 & 1.003 & 1.013 & 1.028 & 1.002 & 1.004 & 0.972 & \uline{0.967} & \textbf{0.960} \\
Appliances & 0.958 & 0.984 & 0.961 & 0.984 & 0.937 & 0.927 & \uline{0.913} & \textbf{0.898} \\
BeijingWeather & 0.847 & 0.979 & 0.798 & 0.854 & \uline{0.791} & 0.839 & \textbf{0.788} & 0.803 \\
\bottomrule
\end{tabular}
\end{table}

\clearpage
\begin{table}[H]
\centering
\caption{Geometric relative MAE by model and dataset on 6 datasets, aggregating horizons 96/192/336/720 and seeds 0--2. Each ratio uses the matching Frozen result.}\label{tab:baseline_datasets_mae}
\fontsize{8.2}{10.2}\selectfont
\setlength{\tabcolsep}{3.2pt}
\begin{tabular}{lrrrrrrrr}
\toprule
Dataset & AdaPTS & $\delta$-Adp+In  & $\delta$-Apt+Out & TS-RAG  & Align-RAG  & TTC4TS+U & TTC4TS+RR & TTC4TS+LW \\
\midrule
\multicolumn{9}{l}{\textbf{Bolt-S}} \\
ETTh1 & 1.004 & 1.026 & 1.030 & 1.023 & 1.002 & 0.998 & \uline{0.997} & \textbf{0.982} \\
ETTh2 & 1.007 & 1.027 & 1.034 & 1.009 & 1.013 & 1.003 & \textbf{0.998} & \uline{1.000} \\
ETTm1 & 1.005 & 1.036 & 1.036 & 1.022 & 1.056 & 1.048 & \uline{0.990} & \textbf{0.978} \\
ETTm2 & 1.016 & 1.037 & 1.052 & 1.020 & 1.021 & \uline{0.990} & \textbf{0.984} & 0.996 \\
Appliances & 1.035 & 1.014 & 1.039 & 1.015 & 1.018 & 1.029 & \textbf{0.997} & \uline{1.009} \\
BeijingWeather & 1.040 & 1.128 & 1.126 & 1.092 & \textbf{1.007} & \uline{1.009} & 1.010 & \textbf{1.007} \\
\midrule
\multicolumn{9}{l}{\textbf{Bolt-B}} \\
ETTh1 & 0.987 & 1.026 & 1.014 & 1.020 & 0.998 & \uline{0.986} & 0.989 & \textbf{0.976} \\
ETTh2 & 1.012 & 1.030 & 1.035 & 1.014 & \uline{0.970} & \textbf{0.963} & 0.983 & 0.983 \\
ETTm1 & 0.996 & 1.040 & 1.034 & 1.005 & 1.020 & 1.004 & \uline{0.990} & \textbf{0.980} \\
ETTm2 & 1.002 & 1.039 & 1.040 & 1.006 & 1.031 & \textbf{0.977} & \uline{0.985} & 0.990 \\
Appliances & 1.024 & 1.018 & 1.039 & 1.017 & 1.010 & \uline{1.004} & \textbf{0.991} & 1.021 \\
BeijingWeather & 1.058 & 1.126 & 1.167 & 1.084 & 1.011 & \uline{1.009} & \textbf{0.998} & 1.010 \\
\midrule
\multicolumn{9}{l}{\textbf{TimesFM}} \\
ETTh1 & 1.010 & 1.036 & 1.034 & 1.032 & \uline{1.000} & 1.005 & 1.002 & \textbf{0.988} \\
ETTh2 & 1.006 & 1.029 & 1.039 & 1.017 & 1.043 & \textbf{0.962} & \uline{0.966} & 0.984 \\
ETTm1 & 1.000 & 1.038 & 1.035 & 1.018 & 1.019 & 1.003 & \uline{0.977} & \textbf{0.964} \\
ETTm2 & 1.000 & 1.038 & 1.033 & 0.999 & 1.061 & 0.985 & \uline{0.984} & \textbf{0.968} \\
Appliances & 1.022 & 1.018 & 1.035 & 1.022 & 0.990 & \uline{0.962} & \textbf{0.961} & 0.969 \\
BeijingWeather & 1.027 & 1.133 & 1.191 & 1.133 & 0.907 & \uline{0.900} & \textbf{0.899} & 1.026 \\
\midrule
\multicolumn{9}{l}{\textbf{Timer}} \\
ETTh1 & 1.002 & 1.033 & 1.033 & 1.017 & 1.012 & \textbf{0.986} & \uline{0.990} & 0.993 \\
ETTh2 & 1.007 & 1.027 & 1.032 & 1.012 & 1.068 & \uline{0.985} & \uline{0.985} & \textbf{0.984} \\
ETTm1 & \uline{0.991} & 1.020 & 1.011 & 1.008 & 1.117 & 1.118 & 1.012 & \textbf{0.990} \\
ETTm2 & 1.003 & 1.024 & 1.028 & 1.003 & 1.143 & \textbf{0.987} & 0.991 & \uline{0.988} \\
Appliances & 0.989 & 1.012 & 1.000 & 0.994 & 1.024 & \uline{0.986} & \textbf{0.985} & 0.993 \\
BeijingWeather & 1.000 & 1.081 & 1.073 & 1.039 & 0.995 & \uline{0.994} & \uline{0.994} & \textbf{0.982} \\
\midrule
\multicolumn{9}{l}{\textbf{Sundial}} \\
ETTh1 & 1.009 & 1.022 & 1.032 & 1.018 & 1.014 & 1.009 & \uline{1.000} & \textbf{0.997} \\
ETTh2 & 1.002 & 1.024 & 1.025 & 1.012 & 1.053 & \textbf{0.982} & \uline{0.984} & 0.987 \\
ETTm1 & 1.016 & 1.029 & 1.044 & 1.022 & 1.030 & 1.010 & \uline{0.993} & \textbf{0.989} \\
ETTm2 & 1.007 & 1.030 & 1.034 & 1.013 & 1.123 & \uline{0.982} & \uline{0.982} & \textbf{0.979} \\
Appliances & 1.028 & 1.010 & 1.035 & 1.027 & 0.993 & \textbf{0.964} & \uline{0.967} & 0.974 \\
BeijingWeather & 1.026 & 1.069 & 1.077 & 1.055 & 0.990 & 0.985 & \uline{0.984} & \textbf{0.966} \\
\midrule
\multicolumn{9}{l}{\textbf{Chronos-2}} \\
ETTh1 & 1.007 & 1.030 & 1.036 & 1.021 & \uline{0.986} & 0.987 & 0.987 & \textbf{0.975} \\
ETTh2 & 1.010 & 1.033 & 1.041 & 1.017 & 1.009 & \textbf{0.977} & \uline{0.980} & 0.985 \\
ETTm1 & 1.015 & 1.039 & 1.054 & 1.026 & 1.003 & 0.993 & \textbf{0.977} & \uline{0.980} \\
ETTm2 & 1.018 & 1.045 & 1.061 & 1.018 & 1.021 & \uline{0.989} & \textbf{0.986} & 0.992 \\
Appliances & 1.029 & 1.015 & 1.036 & 1.015 & 0.981 & \uline{0.977} & \textbf{0.969} & 0.979 \\
BeijingWeather & 1.053 & 1.145 & 1.132 & 1.121 & \textbf{0.947} & 0.971 & \uline{0.956} & 1.003 \\
\bottomrule
\end{tabular}
\end{table}

\subsection{Experiments on More Datasets}
\label{app_more_datasets}

\begin{figure}[H]
\centering\includegraphics[width=\linewidth]{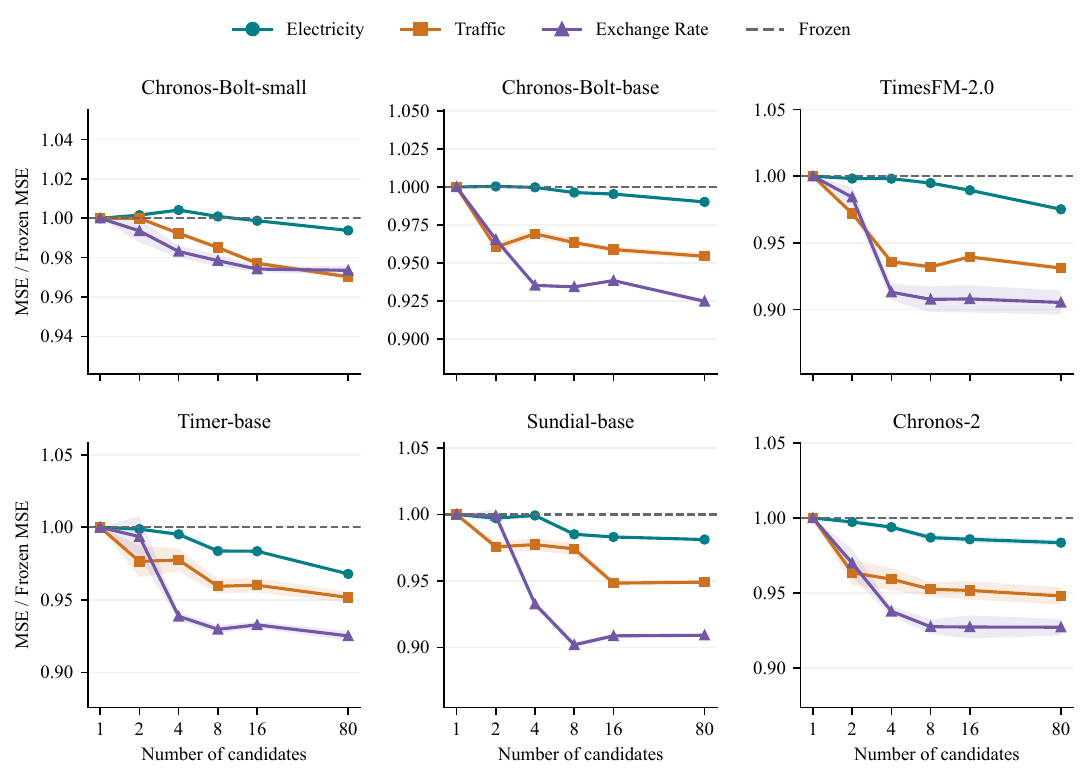}
\caption{Candidate-budget curves on other 3 common datasets. }\label{fig:otherdatasets}
\end{figure}

Figure \ref{fig:otherdatasets} illustrates the impact of the number of candidates on forecasting performance across six pretrained forecasting backbones and three datasets. As the number of candidates increases, most models consistently achieve lower MSE ratios, demonstrating the effectiveness of candidate selection in improving forecasting accuracy. Overall, the results indicate that leveraging more candidate forecasts can effectively enhance the performance of pretrained time-series models.

\subsection{Domain sensitivity and deployment cost}

Giving equal weight to the ETT family, Appliances, and BeijingWeather yields TTC4TS relative MSE 0.919, compared with 0.940 when each dataset receives equal weight. The domain table shows that improvement is largest on BeijingWeather; the overall Classifier failure also hides a gain on that domain. The three domain groups are too few for a broad generalization claim.

\begin{table}[H]
\centering\small
\caption{Domain sensitivity. Within-domain relative MSE averages checkpoints, horizons and seeds geometrically. Lower is better. VS denotes Validation-Selection. VS denotes Validation-Selection.}\label{tab:domain_summary}
\begin{tabular}{lrrrr}
\toprule
Domain & Classifier & Risk & Val.-best & VS \\
\midrule
Appliances & 1.419 & \uuline{0.971} & 1.025 & \textbf{0.957} \\
BeijingWeather & \uuline{0.863} & 0.890 & 0.868 & \textbf{0.842} \\
ETT family & 1.294 & \uuline{0.976} & 0.989 & \textbf{0.962} \\
\bottomrule
\end{tabular}
\end{table}

Timing uses batch-one float32 inference, one warmup and eight timed repeats with CUDA synchronization; median and 90th-percentile latency and peak allocated GPU memory are saved. Loading, input transfer, and service overhead are excluded. The measured implementation makes 61 wrapper calls for a full current-origin pool and 122 when retrospective features are needed, versus one for Frozen. Input and joint views recompute the same input-adapted backbone forecasts separately, and sharing them is a possible implementation optimization but is not credited in these timings.


\subsection{Measured computation}

\begin{table}[H]
\centering
\caption{Median synchronized inference latency (milliseconds) on ETTh1 at horizon 96, seed 0, first test origin, batch one, one warmup and three repeats on a V100. Point decoders and context length 512 are fixed; loading and transfers are excluded. Learned weights do not require retrospective forecasts. Timings are measured in the concurrent experiment environment and are not isolated-service throughput benchmarks.}\label{tab:baseline_latency}
\fontsize{8.3}{10.2}\selectfont
\setlength{\tabcolsep}{3.2pt}
\begin{tabular}{lrrrrrrr}
\toprule
Backbone & Frozen & Input & Output & Joint & Raw demo & Align demo & TTC4TS-LW \\
\midrule
Bolt-S & 42.9 & \textbf{40.7} & 43.0 & \uuline{41.5} & 415.2 & 461.6 & 2594.7 \\
Bolt-B & \textbf{73.5} & 75.8 & \uuline{74.7} & 75.4 & 759.6 & 839.4 & 4618.1 \\
TimesFM & \uuline{50.7} & 52.8 & 51.1 & \textbf{50.3} & 512.6 & 564.8 & 3233.6 \\
Timer & \textbf{9.1} & \uuline{9.7} & 9.8 & 10.5 & 91.0 & 110.2 & 633.7 \\
Sundial & 118.5 & \textbf{104.8} & \uuline{105.1} & 105.2 & 1465.9 & 1164.5 & 7069.2 \\
Chronos-2 & 35.6 & \textbf{32.9} & \uuline{33.1} & 33.6 & 345.4 & 387.1 & 2235.3 \\
\bottomrule
\end{tabular}

{\footnotesize\emph{Note.} Best: bold; second-best: double underline. Ties use displayed precision.\par}
\end{table}

\begin{table}[H]
\centering\small
\caption{Median cold-query inference latency in milliseconds on ETTh1 at horizon 96 and seed 0, measured on one V100 with all channels and a batch of one origin. The 80-candidate pool includes top-10 retrieval demonstrations. Risk includes retrospective forecasts; Classifier scores trajectories directly. Eight timed repeats follow a warmup; data and models reside on the GPU. VS denotes Validation-Selection.}\label{tab:deployment_cost}
\begin{tabular}{lrrrrr}
\toprule
Backbone & Frozen & Uniform & Classifier & Risk & VS \\
\midrule
Bolt-S & \textbf{39.9} & 2516.6 & \uuline{2401.2} & 4919.6 & 2502.3 \\
Bolt-B & \textbf{68.5} & 4298.8 & 4375.0 & 8342.6 & \uuline{4060.2} \\
TimesFM & \textbf{44.8} & 3038.4 & \uuline{2717.2} & 5733.2 & 2920.4 \\
Timer & \textbf{8.6} & 598.8 & \uuline{591.3} & 1186.2 & 1217.8 \\
Sundial & \textbf{98.4} & 7819.9 & \uuline{6047.5} & 12397.8 & 6056.9 \\
Chronos-2 & \textbf{31.5} & 1989.3 & 2094.1 & 3976.2 & \uuline{1958.7} \\
\bottomrule
\end{tabular}
\end{table}

The median synchronized inference latency and the median cold-query inference latency are shown in Tables~\ref{tab:baseline_latency} and \ref{tab:deployment_cost}. It can seen that single-forecast adapters require one backbone prediction, plus their small residual network(s). Ten separate unaligned demonstrations require ten predictions. The current aligned generator also evaluates identity, for eleven wrapper calls. Full TTC4TS evaluates 61 wrapper calls: shared input and composed adapter paths still recompute their input-adjusted forecasts separately. Learned weights require no retrospective candidate pool. These implementation costs must accompany the accuracy comparison, and more candidates are not evidence of matched-latency superiority. 

Offline fitting is also measured separately. Across 432 cells per placement, median standalone fitting times are 38.5 seconds for input adapters, 11.7 seconds for output adapters, and 38.0 seconds for composed adapters. Output-only fitting caches frozen training forecasts and avoids backbone backpropagation. These medians summarize heterogeneous cells in the concurrent experiment environment; they are not a matched-training-cost comparison with TTC4TS.

\section{Detailed Experimental Results of Generator}\label{app:figure_breakdowns}
Tables~\ref{tab:figure2_datasets}--\ref{tab:figure3_model6} disaggregate Figures~\ref{fig:families} and~\ref{fig:scaling}. Entries average per-cell MSE ratios to Frozen geometrically over $H\in\{96,192,336,720\}$ and seeds 0--2. Re-aggregation reproduces every plotted value; these ratios are not absolute errors.

\begin{table}[H]
\centering
\caption{Detailed experiment results of Figure~\ref{fig:families} by model and dataset: ETTh1, ETTh2, ETTm1, ETTm2, Appliances, and BeijingWeather. Relative MSE is averaged geometrically over four horizons {96, 192,336, and 720} and three seeds. A denotes adapters, U augmentations, and R top-10 Retrieval-demo. Every family pool includes the identity once and uses Uniform. Frozen is 1.000, and lower is better. }\label{tab:figure2_datasets}
\fontsize{9}{10.0}\selectfont
\setlength{\tabcolsep}{3.0pt}
\renewcommand{\arraystretch}{1.04}
\begin{tabular}{lrrrrrrrr}
\toprule
Dataset & Frozen & A & U & R & A+R & A+U & A+U+R & U+R \\
\midrule
\multicolumn{9}{l}{\textbf{Chronos-Bolt-small}} \\
ETTh1 & 1.000 & 0.997 & 0.970 & 0.972 & 0.995 & \textbf{0.916} & \uline{0.934} & 0.995 \\
ETTh2 & 1.000 & 1.002 & 0.994 & 0.993 & 0.999 & \textbf{0.972} & \uline{0.978} & 0.999 \\
ETTm1 & 1.000 & 0.991 & 0.953 & 0.943 & 0.970 & \textbf{0.911} & \uline{0.922} & 0.982 \\
ETTm2 & 1.000 & 1.001 & 0.992 & 0.980 & 0.986 & 0.971 & \textbf{0.951} & \uline{0.969} \\
Appliances & 1.000 & 1.010 & 1.000 & \textbf{0.994} & 1.001 & \uline{0.998} & 1.002 & 1.053 \\
BeijingWeather & 1.000 & 0.979 & 0.966 & 0.954 & 0.962 & 0.930 & \uline{0.910} & \textbf{0.907} \\
\midrule
\multicolumn{9}{l}{\textbf{Chronos-Bolt-base}} \\
ETTh1 & 1.000 & 0.993 & 0.963 & 0.964 & 0.989 & \textbf{0.899} & \uline{0.921} & 0.980 \\
ETTh2 & 1.000 & 1.001 & 0.991 & 0.982 & 0.989 & 0.957 & \textbf{0.946} & \uline{0.952} \\
ETTm1 & 1.000 & 0.990 & 0.954 & 0.950 & 0.978 & \uline{0.926} & \textbf{0.923} & 0.976 \\
ETTm2 & 1.000 & 0.994 & 0.984 & 0.975 & 0.982 & 0.963 & \textbf{0.945} & \uline{0.958} \\
Appliances & 1.000 & 1.008 & 1.001 & \uline{0.993} & 0.997 & 1.009 & \textbf{0.991} & 1.008 \\
BeijingWeather & 1.000 & 0.988 & 0.972 & 0.957 & 0.967 & 0.916 & \uline{0.894} & \textbf{0.888} \\
\midrule
\multicolumn{9}{l}{\textbf{TimesFM-2.0}} \\
ETTh1 & 1.000 & 0.995 & 0.974 & 0.966 & 0.982 & \uline{0.947} & \textbf{0.935} & 0.982 \\
ETTh2 & 1.000 & 0.996 & 0.987 & 0.973 & 0.978 & 0.962 & \uline{0.934} & \textbf{0.928} \\
ETTm1 & 1.000 & 0.975 & 0.952 & \uline{0.934} & 0.947 & 0.967 & \textbf{0.928} & 0.947 \\
ETTm2 & 1.000 & 0.960 & 0.950 & \uline{0.940} & 0.946 & 0.946 & \textbf{0.930} & 0.961 \\
Appliances & 1.000 & 0.956 & 0.953 & 0.939 & 0.939 & 0.991 & \uline{0.933} & \textbf{0.921} \\
BeijingWeather & 1.000 & 0.944 & 0.924 & 0.875 & 0.883 & 0.871 & \uline{0.718} & \textbf{0.577} \\
\midrule
\multicolumn{9}{l}{\textbf{Timer-base}} \\
ETTh1 & 1.000 & 1.002 & 0.995 & 0.988 & 0.993 & 1.017 & \uline{0.981} & \textbf{0.971} \\
ETTh2 & 1.000 & 1.004 & 0.998 & 0.990 & 0.993 & 0.983 & \textbf{0.968} & \uline{0.972} \\
ETTm1 & \uline{1.000} & \uline{1.000} & \textbf{0.999} & 1.005 & 1.001 & 1.078 & 1.135 & 1.220 \\
ETTm2 & 1.000 & 1.002 & 0.996 & 0.985 & 0.988 & 0.978 & \uline{0.957} & \textbf{0.955} \\
Appliances & 1.000 & 0.977 & 0.976 & \uline{0.970} & \textbf{0.969} & 0.991 & 0.979 & 0.989 \\
BeijingWeather & 1.000 & 0.999 & 0.991 & 0.989 & 0.995 & \textbf{0.970} & \uline{0.976} & 0.995 \\
\midrule
\multicolumn{9}{l}{\textbf{Sundial-base}} \\
ETTh1 & 1.000 & 1.002 & 0.990 & \uline{0.985} & 0.992 & 1.001 & \textbf{0.982} & 0.990 \\
ETTh2 & 1.000 & 0.997 & 0.994 & 0.987 & 0.988 & 0.989 & \uline{0.971} & \textbf{0.968} \\
ETTm1 & 1.000 & 1.006 & \uline{0.988} & \textbf{0.981} & 0.993 & 1.028 & 0.990 & 0.992 \\
ETTm2 & 1.000 & 0.996 & 0.994 & 0.985 & 0.985 & 1.006 & \uline{0.974} & \textbf{0.968} \\
Appliances & 1.000 & 1.022 & 1.016 & 0.994 & 0.994 & 1.013 & \uline{0.954} & \textbf{0.937} \\
BeijingWeather & 1.000 & 0.988 & 0.983 & 0.978 & 0.982 & \uline{0.972} & \textbf{0.971} & 0.983 \\
\midrule
\multicolumn{9}{l}{\textbf{Chronos-2}} \\
ETTh1 & 1.000 & 0.994 & 0.964 & 0.961 & 0.985 & \textbf{0.897} & \uline{0.910} & 0.963 \\
ETTh2 & 1.000 & 0.991 & 0.980 & 0.970 & 0.977 & 0.950 & \textbf{0.934} & \uline{0.937} \\
ETTm1 & 1.000 & 1.003 & 0.966 & 0.945 & 0.970 & 0.926 & \textbf{0.886} & \uline{0.913} \\
ETTm2 & 1.000 & 0.999 & 0.984 & 0.975 & 0.984 & \uline{0.952} & \textbf{0.944} & 0.968 \\
Appliances & 1.000 & 0.964 & 0.951 & 0.935 & 0.941 & 0.958 & \textbf{0.917} & \uline{0.920} \\
BeijingWeather & 1.000 & 0.860 & 0.863 & 0.854 & \uline{0.849} & 0.910 & 0.868 & \textbf{0.846} \\
\bottomrule
\end{tabular}

{\footnotesize\emph{Note.} Best: bold; second-best: double underline. Ties use displayed precision.\par}
\end{table}

\subsection{Candidate-budget values for every aggregation method}
The following six tables retain all methods and budgets in Figure~\ref{fig:scaling}. Columns denote Frozen (F), Uniform (U), Learned weights (W), Classifier (C), Risk regressor (Risk), Pairwise ranker (Rank), Ensemble learning (E), Retrospective (Rt), Validation-best candidate (VB), Validation-selected TTC4TS (TTC), and unavailable Oracle (O). $K$ is the total candidate budget, not the per-adapter sample count. All tables use the measured budget-specific fits and tie rules in Appendix~\ref{app_syn_define}; in particular, $K=80$ is taken from the budget experiment rather than substituted from the main aggregation table. Bold uses unrounded values and excludes Oracle.
\begin{table}[H]
\centering
\caption{Chronos-Bolt-small: dataset-specific candidate-budget relative MSE underlying Figure~\ref{fig:scaling}. Every entry averages four horizons and three seeds geometrically; candidate budget is $K$. Column definitions are given in Appendix~\ref{app:figure_breakdowns}. O is the unavailable test-label Oracle.}\label{tab:figure3_model1}
\fontsize{8.3}{10.2}\selectfont
\setlength{\tabcolsep}{2.2pt}
\renewcommand{\arraystretch}{1.04}
\begin{tabular}{lrrrrrrrrrrrr}
\toprule
Dataset & $K$ & F & U & W & C & Risk & Rank & E & Rt & VB & VS & O \\
\midrule
ETTh1 & 1 & \textbf{1.000} & \textbf{1.000} & \textbf{1.000} & \textbf{1.000} & \textbf{1.000} & \textbf{1.000} & \textbf{1.000} & \textbf{1.000} & \textbf{1.000} & \textbf{1.000} & 1.000 \\
 & 2 & 1.000 & \uline{0.997} & \textbf{0.996} & 0.999 & \uline{0.997} & 1.002 & \textbf{0.996} & 0.999 & 1.001 & 1.000 & 0.987 \\
 & 4 & 1.000 & \textbf{0.963} & 0.965 & 1.010 & 0.966 & 1.001 & \uline{0.964} & 0.980 & 1.040 & 0.966 & 0.918 \\
 & 8 & 1.000 & 0.984 & \uline{0.972} & 1.026 & 0.985 & 0.995 & \textbf{0.961} & 0.998 & 1.040 & 0.993 & 0.899 \\
 & 16 & 1.000 & 0.988 & \uline{0.978} & 1.081 & 0.987 & 0.997 & \textbf{0.961} & 0.999 & 1.040 & 0.990 & 0.899 \\
 & 80 & 1.000 & 0.972 & \uline{0.906} & 1.098 & 0.982 & 0.985 & \textbf{0.879} & 1.005 & 0.941 & 0.927 & 0.703 \\
\midrule
ETTh2 & 1 & \textbf{1.000} & \textbf{1.000} & \textbf{1.000} & \textbf{1.000} & \textbf{1.000} & \textbf{1.000} & \textbf{1.000} & \textbf{1.000} & \textbf{1.000} & \textbf{1.000} & 1.000 \\
 & 2 & \textbf{1.000} & 1.003 & 1.004 & 1.005 & 1.003 & 1.003 & 1.008 & \uline{1.001} & 1.002 & 1.003 & 0.984 \\
 & 4 & 1.000 & \textbf{0.982} & \uline{0.983} & 1.000 & \textbf{0.982} & \uline{0.983} & 0.991 & \textbf{0.982} & 0.992 & 0.991 & 0.925 \\
 & 8 & 1.000 & \textbf{0.981} & 0.983 & 0.997 & \textbf{0.981} & 0.998 & 0.991 & \uline{0.982} & 0.999 & 1.006 & 0.866 \\
 & 16 & 1.000 & \uline{0.990} & 0.991 & 0.997 & \uline{0.990} & 0.991 & 0.991 & \textbf{0.989} & 0.998 & 1.000 & 0.866 \\
 & 80 & 1.000 & 0.993 & \uline{0.988} & 1.058 & 0.995 & 1.039 & 1.006 & \textbf{0.978} & 1.020 & 1.020 & 0.751 \\
\midrule
ETTm1 & 1 & \textbf{1.000} & \textbf{1.000} & \textbf{1.000} & \textbf{1.000} & \textbf{1.000} & \textbf{1.000} & \textbf{1.000} & \textbf{1.000} & \textbf{1.000} & \textbf{1.000} & 1.000 \\
 & 2 & 1.000 & \textbf{0.987} & \uline{0.989} & 0.997 & \textbf{0.987} & 0.996 & 0.990 & 0.993 & 1.000 & 0.994 & 0.964 \\
 & 4 & 1.000 & \textbf{0.949} & 0.954 & 0.978 & \uline{0.951} & 0.994 & 0.962 & 0.970 & 1.000 & 0.966 & 0.835 \\
 & 8 & 1.000 & \textbf{0.952} & \uline{0.953} & 1.025 & \textbf{0.952} & 0.992 & \uline{0.953} & 0.968 & 0.995 & 0.963 & 0.802 \\
 & 16 & 1.000 & 0.969 & \uline{0.961} & 1.065 & 0.969 & 0.998 & \textbf{0.953} & 0.975 & 0.995 & 0.972 & 0.802 \\
 & 80 & 1.000 & 0.943 & \uline{0.888} & 1.884 & 0.962 & 1.065 & \textbf{0.885} & 0.966 & 0.967 & 0.908 & 0.655 \\
\midrule
ETTm2 & 1 & \textbf{1.000} & \textbf{1.000} & \textbf{1.000} & \textbf{1.000} & \textbf{1.000} & \textbf{1.000} & \textbf{1.000} & \textbf{1.000} & \textbf{1.000} & \textbf{1.000} & 1.000 \\
 & 2 & \uline{1.000} & \uline{1.000} & 1.003 & 1.004 & \uline{1.000} & \uline{1.000} & 1.007 & \textbf{0.998} & 1.002 & \textbf{0.998} & 0.983 \\
 & 4 & 1.000 & \textbf{0.966} & 0.971 & 0.996 & \uline{0.967} & 0.995 & 0.986 & 0.970 & 1.005 & 0.974 & 0.907 \\
 & 8 & 1.000 & \textbf{0.971} & 0.974 & 0.994 & \uline{0.972} & 0.991 & 0.983 & 0.973 & 1.003 & 0.980 & 0.861 \\
 & 16 & 1.000 & \textbf{0.983} & \textbf{0.983} & 1.006 & \uline{0.984} & 0.986 & \uline{0.984} & \uline{0.984} & 1.002 & 0.986 & 0.860 \\
 & 80 & 1.000 & \textbf{0.980} & \textbf{0.980} & 1.457 & \uline{0.983} & 1.106 & 0.987 & 0.986 & 1.009 & 1.021 & 0.775 \\
\midrule
Appliances & 1 & \textbf{1.000} & \textbf{1.000} & \textbf{1.000} & \textbf{1.000} & \textbf{1.000} & \textbf{1.000} & \textbf{1.000} & \textbf{1.000} & \textbf{1.000} & \textbf{1.000} & 1.000 \\
 & 2 & 1.000 & 1.021 & 1.004 & 1.017 & 1.019 & 1.015 & \textbf{0.998} & \uline{0.999} & 1.000 & 1.002 & 0.983 \\
 & 4 & 1.000 & 1.010 & 1.002 & 1.084 & 1.008 & \uline{0.999} & \textbf{0.997} & \textbf{0.997} & 1.018 & \textbf{0.997} & 0.959 \\
 & 8 & \uline{1.000} & 1.012 & 1.006 & 1.088 & 1.011 & 1.005 & \textbf{0.997} & \uline{1.000} & 1.027 & 1.025 & 0.939 \\
 & 16 & 1.000 & 1.001 & \uline{0.998} & 1.082 & 1.001 & 1.009 & \textbf{0.997} & 0.999 & 1.023 & 1.005 & 0.935 \\
 & 80 & 1.000 & 0.994 & 0.969 & 1.466 & 0.994 & 1.023 & \textbf{0.955} & 0.993 & 1.007 & \uline{0.963} & 0.853 \\
\midrule
BeijingWeather & 1 & \textbf{1.000} & \textbf{1.000} & \textbf{1.000} & \textbf{1.000} & \textbf{1.000} & \textbf{1.000} & \textbf{1.000} & \textbf{1.000} & \textbf{1.000} & \textbf{1.000} & 1.000 \\
 & 2 & 1.000 & 0.968 & 0.964 & \uline{0.963} & 0.964 & \textbf{0.961} & \textbf{0.961} & 0.968 & 0.989 & 0.967 & 0.936 \\
 & 4 & 1.000 & 0.907 & 0.913 & 1.032 & \uline{0.870} & \textbf{0.777} & 1.042 & 0.995 & 0.999 & 0.993 & 0.742 \\
 & 8 & 1.000 & 0.930 & 0.904 & \uline{0.802} & 0.900 & \textbf{0.771} & 0.958 & 0.966 & 0.864 & 0.869 & 0.733 \\
 & 16 & 1.000 & 0.953 & 0.939 & 0.938 & \uline{0.929} & \textbf{0.773} & 0.958 & 0.954 & 0.978 & 0.966 & 0.733 \\
 & 80 & 1.000 & 0.954 & 0.892 & 0.966 & 0.947 & \textbf{0.773} & \uline{0.861} & 0.954 & 0.984 & 0.907 & 0.696 \\
\bottomrule
\end{tabular}
\end{table}

\clearpage
\begin{table}[H]
\centering
\caption{Chronos-Bolt-base: dataset-specific candidate-budget relative MSE underlying Figure~\ref{fig:scaling}. Datasets are ETTh1, ETTh2, ETTm1, ETTm2, Appliances, and BeijingWeather. Every entry averages four horizons and three seeds geometrically; candidate budget is $K$. Column definitions are given in Appendix~\ref{app:figure_breakdowns}. O is the unavailable test-label Oracle.}\label{tab:figure3_model2}
\fontsize{8.3}{10.2}\selectfont
\setlength{\tabcolsep}{2.2pt}
\renewcommand{\arraystretch}{1.04}
\begin{tabular}{lrrrrrrrrrrrr}
\toprule
Dataset & $K$ & F & U & W & C & Risk & Rank & E & Rt & VB & VS & O \\
\midrule
ETTh1 & 1 & \textbf{1.000} & \textbf{1.000} & \textbf{1.000} & \textbf{1.000} & \textbf{1.000} & \textbf{1.000} & \textbf{1.000} & \textbf{1.000} & \textbf{1.000} & \textbf{1.000} & 1.000 \\
 & 2 & 1.000 & \uline{0.990} & 0.991 & 0.995 & \textbf{0.989} & 0.997 & 0.992 & 0.997 & 1.001 & 0.999 & 0.982 \\
 & 4 & 1.000 & \textbf{0.959} & 0.962 & 1.009 & \uline{0.961} & 0.982 & 0.965 & 0.985 & 1.035 & 0.993 & 0.899 \\
 & 8 & 1.000 & 0.973 & \uline{0.966} & 1.037 & 0.974 & 1.014 & \textbf{0.960} & 0.994 & 1.030 & 1.005 & 0.880 \\
 & 16 & 1.000 & 0.982 & \uline{0.972} & 1.028 & 0.982 & 1.043 & \textbf{0.960} & 0.995 & 1.029 & 1.008 & 0.880 \\
 & 80 & 1.000 & 0.964 & \uline{0.889} & 1.030 & 0.973 & 1.006 & \textbf{0.858} & 0.988 & 0.949 & 0.956 & 0.683 \\
\midrule
ETTh2 & 1 & \textbf{1.000} & \textbf{1.000} & \textbf{1.000} & \textbf{1.000} & \textbf{1.000} & \textbf{1.000} & \textbf{1.000} & \textbf{1.000} & \textbf{1.000} & \textbf{1.000} & 1.000 \\
 & 2 & \textbf{1.000} & 1.002 & 1.003 & 1.006 & 1.002 & 1.003 & 1.006 & \uline{1.001} & 1.004 & 1.004 & 0.988 \\
 & 4 & 1.000 & \uline{0.969} & 0.971 & 0.987 & \uline{0.969} & 0.980 & 0.995 & \textbf{0.968} & 1.001 & 0.987 & 0.908 \\
 & 8 & 1.000 & \textbf{0.969} & 0.975 & 0.987 & \uline{0.970} & 0.988 & 0.993 & \textbf{0.969} & 1.003 & 1.003 & 0.871 \\
 & 16 & 1.000 & \uline{0.983} & 0.987 & 0.990 & 0.984 & \textbf{0.976} & 0.993 & \uline{0.983} & 1.003 & 0.999 & 0.870 \\
 & 80 & 1.000 & 0.982 & \uline{0.970} & 1.088 & 0.983 & 1.087 & 0.972 & \textbf{0.965} & 0.992 & 0.992 & 0.735 \\
\midrule
ETTm1 & 1 & \textbf{1.000} & \textbf{1.000} & \textbf{1.000} & \textbf{1.000} & \textbf{1.000} & \textbf{1.000} & \textbf{1.000} & \textbf{1.000} & \textbf{1.000} & \textbf{1.000} & 1.000 \\
 & 2 & 1.000 & \textbf{0.985} & 0.988 & 0.995 & \textbf{0.985} & 0.991 & 0.989 & 0.988 & 0.998 & \uline{0.986} & 0.964 \\
 & 4 & 1.000 & \uline{0.956} & 0.959 & 1.007 & \textbf{0.955} & 0.974 & 0.966 & 0.961 & 0.998 & 0.957 & 0.839 \\
 & 8 & 1.000 & 0.963 & 0.960 & 1.025 & 0.960 & 0.971 & \textbf{0.957} & 0.965 & 0.977 & \uline{0.958} & 0.815 \\
 & 16 & 1.000 & 0.976 & 0.969 & 1.032 & 0.974 & 0.973 & \uline{0.957} & 0.977 & 0.980 & \textbf{0.955} & 0.815 \\
 & 80 & 1.000 & 0.950 & 0.893 & 1.375 & 0.966 & 1.015 & \textbf{0.884} & 0.972 & 0.925 & \uline{0.888} & 0.682 \\
\midrule
ETTm2 & 1 & \textbf{1.000} & \textbf{1.000} & \textbf{1.000} & \textbf{1.000} & \textbf{1.000} & \textbf{1.000} & \textbf{1.000} & \textbf{1.000} & \textbf{1.000} & \textbf{1.000} & 1.000 \\
 & 2 & 1.000 & \textbf{0.989} & \uline{0.991} & 1.006 & \textbf{0.989} & 0.997 & 0.993 & \uline{0.991} & 0.999 & 0.992 & 0.964 \\
 & 4 & 1.000 & \textbf{0.956} & \uline{0.957} & 0.993 & 0.958 & 0.981 & 0.959 & \textbf{0.956} & 0.987 & 0.964 & 0.901 \\
 & 8 & 1.000 & \uline{0.963} & \uline{0.963} & 0.968 & 0.965 & 0.970 & \textbf{0.959} & 0.964 & 0.987 & 0.966 & 0.858 \\
 & 16 & 1.000 & 0.978 & 0.976 & 0.980 & 0.979 & 0.971 & \textbf{0.959} & 0.976 & 0.986 & \uline{0.966} & 0.857 \\
 & 80 & 1.000 & 0.975 & 0.963 & 1.371 & 0.978 & 1.019 & \textbf{0.955} & 0.984 & 1.006 & \uline{0.962} & 0.757 \\
\midrule
Appliances & 1 & \textbf{1.000} & \textbf{1.000} & \textbf{1.000} & \textbf{1.000} & \textbf{1.000} & \textbf{1.000} & \textbf{1.000} & \textbf{1.000} & \textbf{1.000} & \textbf{1.000} & 1.000 \\
 & 2 & 1.000 & 1.010 & 0.999 & 1.016 & 1.010 & 1.017 & \textbf{0.996} & \uline{0.997} & 1.000 & 1.001 & 0.981 \\
 & 4 & 1.000 & 0.997 & 0.993 & 1.040 & 0.997 & 1.015 & \uline{0.992} & \textbf{0.991} & 1.000 & 1.001 & 0.946 \\
 & 8 & 1.000 & 0.999 & 0.995 & 1.071 & 0.999 & 1.012 & \textbf{0.990} & \uline{0.993} & 1.067 & 1.029 & 0.930 \\
 & 16 & 1.000 & 0.998 & \uline{0.994} & 1.049 & 0.998 & 1.008 & \textbf{0.990} & 0.996 & 1.067 & 1.019 & 0.927 \\
 & 80 & 1.000 & 0.993 & 0.987 & 1.357 & 0.993 & 1.038 & \textbf{0.982} & 0.990 & 1.060 & \uline{0.984} & 0.868 \\
\midrule
BeijingWeather & 1 & \textbf{1.000} & \textbf{1.000} & \textbf{1.000} & \textbf{1.000} & \textbf{1.000} & \textbf{1.000} & \textbf{1.000} & \textbf{1.000} & \textbf{1.000} & \textbf{1.000} & 1.000 \\
 & 2 & 1.000 & 0.990 & 0.993 & 1.000 & \uline{0.988} & 0.992 & 1.002 & \textbf{0.987} & 1.006 & 0.991 & 0.968 \\
 & 4 & 1.000 & 0.922 & 0.933 & 0.848 & 0.918 & \textbf{0.841} & 0.975 & 0.924 & 0.878 & \uline{0.843} & 0.802 \\
 & 8 & 1.000 & 0.916 & 0.905 & \uline{0.774} & 0.904 & \textbf{0.754} & 0.932 & 0.953 & 0.871 & 0.842 & 0.712 \\
 & 16 & 1.000 & 0.950 & 0.946 & \uline{0.765} & 0.944 & \textbf{0.753} & 0.952 & 0.975 & 0.871 & 0.849 & 0.712 \\
 & 80 & 1.000 & 0.957 & 0.913 & \uline{0.802} & 0.954 & \textbf{0.751} & 0.837 & 0.966 & 0.946 & 0.943 & 0.675 \\
\bottomrule
\end{tabular}
\end{table}

\clearpage
\begin{table}[H]
\centering
\caption{TimesFM-2.0: dataset-specific candidate-budget relative MSE underlying Figure~\ref{fig:scaling}. Datasets are ETTh1, ETTh2, ETTm1, ETTm2, Appliances, and BeijingWeather. Every entry averages four horizons and three seeds geometrically; candidate budget is $K$. Column definitions are given in Appendix~\ref{app:figure_breakdowns}. O is the unavailable test-label Oracle.}\label{tab:figure3_model3}
\fontsize{8.3}{10.2}\selectfont
\setlength{\tabcolsep}{2.2pt}
\renewcommand{\arraystretch}{1.04}
\begin{tabular}{lrrrrrrrrrrrr}
\toprule
Dataset & $K$ & F & U & W & C & Risk & Rank & E & Rt & VB & VS & O \\
\midrule
ETTh1 & 1 & \textbf{1.000} & \textbf{1.000} & \textbf{1.000} & \textbf{1.000} & \textbf{1.000} & \textbf{1.000} & \textbf{1.000} & \textbf{1.000} & \textbf{1.000} & \textbf{1.000} & 1.000 \\
 & 2 & 1.000 & \uline{0.994} & \uline{0.994} & 1.004 & \textbf{0.993} & 0.998 & \uline{0.994} & 0.997 & 1.005 & \uline{0.994} & 0.973 \\
 & 4 & 1.000 & \textbf{0.934} & 0.938 & 1.003 & \uline{0.935} & 0.968 & 0.943 & 0.967 & 1.026 & 1.013 & 0.821 \\
 & 8 & 1.000 & 0.956 & \textbf{0.947} & 0.982 & 0.957 & 0.977 & \uline{0.949} & 0.966 & 1.023 & 1.003 & 0.780 \\
 & 16 & 1.000 & 0.971 & \uline{0.956} & 1.014 & 0.971 & 1.003 & \textbf{0.950} & 0.979 & 1.023 & 0.996 & 0.779 \\
 & 80 & 1.000 & \uline{0.966} & \textbf{0.938} & 1.030 & 0.971 & 0.987 & \textbf{0.938} & 0.990 & 1.043 & 0.972 & 0.713 \\
\midrule
ETTh2 & 1 & \textbf{1.000} & \textbf{1.000} & \textbf{1.000} & \textbf{1.000} & \textbf{1.000} & \textbf{1.000} & \textbf{1.000} & \textbf{1.000} & \textbf{1.000} & \textbf{1.000} & 1.000 \\
 & 2 & \textbf{1.000} & \textbf{1.000} & \uline{1.007} & 1.009 & \textbf{1.000} & \textbf{1.000} & 1.023 & \textbf{1.000} & \uline{1.007} & \textbf{1.000} & 0.977 \\
 & 4 & 1.000 & \uline{0.940} & 0.948 & 0.953 & \textbf{0.938} & 0.961 & 0.995 & \uline{0.940} & 0.962 & 0.944 & 0.850 \\
 & 8 & 1.000 & 0.945 & 0.947 & 0.944 & 0.945 & 0.945 & 0.978 & 0.943 & \uline{0.938} & \textbf{0.926} & 0.801 \\
 & 16 & 1.000 & 0.967 & 0.967 & \textbf{0.936} & 0.966 & 0.962 & 0.989 & 0.949 & 0.938 & \uline{0.937} & 0.799 \\
 & 80 & 1.000 & 0.973 & 0.974 & 1.106 & \uline{0.971} & 1.117 & 1.010 & \textbf{0.955} & 0.996 & 0.996 & 0.722 \\
\midrule
ETTm1 & 1 & \textbf{1.000} & \textbf{1.000} & \textbf{1.000} & \textbf{1.000} & \textbf{1.000} & \textbf{1.000} & \textbf{1.000} & \textbf{1.000} & \textbf{1.000} & \textbf{1.000} & 1.000 \\
 & 2 & 1.000 & \textbf{0.968} & \uline{0.969} & 0.997 & \textbf{0.968} & 1.003 & \uline{0.969} & 0.976 & 1.010 & 0.978 & 0.911 \\
 & 4 & 1.000 & \uline{0.926} & \uline{0.926} & 1.047 & \textbf{0.921} & 0.973 & 0.932 & 0.943 & 1.011 & 0.948 & 0.748 \\
 & 8 & 1.000 & 0.929 & \uline{0.928} & 1.088 & \uline{0.928} & 0.977 & \textbf{0.925} & 0.935 & 1.008 & 0.944 & 0.716 \\
 & 16 & 1.000 & 0.946 & \uline{0.933} & 1.107 & 0.946 & 0.983 & \textbf{0.926} & 0.952 & 1.009 & 0.958 & 0.711 \\
 & 80 & 1.000 & 0.934 & \uline{0.897} & 1.932 & 0.943 & 1.022 & \textbf{0.878} & 0.963 & 0.992 & 0.906 & 0.602 \\
\midrule
ETTm2 & 1 & \textbf{1.000} & \textbf{1.000} & \textbf{1.000} & \textbf{1.000} & \textbf{1.000} & \textbf{1.000} & \textbf{1.000} & \textbf{1.000} & \textbf{1.000} & \textbf{1.000} & 1.000 \\
 & 2 & 1.000 & \textbf{0.964} & 0.974 & 1.015 & \textbf{0.964} & 0.988 & 1.003 & \uline{0.966} & 1.008 & 0.971 & 0.938 \\
 & 4 & 1.000 & \textbf{0.924} & \uline{0.928} & 1.003 & \textbf{0.924} & 0.981 & 0.940 & 0.930 & 1.015 & 0.940 & 0.851 \\
 & 8 & 1.000 & \uline{0.932} & \textbf{0.931} & 1.020 & \uline{0.932} & 0.997 & 0.939 & 0.937 & 1.006 & 0.938 & 0.800 \\
 & 16 & 1.000 & 0.940 & \textbf{0.933} & 1.009 & 0.940 & 0.978 & \uline{0.935} & 0.946 & 0.996 & 0.939 & 0.791 \\
 & 80 & 1.000 & \uline{0.940} & \textbf{0.931} & 1.329 & 0.941 & 1.055 & 0.949 & 0.948 & 1.026 & 0.953 & 0.738 \\
\midrule
Appliances & 1 & \textbf{1.000} & \textbf{1.000} & \textbf{1.000} & \textbf{1.000} & \textbf{1.000} & \textbf{1.000} & \textbf{1.000} & \textbf{1.000} & \textbf{1.000} & \textbf{1.000} & 1.000 \\
 & 2 & 1.000 & \uline{0.963} & \textbf{0.958} & 1.010 & \uline{0.963} & 1.013 & \textbf{0.958} & \uline{0.963} & 1.034 & \uline{0.963} & 0.957 \\
 & 4 & 1.000 & \textbf{0.924} & \textbf{0.924} & 1.011 & \textbf{0.924} & 1.001 & \textbf{0.924} & 0.931 & 1.024 & \uline{0.929} & 0.888 \\
 & 8 & 1.000 & 0.936 & \uline{0.926} & 1.047 & 0.937 & 0.991 & \textbf{0.918} & 0.935 & 1.068 & 0.928 & 0.867 \\
 & 16 & 1.000 & 0.939 & \uline{0.929} & 1.048 & 0.939 & 0.983 & \textbf{0.917} & 0.940 & 1.083 & 0.967 & 0.862 \\
 & 80 & 1.000 & 0.939 & \uline{0.918} & 1.477 & 0.939 & 0.991 & \textbf{0.912} & 0.941 & 1.087 & 0.964 & 0.813 \\
\midrule
BeijingWeather & 1 & \textbf{1.000} & \textbf{1.000} & \textbf{1.000} & \textbf{1.000} & \textbf{1.000} & \textbf{1.000} & \textbf{1.000} & \textbf{1.000} & \textbf{1.000} & \textbf{1.000} & 1.000 \\
 & 2 & 1.000 & 0.886 & 0.854 & 0.787 & 0.780 & \textbf{0.751} & 0.789 & 0.914 & 1.000 & \uline{0.774} & 0.719 \\
 & 4 & 1.000 & 0.729 & 0.691 & 0.687 & 0.640 & \textbf{0.563} & 0.605 & 0.735 & 0.677 & \uline{0.585} & 0.540 \\
 & 8 & 1.000 & 0.766 & 0.711 & 0.860 & 0.674 & \textbf{0.561} & \uline{0.575} & 0.764 & 0.677 & 0.690 & 0.497 \\
 & 16 & 1.000 & 0.891 & 0.888 & 0.917 & 0.709 & \textbf{0.523} & 0.596 & 0.896 & 0.677 & \uline{0.552} & 0.496 \\
 & 80 & 1.000 & 0.875 & 0.861 & 0.572 & 0.694 & \textbf{0.493} & 0.595 & 0.875 & 0.561 & \uline{0.559} & 0.439 \\
\bottomrule
\end{tabular}
\end{table}

\clearpage
\begin{table}[H]
\centering
\caption{Timer-base: dataset-specific candidate-budget relative MSE underlying Figure~\ref{fig:scaling}. Datasets are ETTh1, ETTh2, ETTm1, ETTm2, Appliances, and BeijingWeather. Every entry averages four horizons and three seeds geometrically; candidate budget is $K$. Column definitions are given in Appendix~\ref{app:figure_breakdowns}. O is the unavailable test-label Oracle.}\label{tab:figure3_model4}
\fontsize{8.3}{10.2}\selectfont
\setlength{\tabcolsep}{2.2pt}
\renewcommand{\arraystretch}{1.04}
\begin{tabular}{lrrrrrrrrrrrr}
\toprule
Dataset & $K$ & F & U & W & C & Risk & Rank & E & Rt & VB & VS & O \\
\midrule
ETTh1 & 1 & \textbf{1.000} & \textbf{1.000} & \textbf{1.000} & \textbf{1.000} & \textbf{1.000} & \textbf{1.000} & \textbf{1.000} & \textbf{1.000} & \textbf{1.000} & \textbf{1.000} & 1.000 \\
 & 2 & \textbf{1.000} & 1.004 & \uline{1.001} & 1.003 & 1.004 & \textbf{1.000} & \textbf{1.000} & \textbf{1.000} & 1.004 & 1.003 & 0.999 \\
 & 4 & 1.000 & 1.000 & \uline{0.989} & 1.014 & 0.999 & 1.005 & \textbf{0.985} & 0.995 & 1.006 & 0.993 & 0.942 \\
 & 8 & 1.000 & 1.001 & \textbf{0.994} & 1.062 & 1.002 & 1.003 & \uline{0.995} & 1.001 & 1.003 & 1.003 & 0.920 \\
 & 16 & 1.000 & \uline{0.997} & \uline{0.997} & 1.068 & \uline{0.997} & 1.005 & \textbf{0.995} & 1.000 & 1.003 & 1.004 & 0.920 \\
 & 80 & 1.000 & 0.988 & \textbf{0.972} & 1.121 & 0.989 & 1.034 & \uline{0.975} & 0.992 & 1.019 & 1.012 & 0.850 \\
\midrule
ETTh2 & 1 & \textbf{1.000} & \textbf{1.000} & \textbf{1.000} & \textbf{1.000} & \textbf{1.000} & \textbf{1.000} & \textbf{1.000} & \textbf{1.000} & \textbf{1.000} & \textbf{1.000} & 1.000 \\
 & 2 & \textbf{1.000} & 1.008 & 1.004 & 1.004 & 1.008 & 1.005 & \textbf{1.000} & \uline{1.002} & \textbf{1.000} & \uline{1.002} & 0.994 \\
 & 4 & 1.000 & 0.986 & \uline{0.983} & \textbf{0.979} & 0.985 & \uline{0.983} & 0.992 & 0.988 & \uline{0.983} & \uline{0.983} & 0.932 \\
 & 8 & 1.000 & 0.978 & \textbf{0.975} & 0.982 & \uline{0.977} & 1.002 & 0.986 & 0.996 & 0.987 & 0.992 & 0.846 \\
 & 16 & 1.000 & 0.987 & 0.982 & \textbf{0.947} & 0.986 & 0.995 & 0.986 & 0.992 & 0.987 & \uline{0.968} & 0.846 \\
 & 80 & 1.000 & 0.990 & \uline{0.976} & 1.036 & 0.991 & 1.126 & 1.012 & \textbf{0.971} & 1.016 & 1.009 & 0.760 \\
\midrule
ETTm1 & 1 & \textbf{1.000} & \textbf{1.000} & \textbf{1.000} & \textbf{1.000} & \textbf{1.000} & \textbf{1.000} & \textbf{1.000} & \textbf{1.000} & \textbf{1.000} & \textbf{1.000} & 1.000 \\
 & 2 & 1.000 & 1.000 & \textbf{0.997} & 1.002 & 0.999 & \uline{0.998} & \uline{0.998} & \uline{0.998} & 1.000 & 0.999 & 0.974 \\
 & 4 & 1.000 & 1.047 & 1.013 & 1.086 & 1.039 & 1.004 & \textbf{0.995} & \uline{0.997} & 1.000 & 1.005 & 0.871 \\
 & 8 & \uline{1.000} & 1.026 & 1.005 & 1.104 & 1.021 & 1.005 & \textbf{0.994} & \uline{1.000} & 1.018 & 1.023 & 0.835 \\
 & 16 & 1.000 & 1.002 & \uline{0.997} & 1.080 & 1.002 & 1.009 & \textbf{0.995} & 1.001 & 1.019 & 1.019 & 0.835 \\
 & 80 & 1.000 & 1.005 & \uline{0.978} & 1.597 & 1.000 & 1.026 & \textbf{0.959} & 0.994 & 1.055 & 0.987 & 0.745 \\
\midrule
ETTm2 & 1 & \textbf{1.000} & \textbf{1.000} & \textbf{1.000} & \textbf{1.000} & \textbf{1.000} & \textbf{1.000} & \textbf{1.000} & \textbf{1.000} & \textbf{1.000} & \textbf{1.000} & 1.000 \\
 & 2 & 1.000 & 1.007 & 1.002 & 1.010 & 1.007 & \textbf{0.997} & 0.999 & \uline{0.998} & 1.002 & \uline{0.998} & 0.985 \\
 & 4 & 1.000 & 0.967 & \uline{0.961} & 0.979 & 0.966 & 0.964 & 0.972 & 0.969 & 0.988 & \textbf{0.960} & 0.878 \\
 & 8 & 1.000 & 0.974 & 0.969 & 1.019 & 0.973 & \textbf{0.959} & 0.981 & 0.977 & 0.963 & \uline{0.961} & 0.827 \\
 & 16 & 1.000 & 0.984 & 0.977 & 1.028 & 0.983 & \textbf{0.958} & 0.981 & 0.986 & 0.965 & \uline{0.961} & 0.826 \\
 & 80 & 1.000 & 0.985 & \textbf{0.964} & 1.218 & 0.986 & 0.988 & 0.980 & 0.985 & 0.972 & \uline{0.970} & 0.766 \\
\midrule
Appliances & 1 & \textbf{1.000} & \textbf{1.000} & \textbf{1.000} & \textbf{1.000} & \textbf{1.000} & \textbf{1.000} & \textbf{1.000} & \textbf{1.000} & \textbf{1.000} & \textbf{1.000} & 1.000 \\
 & 2 & 1.000 & \textbf{0.960} & 0.965 & 1.021 & \uline{0.961} & 1.002 & 0.967 & 0.973 & 1.009 & 0.982 & 0.936 \\
 & 4 & 1.000 & 0.960 & \textbf{0.953} & 1.027 & 0.960 & 0.997 & \uline{0.956} & 0.969 & 1.012 & 0.984 & 0.920 \\
 & 8 & 1.000 & 0.963 & \uline{0.953} & 1.007 & 0.963 & 0.989 & \textbf{0.947} & 0.979 & 1.015 & 0.982 & 0.883 \\
 & 16 & 1.000 & 0.976 & \uline{0.963} & 0.995 & 0.974 & 0.992 & \textbf{0.947} & 0.980 & 1.011 & 0.979 & 0.883 \\
 & 80 & 1.000 & 0.970 & \textbf{0.954} & 1.273 & 0.969 & 1.002 & \uline{0.957} & 0.978 & 1.006 & 0.958 & 0.845 \\
\midrule
BeijingWeather & 1 & \textbf{1.000} & \textbf{1.000} & \textbf{1.000} & \textbf{1.000} & \textbf{1.000} & \textbf{1.000} & \textbf{1.000} & \textbf{1.000} & \textbf{1.000} & \textbf{1.000} & 1.000 \\
 & 2 & 1.000 & 1.002 & \uline{0.997} & 1.002 & 1.001 & 1.001 & \textbf{0.996} & 0.998 & 1.000 & 1.002 & 0.979 \\
 & 4 & 1.000 & \uline{0.981} & \textbf{0.980} & 0.989 & \textbf{0.980} & 0.986 & \textbf{0.980} & 0.994 & 0.984 & 0.987 & 0.946 \\
 & 8 & 1.000 & 0.979 & \uline{0.976} & 0.989 & 0.977 & \textbf{0.967} & \uline{0.976} & 0.986 & 0.984 & 0.978 & 0.919 \\
 & 16 & 1.000 & 0.987 & 0.982 & \uline{0.974} & 0.986 & \textbf{0.968} & 0.976 & 0.992 & 0.984 & \uline{0.974} & 0.918 \\
 & 80 & 1.000 & 0.989 & 0.972 & 1.000 & 0.985 & \textbf{0.949} & \uline{0.969} & 0.995 & 1.004 & \textbf{0.949} & 0.864 \\
\bottomrule
\end{tabular}
\end{table}

\clearpage
\begin{table}[H]
\centering
\caption{Sundial-base: dataset-specific candidate-budget relative MSE underlying Figure~\ref{fig:scaling}. Datasets are ETTh1, ETTh2, ETTm1, ETTm2, Appliances, and BeijingWeather. Every entry averages four horizons and three seeds geometrically; candidate budget is $K$. Column definitions are given in Appendix~\ref{app:figure_breakdowns}. O is the unavailable test-label Oracle.}\label{tab:figure3_model5}
\fontsize{8.3}{10.2}\selectfont
\setlength{\tabcolsep}{2.2pt}
\renewcommand{\arraystretch}{1.04}
\begin{tabular}{lrrrrrrrrrrrr}
\toprule
Dataset & $K$ & F & U & W & C & Risk & Rank & E & Rt & VB & VS & O \\
\midrule
ETTh1 & 1 & \textbf{1.000} & \textbf{1.000} & \textbf{1.000} & \textbf{1.000} & \textbf{1.000} & \textbf{1.000} & \textbf{1.000} & \textbf{1.000} & \textbf{1.000} & \textbf{1.000} & 1.000 \\
 & 2 & \textbf{1.000} & 1.005 & \uline{1.002} & 1.009 & 1.005 & 1.003 & \textbf{1.000} & 1.004 & 1.010 & 1.009 & 0.994 \\
 & 4 & 1.000 & 0.990 & \textbf{0.984} & 1.021 & 0.989 & 1.007 & \uline{0.987} & 0.992 & 0.997 & 0.990 & 0.935 \\
 & 8 & 1.000 & 0.996 & 0.999 & 1.121 & 0.999 & 1.008 & 1.000 & \uline{0.993} & \textbf{0.990} & 0.995 & 0.897 \\
 & 16 & 1.000 & \uline{0.992} & 0.994 & 1.119 & \uline{0.992} & 1.032 & 1.000 & 0.993 & \textbf{0.990} & 1.010 & 0.897 \\
 & 80 & 1.000 & 0.985 & 0.978 & 1.251 & 0.987 & 1.051 & \textbf{0.970} & 0.995 & 0.987 & \uline{0.977} & 0.828 \\
\midrule
ETTh2 & 1 & \textbf{1.000} & \textbf{1.000} & \textbf{1.000} & \textbf{1.000} & \textbf{1.000} & \textbf{1.000} & \textbf{1.000} & \textbf{1.000} & \textbf{1.000} & \textbf{1.000} & 1.000 \\
 & 2 & 1.000 & \uline{0.998} & 1.001 & 1.005 & \uline{0.998} & 1.002 & 1.010 & \textbf{0.997} & 1.000 & \uline{0.998} & 0.982 \\
 & 4 & 1.000 & 0.985 & 0.981 & \textbf{0.965} & 0.983 & \uline{0.973} & 0.988 & 0.992 & 0.994 & 0.985 & 0.904 \\
 & 8 & 1.000 & 0.974 & \uline{0.973} & \textbf{0.972} & \textbf{0.972} & 1.006 & 0.974 & 0.988 & 0.983 & 1.000 & 0.822 \\
 & 16 & 1.000 & 0.985 & 0.980 & \textbf{0.942} & 0.984 & 0.978 & \uline{0.974} & 0.994 & 0.983 & 0.987 & 0.822 \\
 & 80 & 1.000 & 0.987 & \textbf{0.972} & 1.054 & 0.986 & 1.032 & 0.983 & \uline{0.982} & 0.990 & 0.991 & 0.734 \\
\midrule
ETTm1 & 1 & \textbf{1.000} & \textbf{1.000} & \textbf{1.000} & \textbf{1.000} & \textbf{1.000} & \textbf{1.000} & \textbf{1.000} & \textbf{1.000} & \textbf{1.000} & \textbf{1.000} & 1.000 \\
 & 2 & \textbf{1.000} & 1.010 & 1.004 & 1.010 & 1.009 & \uline{1.001} & \textbf{1.000} & 1.002 & \textbf{1.000} & 1.002 & 0.992 \\
 & 4 & 1.000 & 0.996 & 0.991 & 1.005 & 0.994 & 1.000 & 0.985 & \uline{0.983} & 1.000 & \textbf{0.977} & 0.882 \\
 & 8 & 1.000 & 0.993 & \uline{0.988} & 1.108 & 0.990 & 1.000 & \textbf{0.984} & 0.992 & 1.005 & \uline{0.988} & 0.841 \\
 & 16 & 1.000 & 0.994 & \uline{0.989} & 1.130 & 0.993 & 1.001 & \textbf{0.984} & 0.993 & 1.003 & \textbf{0.984} & 0.841 \\
 & 80 & 1.000 & 0.981 & 0.963 & 1.827 & 0.987 & 1.016 & \textbf{0.954} & 0.989 & 1.005 & \uline{0.962} & 0.717 \\
\midrule
ETTm2 & 1 & \textbf{1.000} & \textbf{1.000} & \textbf{1.000} & \textbf{1.000} & \textbf{1.000} & \textbf{1.000} & \textbf{1.000} & \textbf{1.000} & \textbf{1.000} & \textbf{1.000} & 1.000 \\
 & 2 & \uline{1.000} & \uline{1.000} & 1.005 & 1.006 & \uline{1.000} & 1.004 & 1.018 & \textbf{0.998} & \uline{1.000} & \uline{1.000} & 0.983 \\
 & 4 & 1.000 & 0.984 & \uline{0.978} & 1.030 & 0.983 & 1.005 & 0.984 & 0.987 & 0.989 & \textbf{0.976} & 0.919 \\
 & 8 & 1.000 & 0.976 & \textbf{0.969} & 1.042 & \uline{0.974} & 0.995 & 0.986 & 0.977 & 0.993 & 0.993 & 0.842 \\
 & 16 & 1.000 & 0.982 & \textbf{0.975} & 1.038 & \uline{0.981} & 0.984 & 0.986 & 0.983 & 0.993 & 0.998 & 0.840 \\
 & 80 & 1.000 & 0.985 & \textbf{0.963} & 1.265 & 0.984 & 1.081 & \uline{0.974} & 0.992 & 0.998 & 0.992 & 0.794 \\
\midrule
Appliances & 1 & \textbf{1.000} & \textbf{1.000} & \textbf{1.000} & \textbf{1.000} & \textbf{1.000} & \textbf{1.000} & \textbf{1.000} & \textbf{1.000} & \textbf{1.000} & \textbf{1.000} & 1.000 \\
 & 2 & \textbf{1.000} & 1.034 & 1.014 & 1.026 & 1.035 & 1.021 & \uline{1.007} & 1.020 & 1.028 & 1.021 & 0.988 \\
 & 4 & 1.000 & 0.966 & \uline{0.962} & 0.997 & 0.966 & 0.979 & \uline{0.962} & \uline{0.962} & 1.020 & \textbf{0.958} & 0.922 \\
 & 8 & 1.000 & 0.972 & \uline{0.961} & 1.007 & 0.972 & 0.987 & \textbf{0.955} & 0.964 & 1.087 & 1.054 & 0.888 \\
 & 16 & 1.000 & 0.988 & \uline{0.975} & 1.003 & 0.987 & 0.987 & \textbf{0.955} & \uline{0.975} & 1.087 & 1.044 & 0.884 \\
 & 80 & 1.000 & 0.994 & \uline{0.940} & 1.445 & 0.993 & 1.035 & \textbf{0.928} & 0.978 & 1.036 & 0.961 & 0.834 \\
\midrule
BeijingWeather & 1 & \textbf{1.000} & \textbf{1.000} & \textbf{1.000} & \textbf{1.000} & \textbf{1.000} & \textbf{1.000} & \textbf{1.000} & \textbf{1.000} & \textbf{1.000} & \textbf{1.000} & 1.000 \\
 & 2 & 1.000 & 0.991 & \uline{0.988} & \textbf{0.987} & 0.990 & 0.990 & 0.989 & 0.990 & 1.000 & 0.990 & 0.974 \\
 & 4 & 1.000 & 0.953 & \uline{0.946} & 0.956 & 0.950 & 0.956 & \textbf{0.940} & 0.954 & 0.954 & 0.951 & 0.915 \\
 & 8 & 1.000 & 0.957 & 0.952 & 0.967 & 0.953 & 0.950 & \textbf{0.942} & 0.956 & 0.963 & \uline{0.949} & 0.896 \\
 & 16 & 1.000 & 0.973 & 0.963 & 0.969 & 0.968 & \uline{0.949} & \textbf{0.942} & 0.969 & 0.963 & \uline{0.949} & 0.895 \\
 & 80 & 1.000 & 0.978 & \uline{0.955} & 1.048 & 0.973 & \textbf{0.952} & 0.965 & 0.972 & 0.993 & 0.965 & 0.868 \\
\bottomrule
\end{tabular}
\end{table}

\clearpage
\begin{table}[H]
\centering
\caption{Chronos-2: dataset-specific candidate-budget relative MSE underlying Figure~\ref{fig:scaling}. Datasets are ETTh1, ETTh2, ETTm1, ETTm2, Appliances, and BeijingWeather. Every entry averages four horizons and three seeds geometrically; candidate budget is $K$. Column definitions are given in Appendix~\ref{app:figure_breakdowns}. O is the unavailable test-label Oracle.}\label{tab:figure3_model6}
\fontsize{8.3}{10.2}\selectfont
\setlength{\tabcolsep}{2.2pt}
\renewcommand{\arraystretch}{1.04}
\begin{tabular}{lrrrrrrrrrrrr}
\toprule
Dataset & $K$ & F & U & W & C & Risk & Rank & E & Rt & VB & VS & O \\
\midrule
ETTh1 & 1 & \textbf{1.000} & \textbf{1.000} & \textbf{1.000} & \textbf{1.000} & \textbf{1.000} & \textbf{1.000} & \textbf{1.000} & \textbf{1.000} & \textbf{1.000} & \textbf{1.000} & 1.000 \\
 & 2 & 1.000 & \uline{0.996} & \textbf{0.995} & 1.000 & \uline{0.996} & \textbf{0.995} & \textbf{0.995} & 0.997 & 1.007 & 1.001 & 0.982 \\
 & 4 & 1.000 & 0.935 & \uline{0.932} & 0.959 & 0.934 & 0.965 & \textbf{0.930} & 0.950 & 1.024 & 0.953 & 0.853 \\
 & 8 & 1.000 & 0.960 & \uline{0.947} & 0.986 & 0.956 & 0.975 & \textbf{0.933} & 0.968 & 1.017 & 0.964 & 0.846 \\
 & 16 & 1.000 & 0.973 & \uline{0.957} & 0.986 & 0.971 & 0.977 & \textbf{0.933} & 0.986 & 1.017 & 0.980 & 0.845 \\
 & 80 & 1.000 & 0.961 & 0.883 & 1.155 & 0.968 & 0.979 & \textbf{0.839} & 0.978 & 0.975 & \uline{0.869} & 0.709 \\
\midrule
ETTh2 & 1 & \textbf{1.000} & \textbf{1.000} & \textbf{1.000} & \textbf{1.000} & \textbf{1.000} & \textbf{1.000} & \textbf{1.000} & \textbf{1.000} & \textbf{1.000} & \textbf{1.000} & 1.000 \\
 & 2 & 1.000 & 0.990 & \uline{0.989} & \textbf{0.987} & 0.990 & 1.005 & 0.990 & \uline{0.989} & 0.992 & 0.992 & 0.965 \\
 & 4 & 1.000 & 0.928 & 0.922 & \uline{0.918} & 0.927 & 0.956 & \textbf{0.913} & 0.931 & 0.923 & 0.931 & 0.835 \\
 & 8 & 1.000 & 0.938 & 0.929 & \uline{0.911} & 0.940 & 0.952 & \textbf{0.907} & 0.944 & 0.956 & 0.947 & 0.817 \\
 & 16 & 1.000 & 0.963 & 0.951 & \uline{0.917} & 0.963 & 0.951 & \textbf{0.909} & 0.966 & 0.956 & 0.942 & 0.817 \\
 & 80 & 1.000 & 0.970 & \uline{0.950} & 1.051 & 0.971 & 1.130 & \textbf{0.915} & 0.974 & 0.965 & 0.953 & 0.725 \\
\midrule
ETTm1 & 1 & \textbf{1.000} & \textbf{1.000} & \textbf{1.000} & \textbf{1.000} & \textbf{1.000} & \textbf{1.000} & \textbf{1.000} & \textbf{1.000} & \textbf{1.000} & \textbf{1.000} & 1.000 \\
 & 2 & \textbf{1.000} & 1.006 & \uline{1.002} & 1.017 & 1.005 & 1.014 & \uline{1.002} & 1.005 & 1.004 & 1.011 & 0.968 \\
 & 4 & 1.000 & \textbf{0.932} & 0.937 & 0.962 & \uline{0.933} & 0.964 & 0.934 & \uline{0.933} & 0.984 & \uline{0.933} & 0.814 \\
 & 8 & 1.000 & 0.947 & \uline{0.938} & 1.017 & 0.945 & 0.977 & \textbf{0.935} & 0.941 & 0.994 & \textbf{0.935} & 0.787 \\
 & 16 & 1.000 & 0.968 & \uline{0.951} & 1.037 & 0.968 & 0.972 & \textbf{0.935} & 0.964 & 0.992 & \textbf{0.935} & 0.786 \\
 & 80 & 1.000 & 0.945 & \uline{0.885} & 1.945 & 0.961 & 0.988 & \textbf{0.878} & 0.958 & 0.907 & 0.899 & 0.640 \\
\midrule
ETTm2 & 1 & \textbf{1.000} & \textbf{1.000} & \textbf{1.000} & \textbf{1.000} & \textbf{1.000} & \textbf{1.000} & \textbf{1.000} & \textbf{1.000} & \textbf{1.000} & \textbf{1.000} & 1.000 \\
 & 2 & 1.000 & 0.999 & \uline{0.998} & 1.013 & 0.999 & 1.007 & 1.002 & \textbf{0.997} & 1.002 & \textbf{0.997} & 0.972 \\
 & 4 & 1.000 & \uline{0.951} & 0.952 & 0.991 & \textbf{0.950} & 0.955 & 0.962 & \textbf{0.950} & 0.993 & \textbf{0.950} & 0.875 \\
 & 8 & 1.000 & 0.960 & \textbf{0.957} & 1.006 & \uline{0.959} & 0.961 & 0.961 & \uline{0.959} & 0.994 & 0.961 & 0.839 \\
 & 16 & 1.000 & 0.975 & 0.967 & 1.015 & 0.974 & \uline{0.962} & \textbf{0.961} & 0.972 & 0.992 & 0.966 & 0.838 \\
 & 80 & 1.000 & 0.975 & \textbf{0.960} & 1.526 & 0.977 & 1.029 & \uline{0.965} & 0.986 & 1.006 & 0.999 & 0.785 \\
\midrule
Appliances & 1 & \textbf{1.000} & \textbf{1.000} & \textbf{1.000} & \textbf{1.000} & \textbf{1.000} & \textbf{1.000} & \textbf{1.000} & \textbf{1.000} & \textbf{1.000} & \textbf{1.000} & 1.000 \\
 & 2 & 1.000 & \uline{0.954} & 0.956 & 0.956 & 0.956 & 0.980 & 0.957 & 0.976 & 1.000 & \textbf{0.948} & 0.916 \\
 & 4 & 1.000 & 0.911 & \uline{0.906} & 0.933 & 0.909 & 0.969 & \textbf{0.905} & 0.920 & 1.010 & 0.914 & 0.848 \\
 & 8 & 1.000 & 0.918 & \uline{0.908} & 0.950 & 0.918 & 0.936 & \textbf{0.899} & 0.923 & 1.054 & 0.941 & 0.833 \\
 & 16 & 1.000 & 0.945 & \uline{0.927} & 0.947 & 0.942 & 0.950 & \textbf{0.899} & 0.948 & 1.054 & 0.933 & 0.830 \\
 & 80 & 1.000 & 0.935 & \uline{0.898} & 1.511 & 0.937 & 0.972 & \textbf{0.887} & 0.946 & 0.959 & 0.920 & 0.780 \\
\midrule
BeijingWeather & 1 & \textbf{1.000} & \textbf{1.000} & \textbf{1.000} & \textbf{1.000} & \textbf{1.000} & \textbf{1.000} & \textbf{1.000} & \textbf{1.000} & \textbf{1.000} & \textbf{1.000} & 1.000 \\
 & 2 & 1.000 & 0.868 & 0.841 & 0.984 & 0.849 & \textbf{0.812} & \uline{0.819} & 0.866 & 0.834 & 0.843 & 0.789 \\
 & 4 & 1.000 & 0.821 & 0.793 & 0.783 & 0.788 & \textbf{0.762} & \uline{0.776} & 0.823 & 0.855 & 0.781 & 0.730 \\
 & 8 & 1.000 & 0.845 & 0.805 & 0.892 & 0.796 & \textbf{0.756} & \uline{0.779} & 0.844 & 0.828 & 0.783 & 0.720 \\
 & 16 & 1.000 & 0.851 & 0.834 & 0.875 & 0.824 & \textbf{0.752} & 0.814 & 0.852 & 0.820 & \uline{0.810} & 0.719 \\
 & 80 & 1.000 & 0.854 & 0.803 & 0.888 & 0.829 & \textbf{0.770} & \uline{0.784} & 0.853 & 0.822 & 0.815 & 0.700 \\
\bottomrule
\end{tabular}
\end{table}

\section{Candidate-informed UQ}\label{app:uncertainty}

A small network receives five common observable features: log-one-plus context standard deviation, log-one-plus RMS first difference of context, log-one-plus square root of the identity's retrospective MSE, relative forecast lead, and log-one-plus absolute displacement of the forecast from the last observation. Retrospective MSE averages the already observed horizon and channels. Seven additional features describe current candidates: log-one-plus standard deviation within each of the 3 generator families, standard deviation of the 3 family means, family-balanced RMS displacement from $\mu$, and the signed log-one-plus displacements of the empirical 10th and 90th candidate percentiles from $\mu$.

\subsection{Comparators and information access}
All methods use the same targets and calibration origins. The selected TTC point predictor is shared across TTC interval methods, except that raw candidate quantiles and CQR use their own central quantile for scoring.
\begin{itemize}
\item \textbf{Frozen Gaussian} centers intervals at Frozen and uses Gaussian quantiles times per-lead/channel residual RMS estimated on the selection block.
\item \textbf{Frozen/TTC split CP} use absolute calibration residuals around their respective point forecasts, giving a constant half-width for each lead and channel.
\item \textbf{Raw candidate quantiles} use empirical lower/upper candidate percentiles without calibration. \textbf{Candidate-quantile CP} expands those endpoints using the corrected quantile of $\max(\ell_i-Y_i,Y_i-u_i,0)$ on calibration.
\item \textbf{History-scale CP} matches the scale head and training objective but uses only the five common features. \textbf{Candidate-spread CP} uses $s=\sqrt{d^2+b^2}$, where $b$ is the greater of 0.01 and 5\% of selection-block residual RMS, separately for each lead and channel. \textbf{Candidate-scale CP} uses the twelve-feature learned scale.
\item \textbf{CQR (non-shrinking)} fits a $12\!\to\!32\!\to\!7$ residual-quantile MLP with pinball loss at levels 0.025, 0.05, 0.10, 0.50, 0.90, 0.95, and 0.975. Outputs are sorted before the loss to prevent crossing. Optimization matches the scale heads. Calibration uses $\max(\ell_i-Y_i,Y_i-u_i,0)$, a conservative non-shrinking variant of CQR \citep{romano2019conformalized}, with the same candidate features available to the proposed scale head.
\item \textbf{Frozen/TTC rolling CP} maintain the most recent $n$ complete residual horizons, initially the calibration block, and replace the oldest after each observed test horizon. \textbf{Candidate-scale rolling CP} does the same with normalized residuals and a fixed learned scale head.
\item \textbf{TTC bounded ACI} combines rolling residuals with $\alpha_{t+1,h,c}=\Pi_{[1/(n+1),\,0.5]}[\alpha_{t,h,c}+0.01(\alpha-\mathbf 1\{Y_{t,h,c}\notin I_{t,h,c}\})]$, initialized at the desired $\alpha$. This explicit finite-width, projected adaptation of ACI \citep{gibbs2021adaptive} does not inherit the unrestricted ACI guarantee. Adaptive updates occur only after the current horizon has become observed.
\end{itemize}
Fixed-calibration methods never use test labels for interval updates. Online comparators use previously completed test horizons as historical feedback, never the current or future horizon. Generator, point-aggregator, and uncertainty-head parameters remain frozen in both settings. Sequential residual modeling such as SPCI \citep{xu2023sequential} is related work, not a reproduced comparator here; neither native probabilistic foundation-model interfaces nor bootstrap-refitted backbones are evaluated.

\subsection{Metrics and measured comparison}
We evaluate pointwise coverage at nominal 80\%, 90\%, and 95\%, interval width, and the interval score
\begin{equation}
 \mathrm{IS}_{\alpha}(\ell,u;y)=(u-\ell)+\frac{2}{\alpha}(\ell-y)_+
 +\frac{2}{\alpha}(y-u)_+.
\end{equation}
Weighted interval score (WIS) combines these three intervals with the supplied central prediction $m$:
\begin{equation}
 \mathrm{WIS}=\frac{0.5|y-m|+\sum_{\alpha\in\{0.2,0.1,0.05\}}
 (\alpha/2)\mathrm{IS}_{\alpha}}{3.5}.
\end{equation}
This follows interval-format forecast evaluation \citep{bracher2021evaluating}. The central prediction is the fixed point forecast for scale/residual methods and the 0.5 quantile for candidate-quantile/CQR methods. 
Table~\ref{tab:uq_datasets} reports model--dataset variation rather than treating aggregate coverage as uniform validity.

Simultaneous per-channel trajectory bands further illustrate the coverage--width tradeoff. Candidate max-score CP covers 90.1\% of complete trajectories at mean width 5.875, compared with 88.5\% and 4.916 for TTC max-residual CP. Higher trajectory coverage therefore comes with wider bands. The extension supports calibrated interval construction from TTC4TS candidates, while showing that candidate diversity alone and a more complex scale head do not ensure better uncertainty estimates.

\begin{table}[H]
\centering
\caption{Fixed-calibration uncertainty by model and dataset at horizon 96 on ETTh1, ETTh2, ETTm1, ETTm2, Appliances, and BeijingWeather. Each pair gives 90\% pointwise coverage (C, percent) and geometric relative WIS (R) over seeds 0--2. WIS uses 80/90/95\% intervals and is normalized to Frozen split CP in each cell. CQR uses the declared non-shrinking correction.}\label{tab:uq_datasets}
\fontsize{8.4}{10.4}\selectfont
\setlength{\tabcolsep}{3pt}
\begin{tabular}{lrrrrrrrr}
\toprule
& \multicolumn{2}{c}{Frozen split} & \multicolumn{2}{c}{TTC split} & \multicolumn{2}{c}{Candidate-scale} & \multicolumn{2}{c}{CQR} \\
Dataset & C & R & C & R & C & R & C & R \\
\midrule
\multicolumn{9}{l}{\textbf{Chronos-Bolt-small}} \\
ETTh1 & \uline{94.6} & \uline{1.000} & 94.9 & \textbf{0.967} & \textbf{93.4} & 1.037 & 95.5 & 1.020 \\
ETTh2 & 86.2 & \textbf{1.000} & 86.6 & \uline{1.005} & \textbf{87.7} & 1.072 & \uline{92.4} & 1.068 \\
ETTm1 & \uline{92.8} & \uline{1.000} & 93.4 & \textbf{0.958} & \textbf{92.1} & 1.023 & 94.8 & 1.110 \\
ETTm2 & 87.8 & \uline{1.000} & \uline{87.9} & \textbf{0.994} & \uline{87.9} & 1.026 & \textbf{90.8} & 1.030 \\
Appliances & 87.7 & \uline{1.000} & 86.1 & \textbf{0.998} & \uline{87.8} & 1.006 & \textbf{91.3} & 1.020 \\
BeijingWeather & \uline{88.0} & 1.000 & \textbf{88.7} & 0.988 & \textbf{88.7} & \textbf{0.948} & 94.0 & \uline{0.976} \\
\midrule
\multicolumn{9}{l}{\textbf{Chronos-Bolt-base}} \\
ETTh1 & 95.0 & \uline{1.000} & \uline{94.9} & \textbf{0.961} & \textbf{93.5} & 1.016 & 95.9 & 1.004 \\
ETTh2 & 87.3 & \uline{1.000} & \uline{87.6} & \textbf{0.992} & \textbf{87.8} & 1.088 & \textbf{92.2} & 1.065 \\
ETTm1 & \uline{92.9} & \uline{1.000} & 93.2 & \textbf{0.979} & \textbf{92.4} & 1.036 & 94.4 & 1.129 \\
ETTm2 & 87.9 & \uline{1.000} & \uline{88.0} & \textbf{0.992} & 87.6 & 1.029 & \textbf{90.5} & 1.038 \\
Appliances & \uline{87.3} & \textbf{1.000} & 86.1 & \uline{1.001} & 87.1 & 1.016 & \textbf{91.4} & 1.036 \\
BeijingWeather & 87.8 & 1.000 & \uline{88.8} & 0.998 & \textbf{89.5} & \textbf{0.956} & 93.5 & \uline{0.957} \\
\midrule
\multicolumn{9}{l}{\textbf{TimesFM-2.0}} \\
ETTh1 & 95.9 & \uline{1.000} & \uline{95.1} & \textbf{0.976} & \textbf{94.7} & 1.009 & 95.9 & 1.017 \\
ETTh2 & 84.8 & \textbf{1.000} & 84.3 & \uline{1.020} & \uline{86.1} & 1.032 & \textbf{90.8} & 1.090 \\
ETTm1 & \uline{94.0} & 1.000 & 94.3 & \textbf{0.939} & \textbf{92.8} & \uline{0.957} & 94.4 & 1.011 \\
ETTm2 & \textbf{88.2} & \uline{1.000} & \textbf{88.2} & \textbf{0.967} & \uline{87.5} & 1.006 & \textbf{91.8} & 1.009 \\
Appliances & 86.7 & 1.000 & \uline{87.0} & \textbf{0.962} & 86.9 & \uline{0.985} & \textbf{91.2} & 0.994 \\
BeijingWeather & \uline{89.5} & \textbf{1.000} & \uline{89.5} & \uline{1.035} & \textbf{90.4} & 1.056 & 93.8 & 1.036 \\
\midrule
\multicolumn{9}{l}{\textbf{Timer-base}} \\
ETTh1 & 95.3 & 1.000 & 95.5 & \textbf{0.976} & \textbf{93.4} & \uline{0.987} & \uline{95.1} & 1.015 \\
ETTh2 & 86.8 & \uline{1.000} & 86.9 & \textbf{0.990} & \uline{87.5} & 1.089 & \textbf{91.3} & 1.086 \\
ETTm1 & \uline{93.9} & \uline{1.000} & 94.0 & \textbf{0.990} & \textbf{92.3} & 1.027 & 94.6 & 1.095 \\
ETTm2 & \uline{87.6} & \uline{1.000} & 87.3 & \textbf{0.983} & 86.3 & 1.053 & \textbf{90.7} & 1.020 \\
Appliances & 86.9 & 1.000 & 87.0 & \textbf{0.994} & \uline{87.9} & \uline{0.996} & \textbf{91.7} & 1.010 \\
BeijingWeather & 88.1 & \uline{1.000} & \uline{88.4} & 1.026 & \textbf{89.1} & 1.020 & 94.1 & \textbf{0.974} \\
\midrule
\multicolumn{9}{l}{\textbf{Sundial-base}} \\
ETTh1 & 95.2 & 1.000 & \uline{95.0} & \textbf{0.982} & \textbf{94.9} & \uline{0.984} & 95.4 & 1.024 \\
ETTh2 & 87.0 & \textbf{1.000} & 87.0 & \uline{1.007} & \textbf{87.6} & 1.114 & \uline{92.7} & 1.089 \\
ETTm1 & \uline{94.5} & \uline{1.000} & 94.7 & \textbf{0.996} & \textbf{93.4} & 1.015 & 94.7 & 1.079 \\
ETTm2 & \textbf{89.0} & \uline{1.000} & \uline{88.3} & \textbf{0.994} & 88.0 & 1.047 & 92.2 & 1.032 \\
Appliances & \textbf{89.0} & 1.000 & 87.4 & \textbf{0.973} & 88.1 & \uline{0.983} & \uline{91.8} & 0.986 \\
BeijingWeather & 88.5 & 1.000 & \uline{88.6} & 0.915 & \textbf{88.8} & \uline{0.904} & 94.0 & \textbf{0.902} \\
\midrule
\multicolumn{9}{l}{\textbf{Chronos-2}} \\
ETTh1 & \uline{95.2} & 1.000 & \textbf{94.9} & \textbf{0.943} & \uline{95.2} & \uline{0.984} & 96.1 & 0.986 \\
ETTh2 & \uline{86.4} & \textbf{1.000} & 85.5 & \uline{1.037} & 85.8 & 1.112 & \textbf{91.9} & 1.058 \\
ETTm1 & \uline{93.8} & \uline{1.000} & 93.9 & \textbf{0.973} & \textbf{92.5} & 1.031 & 94.7 & 1.090 \\
ETTm2 & 87.8 & \uline{1.000} & \uline{88.4} & \textbf{0.997} & 88.1 & 1.040 & \textbf{91.0} & 1.043 \\
Appliances & 87.6 & 1.000 & \uline{87.7} & \textbf{0.972} & \textbf{88.7} & \uline{0.975} & 92.9 & 1.001 \\
BeijingWeather & 88.1 & \uline{1.000} & \uline{88.9} & 1.001 & \textbf{91.0} & 1.010 & 93.6 & \textbf{0.993} \\
\bottomrule
\end{tabular}

{\footnotesize\emph{Note.}  Coverage ranks minimize distance to 90 percent; WIS ranks minimize error.\par}
\end{table}


A marginal 90\% interval at each step does not imply 90\% coverage of an entire trajectory. For a simultaneous band within each channel, we calibrate $R_{i,c}^{\max}=\max_{h\leq96}|Y_{i,h,c}-\mu_{i,h,c}|/s_{i,h,c}$, then use its corrected 90th percentile for all leads. Table~\ref{tab:uq_simultaneous} compares constant-scale and candidate-scale versions. These bands target each channel's trajectory separately, not the joint event across every channel, and retain the temporal-dependence limitation of the marginal experiment.

\begin{table}[H]
\centering
\caption{Simultaneous 90\% bands for each channel's full 96-step trajectory, over ETTh1, ETTh2, ETTm1, ETTm2, Appliances, and BeijingWeather and seeds 0--2. Calibration uses the maximum normalized error across the horizon, separately for each channel. C is whole-trajectory coverage (percent), and W is mean width in fitting-standardized units. These bands do not target simultaneous coverage across all channels.}\label{tab:uq_simultaneous}
\fontsize{9}{10.4}\selectfont
\setlength{\tabcolsep}{5pt}
\begin{tabular}{lrrrr}
\toprule
& \multicolumn{2}{c}{TTC max-residual CP} & \multicolumn{2}{c}{Candidate max-score CP} \\
Model & C & W & C & W \\
\midrule
Chronos-Bolt-small & \uline{88.7} & 4.856 & \textbf{90.2} & 5.848 \\
Chronos-Bolt-base & \uline{88.3} & 4.855 & \textbf{90.1} & 5.879 \\
TimesFM-2.0 & \uline{89.5} & 5.003 & \textbf{90.0} & 5.844 \\
Timer-base & \uline{88.3} & 5.013 & \textbf{90.0} & 6.025 \\
Sundial-base & \uline{88.1} & 4.912 & \textbf{90.3} & 5.821 \\
Chronos-2 & \uline{88.1} & 4.856 & \textbf{90.2} & 5.834 \\
\bottomrule
\end{tabular}

{\footnotesize\emph{Note.} Best: bold; second-best: double underline. Ties use displayed precision. Ranks minimize distance to nominal 90 percent coverage; width is descriptive.\par}
\end{table}

\section{Illustrative successful forecasts}\label{app:forecast_examples}
\begin{figure}[H]
\centering\includegraphics[width=\linewidth]{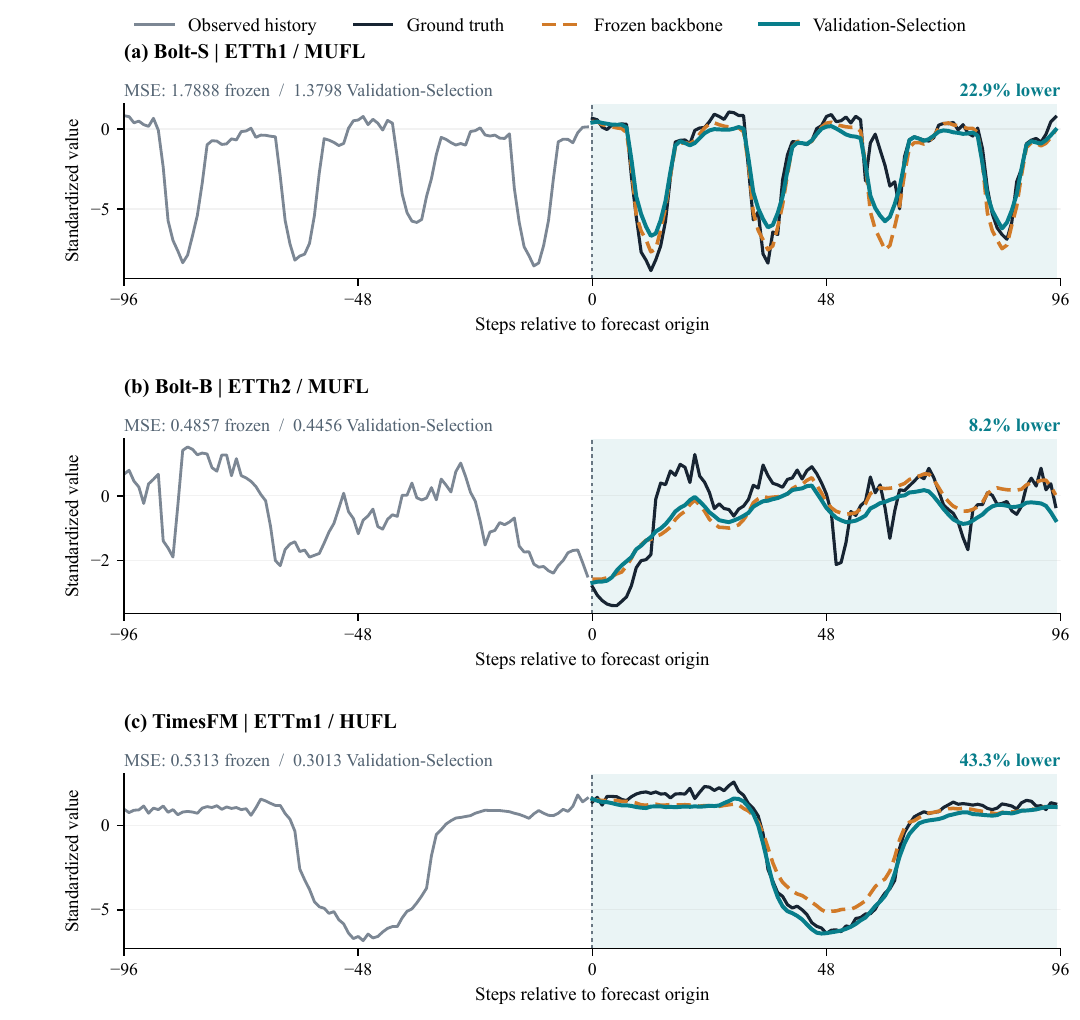}
\caption{Saved successful forecasts for (a) Chronos-Bolt-small on ETTh1 (MUFL), (b) Chronos-Bolt-base on ETTh2 (MUFL), and (c) TimesFM-2.0 on ETTm1 (HUFL), at horizon 96 and seed 0. Gray is observed history, black is ground truth, orange dashed is Frozen, and teal is validation-selected TTC4TS. Shading marks the forecast interval. MSE and percentage reductions refer only to the plotted channel and horizon. These post-hoc illustrations follow Appendix~\ref{app:forecast_examples}'s selection rule.}\label{fig:forecast_examples1}
\end{figure}

\begin{figure}[H]
\centering\includegraphics[width=\linewidth]{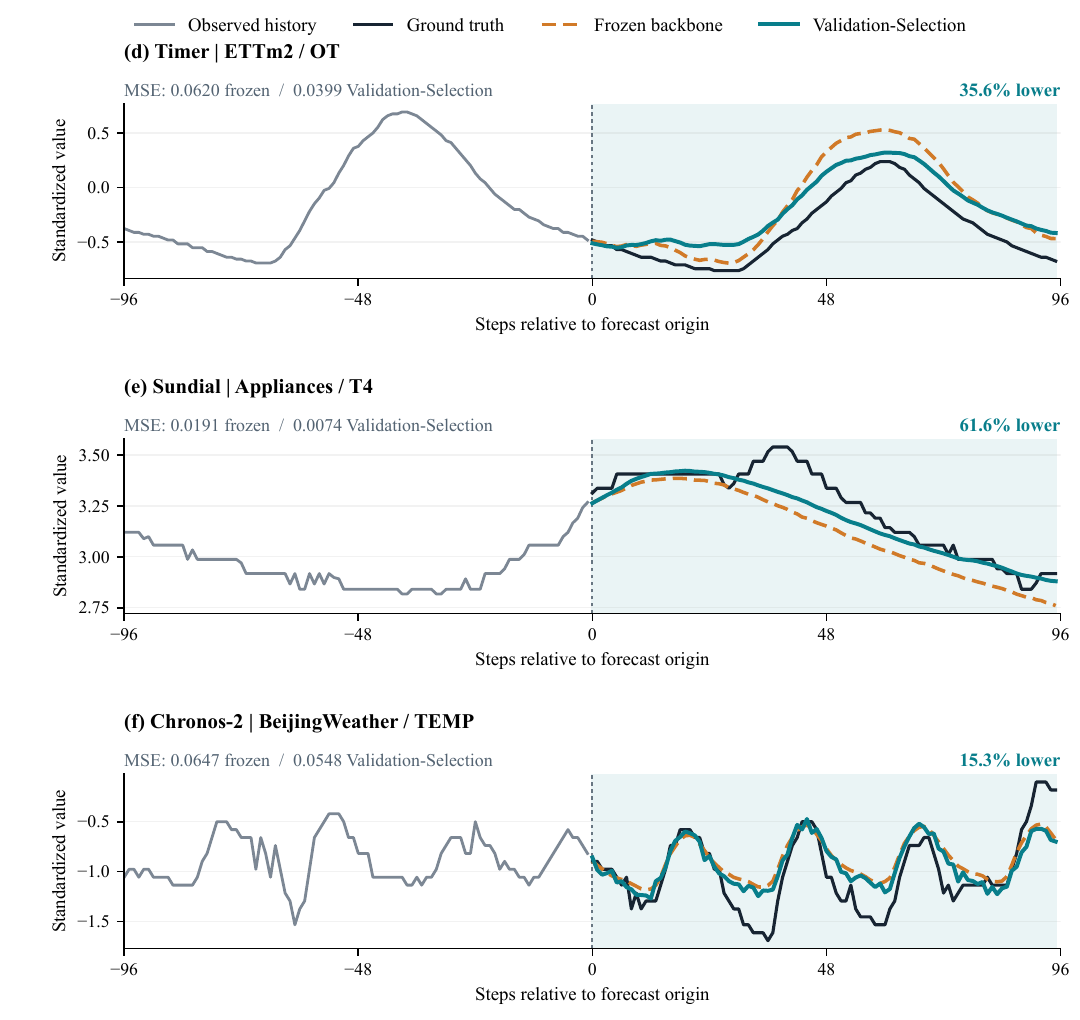}
\caption{Saved successful forecasts for (d) Timer-base on ETTm2 (OT), (e) Sundial-base on Appliances (T4), and (f) Chronos-2 on BeijingWeather (TEMP), at horizon 96 and seed 0. Each panel compares the same observed history and future targets for Frozen and validation-selected TTC4TS. Colors, shading and error annotations follow Figure~\ref{fig:forecast_examples1}. These are post-hoc success illustrations, not representative estimates of average gains.}\label{fig:forecast_examples2}
\end{figure}

Figures~\ref{fig:forecast_examples1} and~\ref{fig:forecast_examples2} visualize saved forecasts from six fixed model--dataset pairs, one for each checkpoint and dataset, at $H=96$ and seed 0. Each panel shows 96 observed steps and the next 96 true and predicted values. The curves come from the revised campaign's saved Frozen and validation-selected TTC4TS predictions, and no fitting or forecast regeneration was performed for these plots.

These are explicitly post-hoc success illustrations. Within each fixed pair, eligible origin--channel combinations have positive TTC4TS MSE reduction and TTC4TS MSE below half the variance of the true future. We select the observed gain nearest the 75th percentile of eligible gains, resolving ties by origin and channel order. This selection uses test targets only to choose an illustration; it does not alter the deployed predictor, validation choices, or aggregate results. The examples are not a random sample or an estimate of typical performance. Panels use fitting-standardized coordinates and individual vertical scales. Table~\ref{tab:forecast_examples} identifies the selected origins, channels and deployment rules; the accompanying CSV and manifest include timestamps, exact errors and source hashes.

\begin{table}[H]
\centering
\caption{Provenance for the six post-hoc forecast illustrations on ETTh1, ETTh2, ETTm1, ETTm2, Appliances, and BeijingWeather. All use horizon 96 and seed 0. Origin $t$ is a zero-based dataset index; channel MSEs use adapter-prefix standardization. The displayed rule was chosen on validation before these examples were selected.}\label{tab:forecast_examples}
\fontsize{9}{10.8}\selectfont
\setlength{\tabcolsep}{3.0pt}
\renewcommand{\arraystretch}{1.04}
\begin{tabular}{llllrr}
\toprule
Panel / model & Dataset & Channel & Rule & $t$ & Gain (\%) \\
\midrule
(a) Bolt-S & ETTh1 & MUFL & Ensemble learning & 11808 & 22.9 \\
(b) Bolt-B & ETTh2 & MUFL & Val.-best & 14112 & 8.2 \\
(c) TimesFM & ETTm1 & HUFL & Val.-best & 53664 & 43.3 \\
(d) Timer & ETTm2 & OT & Ensemble learning & 46560 & 35.6 \\
(e) Sundial & Appliances & T4 & Ensemble learning & 19148 & 61.6 \\
(f) Chronos-2 & BeijingWeather & TEMP & Retro. & 35539 & 15.3 \\
\bottomrule
\end{tabular}
\end{table}

\clearpage
\clearpage

\clearpage
\subsection{Intervals on prespecified forecast windows}
\begin{figure}[H]
\centering\includegraphics[width=\linewidth]{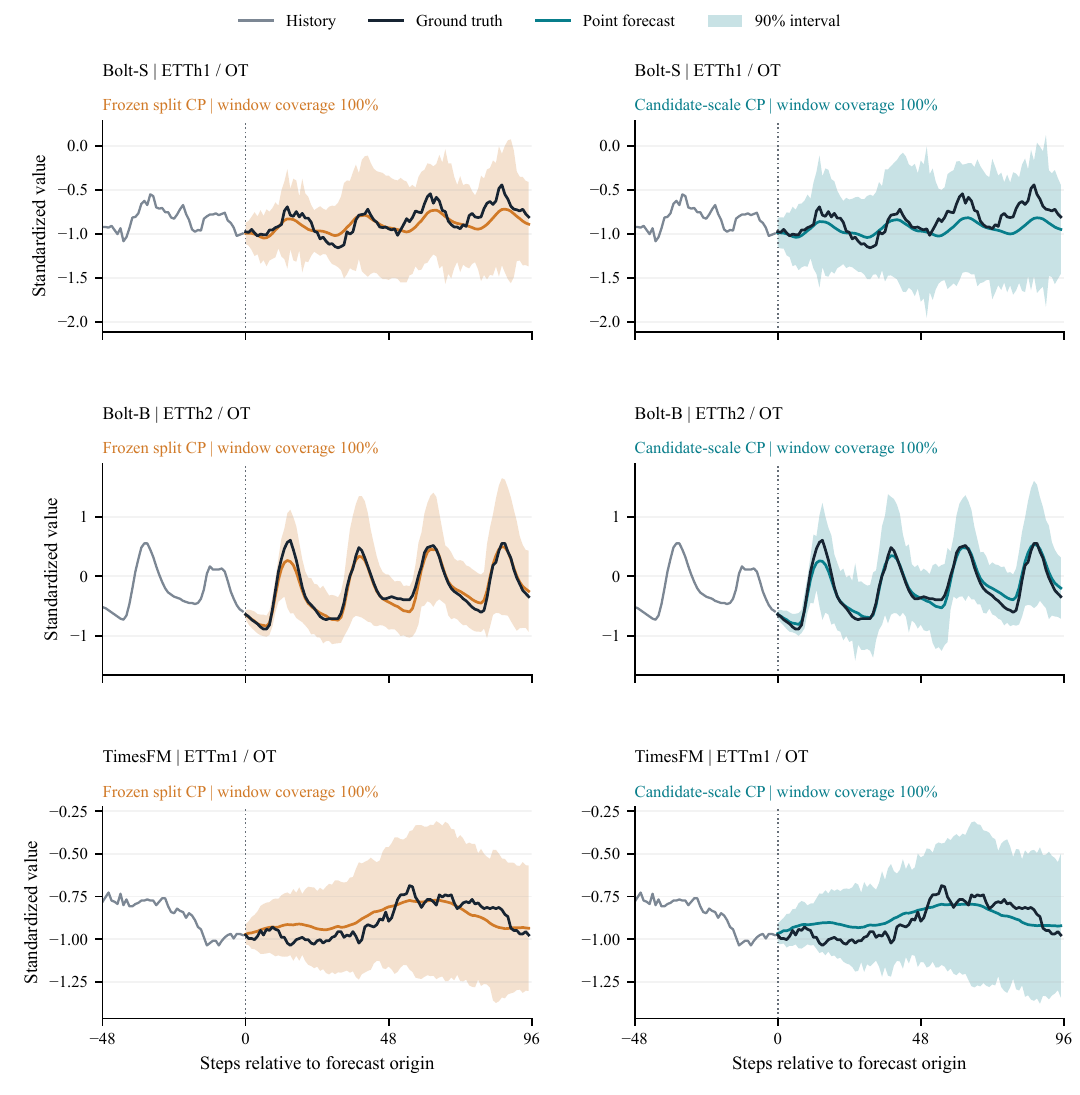}
\caption{Nominal 90\% intervals for Chronos-Bolt-small on ETTh1 (OT), Chronos-Bolt-base on ETTh2 (OT), and TimesFM-2.0 on ETTm1 (OT), at the first test origin, horizon 96, seed 0. Left: Frozen split CP. Right: Candidate-scale CP around the refitted/selected TTC forecast. Shading denotes prediction intervals; black is ground truth, gray is observed history, and colored curves are point predictions. These windows are fixed in advance of the uncertainty results.}\label{fig:uq_intervals1}
\end{figure}

\begin{figure}[H]
\centering\includegraphics[width=\linewidth]{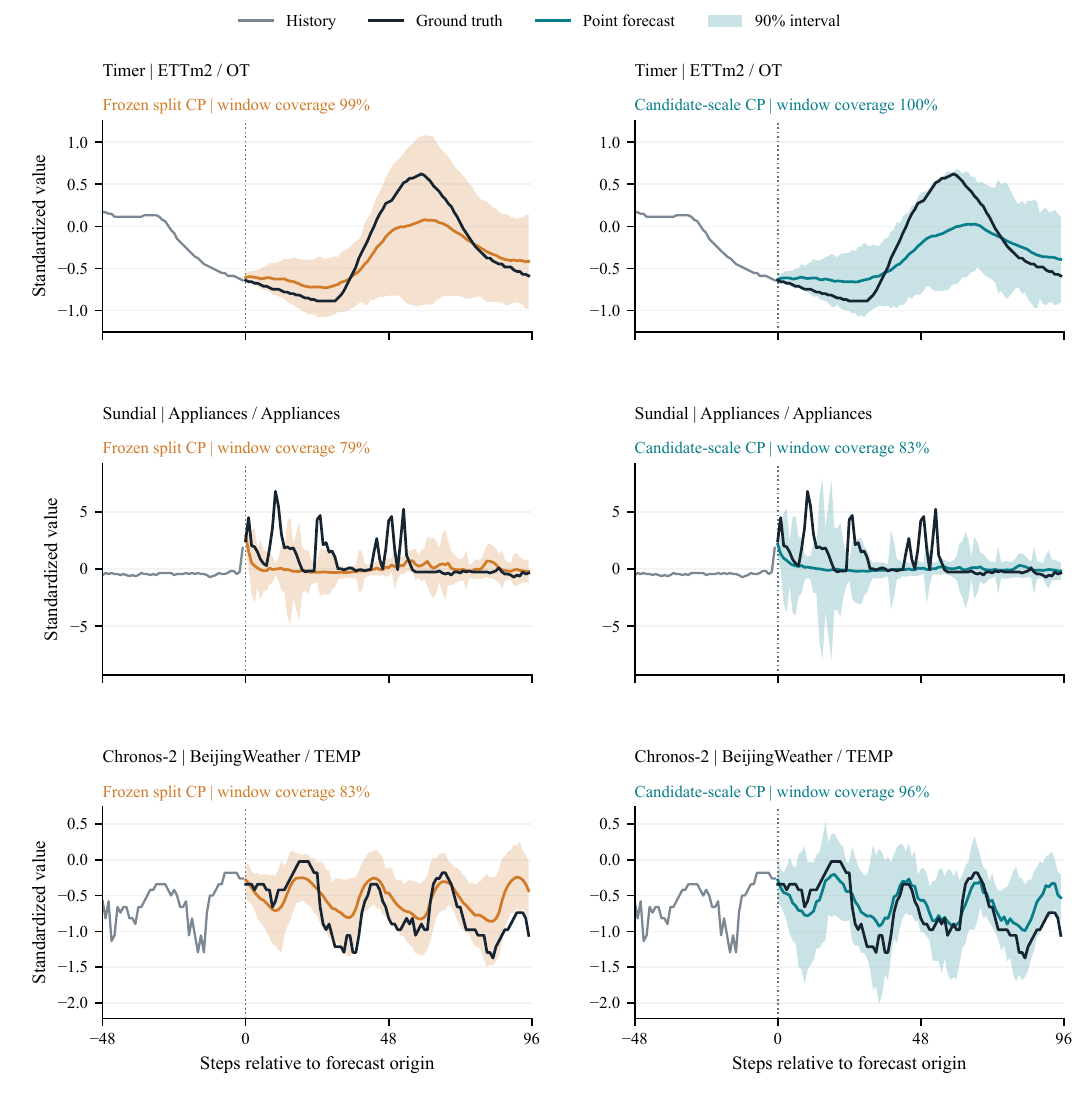}
\caption{Nominal 90\% intervals for Timer-base on ETTm2 (OT), Sundial-base on Appliances (Appliances), and Chronos-2 on BeijingWeather (TEMP), at the first test origin, horizon 96, seed 0. Left: Frozen split CP. Right: Candidate-scale CP. Colors, history length, shading, and matched vertical limits follow Figure~\ref{fig:uq_intervals1}. Coverage failures are retained rather than filtering examples by success.}\label{fig:uq_intervals2}
\end{figure}

Figures~\ref{fig:uq_intervals1} and~\ref{fig:uq_intervals2} show the first test origin for six prespecified model--dataset pairs, at seed 0 and horizon 96. The channel is OT for ETT, Appliances for the energy dataset, and TEMP for BeijingWeather. No origin or channel is selected by test coverage, interval width, or prediction error. Each row uses identical vertical limits for Frozen split CP and Candidate-scale CP, with 48 observed steps followed by the 96-step forecast. Annotations report coverage within that single window; they are descriptive, not evidence of nominal coverage for every query. All curves are extracted from saved experiment endpoints.
\clearpage


\section{Candidate and generator comparisons}
\IfFileExists{tables/candidates.tex}{\begin{table}[H]
\centering
\caption{Every individual generator operator on ETTh1, ETTh2, ETTm1, ETTm2, Appliances, and BeijingWeather, evaluated on identical test origins. No candidate is omitted for poor performance.}\label{tab:candidates}
\fontsize{8.7}{10.6}\selectfont
\setlength{\tabcolsep}{4pt}
\begin{tabular}{lrr@{\hspace{16pt}}lrr}
\toprule
Candidate & Rel. MSE & Rel. MAE & Candidate & Rel. MSE & Rel. MAE \\
\midrule
ewma & 1.029 & 1.044 & joint=0.284211 & 1.013 & 1.062 \\
horizontal & 1.380 & 1.236 & joint=0.3 & 1.015 & 1.067 \\
identity & 1.000 & \textbf{1.000} & magnitude flip & 0.995 & 1.005 \\
input=0.0157895 & 1.001 & \textbf{1.000} & magnitude warp & 0.981 & 1.003 \\
input=0.0315789 & 1.000 & \uline{1.001} & output=0.0157895 & 1.000 & \uline{1.001} \\
input=0.0473684 & 1.001 & 1.002 & output=0.0315789 & 1.000 & 1.002 \\
input=0.0631579 & 1.000 & 1.003 & output=0.0473684 & 1.000 & 1.004 \\
input=0.0789474 & 0.999 & 1.004 & output=0.0631579 & 1.000 & 1.005 \\
input=0.0947368 & 0.998 & 1.005 & output=0.0789474 & 1.000 & 1.008 \\
input=0.110526 & 0.998 & 1.006 & output=0.0947368 & 1.001 & 1.010 \\
input=0.126316 & 0.997 & 1.007 & output=0.110526 & 1.002 & 1.012 \\
input=0.142105 & 0.997 & 1.008 & output=0.126316 & 1.002 & 1.015 \\
input=0.157895 & 0.997 & 1.009 & output=0.142105 & 1.003 & 1.018 \\
input=0.173684 & 0.997 & 1.009 & output=0.157895 & 1.004 & 1.021 \\
input=0.189474 & 0.997 & 1.010 & output=0.173684 & 1.005 & 1.024 \\
input=0.205263 & 0.997 & 1.011 & output=0.189474 & 1.007 & 1.027 \\
input=0.221053 & 0.997 & 1.012 & output=0.205263 & 1.008 & 1.031 \\
input=0.236842 & 0.996 & 1.013 & output=0.221053 & 1.009 & 1.034 \\
input=0.252632 & 0.995 & 1.014 & output=0.236842 & 1.011 & 1.038 \\
input=0.268421 & 0.995 & 1.015 & output=0.252632 & 1.013 & 1.042 \\
input=0.284211 & 0.995 & 1.016 & output=0.268421 & 1.015 & 1.046 \\
input=0.3 & 0.994 & 1.017 & output=0.284211 & 1.017 & 1.051 \\
jitter & 0.999 & \uline{1.001} & output=0.3 & 1.019 & 1.055 \\
joint=0.0157895 & 1.000 & \uline{1.001} & retrieval=demo1 & \uline{0.977} & 1.004 \\
joint=0.0315789 & 1.000 & 1.003 & retrieval=demo10 & 1.013 & 1.018 \\
joint=0.0473684 & 1.001 & 1.005 & retrieval=demo2 & \textbf{0.971} & 1.006 \\
joint=0.0631579 & 1.001 & 1.008 & retrieval=demo3 & 0.983 & 1.010 \\
joint=0.0789474 & 0.999 & 1.011 & retrieval=demo4 & 1.001 & 1.012 \\
joint=0.0947368 & 0.999 & 1.014 & retrieval=demo5 & 0.991 & 1.011 \\
joint=0.110526 & 0.999 & 1.017 & retrieval=demo6 & 0.999 & 1.014 \\
joint=0.126316 & 1.000 & 1.020 & retrieval=demo7 & 0.999 & 1.014 \\
joint=0.142105 & 1.001 & 1.024 & retrieval=demo8 & 1.000 & 1.012 \\
joint=0.157895 & 1.002 & 1.027 & retrieval=demo9 & 0.999 & 1.014 \\
joint=0.173684 & 1.003 & 1.031 & slope=minus & 1.083 & 1.071 \\
joint=0.189474 & 1.004 & 1.035 & slope=plus & 1.108 & 1.083 \\
joint=0.205263 & 1.006 & 1.040 & spike & 0.999 & \uline{1.001} \\
joint=0.221053 & 1.007 & 1.044 & time=flip & 2.108 & 1.588 \\
joint=0.236842 & 1.008 & 1.048 & time=warp & 1.269 & 1.160 \\
joint=0.252632 & 1.010 & 1.053 & vertical=minus & 1.000 & \textbf{1.000} \\
joint=0.268421 & 1.011 & 1.057 & vertical=plus & 1.000 & \textbf{1.000} \\
\bottomrule
\end{tabular}
\end{table}
}{}

\end{document}